%% file: main.tex
\documentclass[11pt, a4paper, logo, copyright, nonumbering]{xiaomi}

\usepackage[authoryear, sort&compress, round]{natbib}
\usepackage{dblfloatfix}
\usepackage{ulem}
\usepackage{caption}
\usepackage{dramatist}
\usepackage{xspace}
\usepackage{pifont} 
\usepackage{multirow}
\usepackage{tcolorbox}
\usepackage{xltabular}
\usepackage{longtable}
\usepackage{hyperref}
\usepackage{wrapfig}

\usepackage{amsfonts}
\usepackage{amsmath}
\usepackage{amssymb}
\usepackage{lineno}
\usepackage{multirow}
\usepackage{adjustbox}

\usepackage[bottom]{footmisc}

\usepackage{CJKutf8}
\usepackage{setspace}
\usepackage{makecell}
\usepackage{graphicx}
\usepackage{subcaption}
\usepackage{multicol} 

\usepackage[utf8]{inputenc} 
\usepackage[T1]{fontenc}    
\usepackage{hyperref}
\usepackage{cleveref} 
\usepackage{url}            
\usepackage{booktabs}       
\usepackage{amsfonts}       
\usepackage{nicefrac}       
\usepackage{microtype}      
\usepackage{xcolor}         
\usepackage{colortbl} 
\usepackage{tcolorbox}
\usepackage[misc]{ifsym}
\usepackage{float}
\definecolor{BrickRed}{rgb}{.72,0,0}
\definecolor{darkgreen}{rgb}{0.0, 0.5, 0.0}
\definecolor{ForestGreen}{RGB}{34,139,34}
\definecolor{LakeBlue}{RGB}{0,61,153}
\definecolor{MiOrange}{RGB}{255,225,204}
\definecolor{Hex}{RGB}{225,213,231}

\hypersetup{
    colorlinks=true,
    allcolors=LakeBlue
}

\newcommand{\METHODNAME}{Q-Mask: Query-driven Causal Masks for Text Anchoring in OCR-Oriented Vision-Language Models}

\title{\vspace{-1.8cm}\centering \METHODNAME~}

\author{
    {
    \vspace{-0.66cm}
        Longwei~Xu$^*$~
        Feng~Feng$^*$~
        Shaojie~Zhang$^*$~
        Xin~Chen$^*$~
        Hang~Li
        Anan~Du~
        Hailong~Yu~
        Pei~Fu~
        Zhenbo~Luo$^\dagger$~
        Jian~Luan
    } \\
MiLM Plus, Xiaomi Inc\\
~~
}

\begin{document}

{\let
    \thefootnote \relax 
    \footnote{$^*$ Equal contribution; $\dagger$ Corresponding author.} \\
    \footnote{$^{\textrm{\Letter}}$\{xulongwei, fengfeng6, shaojiezhang5, chenxin17,luozhenbo, luanjian\}@xiaomi.com}
    \setcounter{footnote}{0}\footnotetext{TABench is publicly available at \url{https://huggingface.co/datasets/loongwayX/TABench}.}
}

\vspace{-0.3cm}\input{main/sections/abstract}

\maketitle
\begin{figure}[H]
    \vspace{-0.5cm}
    \centering
    \includegraphics[width=0.75\linewidth]{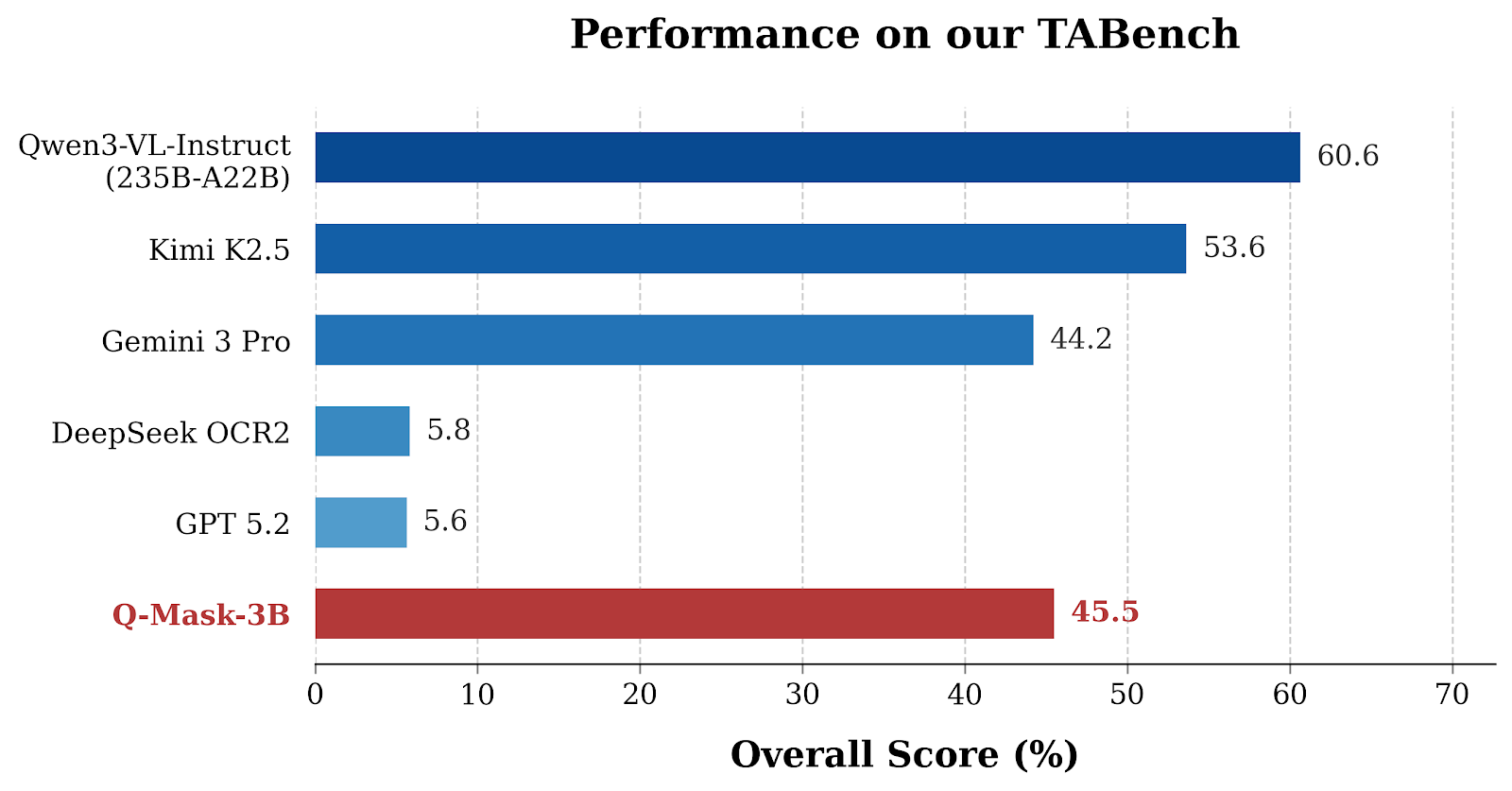}
    \caption{Performance comparison of mainstream general-purpose VLMs~\cite{bai2025qwen3vl,team2026kimi,gemini3pro2025,openai2025gpt5.2} and OCR-specific VLMs~\cite{wei2026deepseek} on the proposed TABench.\protect\footnotemark}
    \label{fig: benchmark}
\end{figure}

\input{main/sections/introduction}
\input{main/sections/related_work}
\input{main/sections/method}
\input{main/sections/TextAnchorBench}
\input{main/sections/Experiments}
\input{main/sections/conclusion}
\input{main/sections/license}
\clearpage
\suppressfloats[t]
\begin{center}
    \LARGE \bf Supplementary Material 
\end{center}
\vspace{1cm}

\setcounter{section}{0}
\setcounter{figure}{0}
\setcounter{table}{0}
\renewcommand{\thesection}{S\arabic{section}}
\renewcommand{\thefigure}{S\arabic{figure}}
\renewcommand{\thetable}{S\arabic{table}}

\input{Supplymentary_Material/sections/details_of_TABench}
\input{Supplymentary_Material/sections/details_of_TA-26M}
\input{Supplymentary_Material/sections/details_of_training}
\input{Supplymentary_Material/sections/Q-Mask_Visualization}
\clearpage
\bibliography{merged_reference} 

\end{document}

%% file: main/sections/abstract.tex
\begin{abstract}
\vspace{-0.5cm}
Optical Character Recognition (OCR) is increasingly regarded as a foundational capability for modern vision-language models (VLMs), enabling them not only to read text in images but also to support downstream reasoning in real-world visual question answering (VQA). However, practical applications further require reliable text anchors, i.e., accurately grounding queried text to its corresponding spatial region. To systematically evaluate this capability, we introduce TextAnchor-Bench (TABench), a benchmark for fine-grained text–region grounding, which reveals that both general-purpose and OCR-specific VLMs still struggle to establish accurate and stable text anchors. To address this limitation, we propose Q-Mask, a precise OCR framework built upon a causal query-driven mask decoder (CQMD). Inspired by chain-of-thought reasoning, Q-Mask performs causal visual decoding that sequentially generates query-conditioned visual masks before producing the final OCR output. This visual CoT paradigm disentangles where the text is from what the text is, enforcing grounded evidence acquisition prior to recognition and enabling explicit text anchor construction during inference. To train CQMD, we construct TextAnchor-26M, a large-scale dataset of image–text pairs annotated with fine-grained masks corresponding to specific textual elements, encouraging stable text–region correspondences and injecting strong spatial priors into VLM training. Extensive experiments demonstrate that Q-Mask substantially improves text anchoring and understanding across diverse visual scenes.
\end{abstract}

%% file: main/sections/introduction.tex
\section{Introduction}
Optical Character Recognition (OCR) serves as a fundamental bridge between visual perception and language understanding, enabling machines to transform textual information embedded in images into structured, machine-readable representations that support downstream reasoning, retrieval, and decision-making across numerous real-world applications such as document understanding, scene text analysis, and human–computer interaction~\cite{subramani2011survey,shen2023survey,huang2024detection}. More recently, with the evolution of large language models (LLMs) and vision language models (VLMs), OCR has been considered as a foundational capability, enabling models to perceive and reason about textual elements in complex visual scenes and documents~\cite{achiam2023gpt,Monkey,bai2023qwenvl,Instructdoc,wang2024qwen2vl,bai2025qwen3vl}. 

However, in real-world visual question answering (VQA) scenarios, it is not only necessary for VLMs to extract text, but also to establish accurate text anchors, which means the corresponding region of the text. For example, for interactive devices such as smart glasses, the model must not only understand the text embedded in images but also accurately identify the spatial locations of the text in the visual scene to enable precise interaction and operation. To systematically evaluate the ability of state-of-the-art (SOTA) VLMs~\cite{bai2025qwen3vl,team2026kimi,gemini3pro2025,wei2026deepseek,openai2025gpt5.2} to accurately anchor text to its corresponding spatial regions in images, we propose a comprehensive benchmark, TextAnchor-Bench (\textbf{TABench}). As shown in Fig.~\ref{fig: benchmark}, despite substantial advances in general-purpose VLMs~\cite{bai2025qwen3vl,team2026kimi,gemini3pro2025,openai2025gpt5.2} and OCR-specific VLMs~\cite{wei2026deepseek}, these models still exhibit limited capability to accurately establish text anchors and lack the ability to perceive the spatial distribution of text embedded in images.

\begin{figure}[t]
    \centering
    \includegraphics[width=0.98\textwidth]{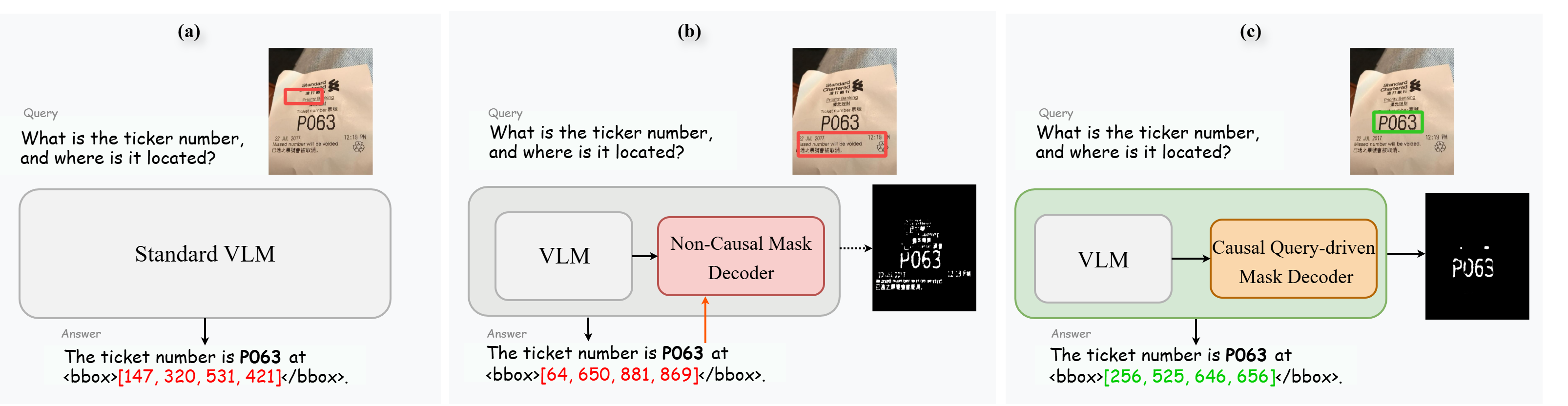}
    \caption{Comparison of the training paradigm of OCR-specific VLMs. (a) Standard VLMs~\cite{achiam2023gpt,Monkey,bai2023qwenvl,Instructdoc,wang2024qwen2vl,bai2025qwen3vl} can recognize text but lack explicit mechanisms to establish reliable text anchors. (b) Existing mask-based methods~\cite{Martern,TokenFD} enhance spatial perception but fail to explicitly model text–region correspondence during inference. (c) Q-Mask introduces a causal query-driven mask decoder (CQMD) that explicitly grounds the queried text token within its spatial region prior to recognition.}
    \label{fig: motivation}
\end{figure}

We suppose that the inability to use text anchors is closely related to the VLM's training paradigm. As shown in Fig.~\ref{fig: motivation} (a), standard VLMs~\cite{achiam2023gpt,Monkey,bai2023qwenvl,Instructdoc,wang2024qwen2vl,bai2025qwen3vl} adopt large numbers of image-text pairs for end-to-end training, enabling the model to develop image understanding and text recognition capabilities while remaining unconcerned about the text anchor within the image. Although existing VLMs enhance text anchor capabilities by incorporating coordinate-based supervision, they still lack a robust, consistent mechanism for constructing stable text anchors during downstream tasks. Moreover, as shown in Fig.~\ref{fig: motivation} (b), recent methods~\cite{Martern,TokenFD} design a pre-training task and a non-causal mask decoder that use character masks in the image as additional supervision signals to enhance spatial perception, but they can't explicitly capture text–region correspondence or establish explicit text anchors during inference. Therefore, it remains fundamentally challenging for current VLMs to explicitly establish reliable text anchors and fully leverage fine-grained spatial priors for precise text–region grounding in complex visual scenes.

To enable VLMs to explicitly establish reliable text anchors and exploit fine-grained spatial priors, we introduce Q-Mask in this paper, a framework for precise OCR via a causal query-driven mask decoder (CQMD). Inspired by the success of chain-of-thought (CoT) reasoning in LLMs, we introduce a causal decoding process, in which the model sequentially generates the visual masks and final answers. As shown in Fig.~\ref{fig: motivation} (c), this visual CoT paradigm explicitly enforces a structured reasoning process that separates \emph{where} the text is located from \emph{what} the text content is, enabling the model to leverage fine-grained spatial priors for reliable text anchoring. As a core component, the CQMD predicts query-conditioned visual masks that anchor the queried text token to its corresponding spatial region before generating the final OCR output. To train the CQMD, we construct the \textbf{TextAnchor-26M} dataset, comprising large-scale image–text pairs annotated with fine-grained target masks that explicitly delineate the spatial regions corresponding to specific textual elements. This dataset provides explicit supervision for learning text–region correspondences and encourages VLMs to form stable textual anchors during training. In contrast to conventional image–text pre-training corpora, TextAnchor-26M incorporates spatial priors that narrow the gap between text recognition and precise spatial grounding. Building on this dataset, we develop Q-Mask, an OCR-specific VLM that progressively infers the spatial region associated with a queried text token by first localizing a candidate region likely to contain the target text, thereby establishing an explicit text anchor instead of directly predicting textual content from the entire image.

Our contributions can be summarized as follows:
\begin{itemize}
    \item We introduce \textbf{TABench}, a comprehensive benchmark designed to systematically assess the capability of VLMs to establish precise text anchors and perform fine-grained text–region grounding in complex visual scenes.
    
    \item We propose Q-Mask, a novel framework that explicitly constructs textual anchors through a structured reasoning process. By first localizing candidate text regions and then performing OCR conditioned on spatially grounded visual evidence, the framework effectively disentangles \emph{where} the text is located from \emph{what} the text is.
    
    \item Extensive experiments demonstrate that Q-Mask significantly improves text anchoring and understanding in diverse benchmarks, leading to more reliable text localization and recognition in complex and interactive visual scenes.
\end{itemize}

%% file: main/sections/related_work.tex
\section{Related Work}
\label{sec:related_work}

\subsection{OCR-oriented Vision-Language Models}
Multimodal large language models combine a visual encoder with an autoregressive language model to support perception and instruction following on text-rich images and documents.
Existing approaches can be broadly grouped into OCR-dependent and OCR-free paradigms.
OCR-dependent methods~\cite{LayTextLLM,MOAI,Cream,Instructdoc,Doclayllm} rely on external OCR engines and inject recognized texts, and sometimes geometric metadata, as auxiliary tokens, which may increase context length and propagate upstream errors.
In contrast, OCR-free approaches~\cite{Monkey,MiNi-Monkey,liu2024textmonkey} aim to directly map pixels to task outputs, and have explored high-resolution processing, compact visual tokenization, and specialized architectural components to better handle dense text and complex layouts, including token compression for long-context inputs~\cite{deepseekocr,deepseekocr-2}.
Recent OCR-oriented vision-language models increasingly emphasize explicit localization, such as text spotting and layout-guided parsing, as part of their primary design objectives~\cite{hunyuanocr,paddleocrvl,paddleocrvl1.5,niu2025mineru25decoupledvisionlanguagemodel}.
These developments suggest that performance on text-rich inputs, including dense documents, tables, formulas, and scene text, is often constrained by localization reliability, motivating methods that more explicitly incorporate spatial cues into vision-language reasoning.

\subsection{Spatial Supervision for Text Anchoring}
To improve text anchoring in vision-language models, recent work has introduced explicit spatial supervision and fine-grained region--text alignment beyond image-level objectives~\cite{CLIP,SigLIP,SAM,oCLIP}. In the OCR domain, ODM~\cite{ODM} improves detection and spotting through additional text--image alignment during pretraining, while TokenVL~\cite{TokenFD} introduces token-level supervision to associate linguistic units with localized visual regions. Mask-augmented formulations have also been explored for document question answering; Martern~\cite{Martern} proposes a VQA-style mask generation objective to provide explicit spatial supervision. However, in these designs, mask prediction is often conditioned on hidden states derived from the full question--answer sequence, which can couple the learned masks with answer tokens and reduce their suitability as a spatial prior available before autoregressive decoding. Motivated by this observation, we study a query-driven causal masking formulation in which spatial cues are predicted from image and query tokens prior to answer generation.

%% file: main/sections/method.tex
\section{Method}
\subsection{Overall Architecture}
\label{sec:overall_architecture}
\begin{figure*}[t]
  \centering
  \includegraphics[width=1.0\textwidth]{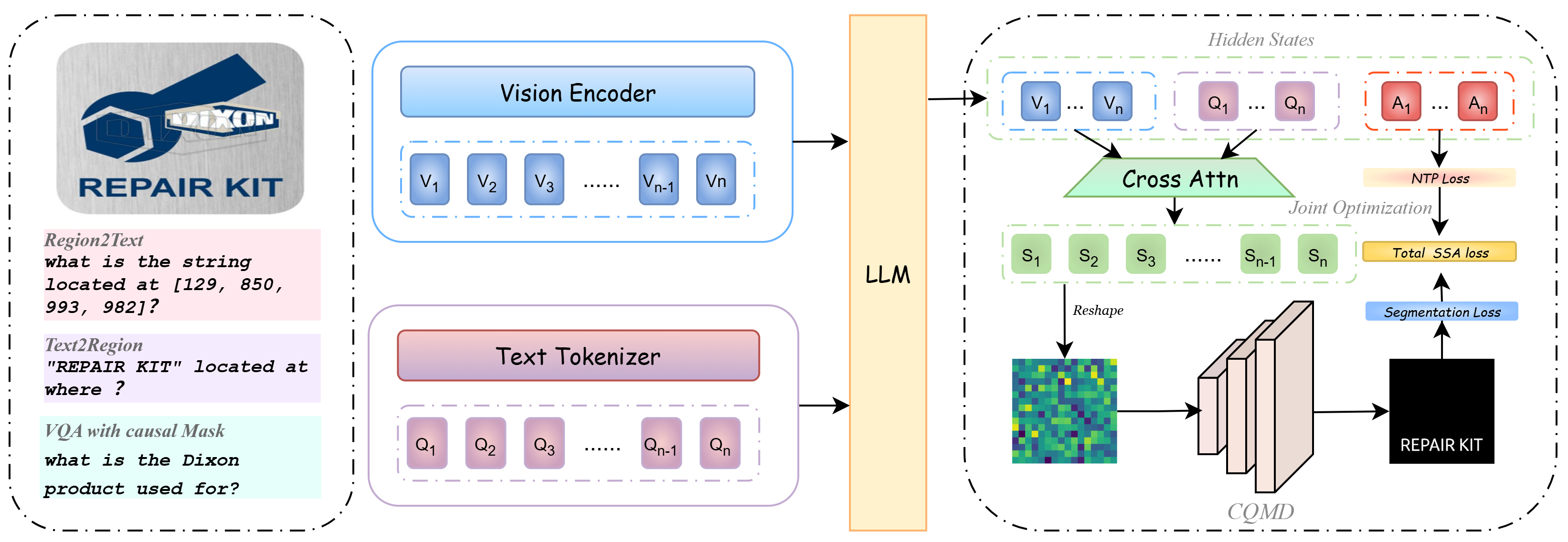}
  \caption{Overview of the proposed architecture. After a LLM processes the concatenated visual and textual embeddings, the CQMD module extracts only the hidden states of visual tokens ($\mathbf{H}_{img}$) and query tokens ($\mathbf{H}_{q}$). Using cross-attention, Q-Mask predicts spatial masks prior to autoregressive answer generation. The model is trained with the next-token prediction (NTP) loss and a segmentation loss.}
  \label{fig:overall_architecture}
\end{figure*}

Our framework builds upon a standard multimodal large language model (MLLM). Given a high-resolution image, the visual encoder produces visual embeddings $\mathbf{V} = [\mathbf{v}_1, \dots, \mathbf{v}_n]$. During training, the textual query and the target answer are tokenized into query embeddings $\mathbf{Q} = [\mathbf{q}_1, \dots, \mathbf{q}_l]$ and answer embeddings $\mathbf{A} = [\mathbf{a}_1, \dots, \mathbf{a}_m]$.

These embeddings are concatenated and fed into the LLM. Let $\mathbf{H}_{out} \in \mathbb{R}^{(n+l+m) \times d}$ denote the final hidden states produced by the last LLM layer. Instead of using the full hidden-state sequence, we retain only the contextualized hidden states corresponding to the visual tokens and the query tokens:
\begin{equation}
\mathbf{H}_{img} = \mathbf{H}_{out}[\mathcal{I}_{img}], \quad \mathbf{H}_{q} = \mathbf{H}_{out}[\mathcal{I}_{q}],
\end{equation}
where $\mathcal{I}_{img}$ and $\mathcal{I}_{q}$ denote the index sets of image and query tokens in the concatenated sequence, respectively. The resulting $\mathbf{H}_{img}$ and $\mathbf{H}_{q}$ serve as the only inputs to the proposed Causal Query-Driven Mask Decoder (CQMD), described below.

\subsection{Causal Query-Driven Mask Decoder (CQMD)}
To bridge the spatial gap between visual and language modalities, several recent approaches~\cite{TokenFD, Martern} have introduced auxiliary mask-generation modules. A common design is to condition mask prediction on hidden states derived from the \emph{entire} question--answer sequence, which can be abstracted as $\mathbf{H}_{QA} = [\mathbf{H}_{q}, \mathbf{H}_{a}]$, and to query visual features using $\mathbf{H}_{QA}$. This design is potentially misaligned with autoregressive inference: the full answer sequence is unavailable when spatial guidance is needed, so the resulting mask cannot be interpreted as a prior defined before answer decoding and may primarily act as a training-time regularizer.

We formalize the desired causal mechanism by introducing a latent spatial support variable $S$ (e.g., a mask or a set of relevant regions) that mediates between the image $I$ and the answer $A=(a_1,\dots,a_T)$ given the query $Q$:
\begin{equation}
P(A \mid I, Q) = \int P(A \mid S, Q)\, P(S \mid I, Q)\, dS,
\label{eq:causal_decomp}
\end{equation}
where $P(S\mid I,Q)$ is a query-driven spatial prior and $P(A\mid S,Q)$ generates the answer conditioned on the spatial support. In an autoregressive setting, a causal spatial prior should satisfy
\begin{equation}
S \perp A_{\text{future}} \mid (I,Q),
\label{eq:causal_indep}
\end{equation}
i.e., $S$ must not depend on answer tokens that have not yet been generated.

Motivated by the causal requirement in Eq.~\ref{eq:causal_indep}, we propose the Causal Query-Driven Mask Decoder (CQMD), which predicts spatial cues exclusively from the visual and query hidden states $\mathbf{H}_{img}$ and $\mathbf{H}_{q}$ extracted in Sec.~\ref{sec:overall_architecture}.
Concretely, CQMD computes query-aware spatial features via cross-attention from visual tokens to query tokens:
\begin{equation}
\mathbf{Attn} =
\operatorname{softmax}\!\left(\frac{(\mathbf{H}_{img}\mathbf{W}_{query})(\mathbf{H}_{q}\mathbf{W}_{key})^\top}{\sqrt{d}}\right)
(\mathbf{H}_{q}\mathbf{W}_{value}),
\label{eq:qmask_attn}
\end{equation}
\begin{equation}
\mathbf{S} = \operatorname{ReLU}(\mathbf{Attn}\mathbf{W}_1 + \mathbf{b}_1)\mathbf{W}_2 + \mathbf{b}_2,
\label{eq:qmask_mlp}
\end{equation}
where $\mathbf{W}_{query}, \mathbf{W}_{key}, \mathbf{W}_{value}, \mathbf{W}_1, \mathbf{W}_2$ are learnable projections, and $\mathbf{S}\in\mathbb{R}^{n\times d}$ denotes the resulting query-aware spatial features, shown as $S_1,\dots,S_n$ in Fig.~\ref{fig:overall_architecture}.

To enforce Eq.~\ref{eq:causal_indep}, Q-Mask is structurally prevented from conditioning on answer-token representations. Let $\mathbf{H}_{a}$ denote the hidden states of answer tokens in $\mathbf{H}_{out}$. Our construction enforces
\begin{equation}
\mathbf{S} = f(\mathbf{H}_{img}, \mathbf{H}_{q})
\quad \Longrightarrow \quad
\frac{\partial \mathbf{S}}{\partial \mathbf{H}_{a}} = \mathbf{0},
\label{eq:grad_block}
\end{equation}
which follows directly from the computational graph. This constraint means that $\mathbf{S}$ is computed independently of answer-token representations and should therefore be understood as a latent, query-conditioned spatial support established prior to answer decoding. In our implementation, this support is not provided to the language model as an explicit decoder input. Instead, mask supervision regularizes the shared representations so that autoregressive answer generation is implicitly grounded in query-conditioned spatial support.
\begin{figure}[tbp]
    \centering
    \includegraphics[width=1.0\linewidth]{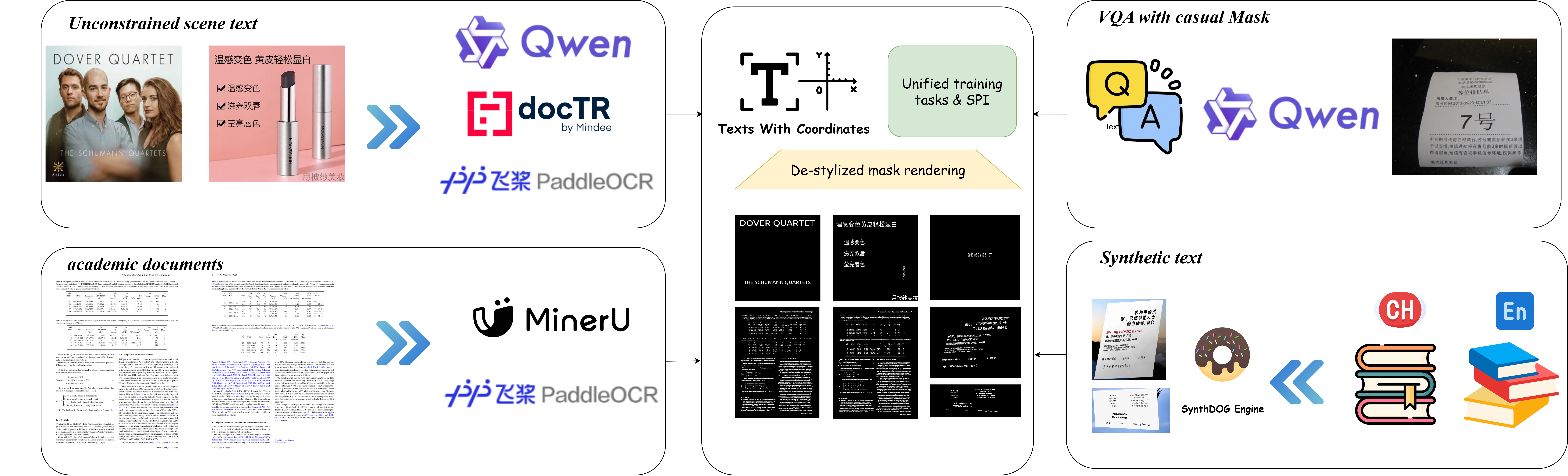}
    \caption{Overview of the TextAnchor-26M construction pipeline. We aggregate four data sources: (1) unconstrained scene text mined from large-scale web corpora; (2) academic documents (e.g., arXiv pages); (3) synthetic text rendered by SynthDog with multilingual fonts; and (4) VQA-with-causal-mask samples generated by prompting a VLM on precisely annotated regions. We first obtain transcripts and bounding boxes using expert models or rendering annotations, and then apply stochastic prior injection (SPI) and de-stylized mask rendering to produce unified supervision for Q-Mask training.}
    \label{fig:data_engine}
\end{figure}
\paragraph{Training objective.}
We reshape the 1D query-aware spatial features $\mathbf{S}$ into a 2D spatial feature map
$\mathbf{S}^{2D}=\operatorname{Reshape}(\mathbf{S})$
and apply an upsampling decoder $\phi$ (implemented with transposed convolutions)
to predict the segmentation mask:
\begin{equation}
\tilde{\mathbf{M}} = \phi(\mathbf{S}^{2D}).
\label{eq:qmask_decode}
\end{equation}
We optimize a \emph{Spatial Supervision Alignment (SSA)} objective that jointly
trains the language modeling and mask prediction branches:
\begin{equation}
\mathcal{L}_{\text{SSA}} = \lambda_{\text{txt}} \mathcal{L}_{\text{NTP}}
+ \lambda_{\text{seg}} \mathcal{L}_{\text{mask}},
\label{eq:ssa_loss}
\end{equation}
where $\mathcal{L}_{\text{NTP}}$ is the next-token prediction loss, and
$\mathcal{L}_{\text{mask}} = \mathcal{L}_{\text{Dice}}(\tilde{\mathbf{M}},\mathbf{M})
+ \mathcal{L}_{\text{CE}}(\tilde{\mathbf{M}},\mathbf{M})$ combines Dice loss and cross-entropy loss
applied to the predicted mask $\tilde{\mathbf{M}}$ and the ground-truth mask $\mathbf{M}$.

\subsection{TextAnchor-26M Dataset Construction}
\label{sec:textanchor_data}
Reliable text anchoring in vision-language models is constrained by the availability of scalable, high-fidelity localization supervision. To support Q-Mask training, we construct the \textbf{TextAnchor-26M} dataset, designed to align with the evaluation protocol of TABench and to provide diverse text regions with consistent geometric supervision. TextAnchor-26M contains approximately \textbf{26.7M} image--text instances with bounding boxes and corresponding masks, drawn from four sources: (i) unconstrained scene-text image--text pairs mined from large-scale multimodal corpora, including Wukong~\cite{gu2022wukong} and TextDiffuser-MARIO-10M~\cite{chen2024textdiffuser,chen2023textdiffuser}; (ii) academic document pages collected from arXiv and parsed by document analyzers~\cite{niu2025mineru25decoupledvisionlanguagemodel,paddleocrvl1.5} to obtain line-level regions and transcripts; (iii) typography-rich synthetic text rendered with SynthDog~\cite{kim2022ocr} using a large private corpus and diverse free-use fonts; and (iv) a small VQA-with-causal-mask subset generated by prompting a VLM on precisely annotated regions~\cite{yu2023icdar}. For the scene-text subset, we use an automated multi-engine agreement strategy. A high-confidence subset is first selected based on strict geometric consistency under a high IoU threshold, and a large vision-language model is then used as an auxiliary adjudicator for a subset of the remaining uncertain cases. Unresolved inconsistencies are discarded to reduce semantic noise. The fourth subset is used only in Stage~2 training, while the first three sources provide the main large-scale text-anchoring supervision.

\paragraph{De-stylized mask rendering.}
Instead of attempting to recover pixel-accurate stylized glyph contours, we render \emph{de-stylized} binary masks from transcripts and bounding boxes by drawing the text on a blank canvas with standard fonts and then scaling and aligning it to fit the target box. This design reduces sensitivity to low-level textures and encourages modality-agnostic alignment between linguistic content and spatial support.

\begin{figure}[t]
    \centering
    \includegraphics[width=0.95\linewidth]{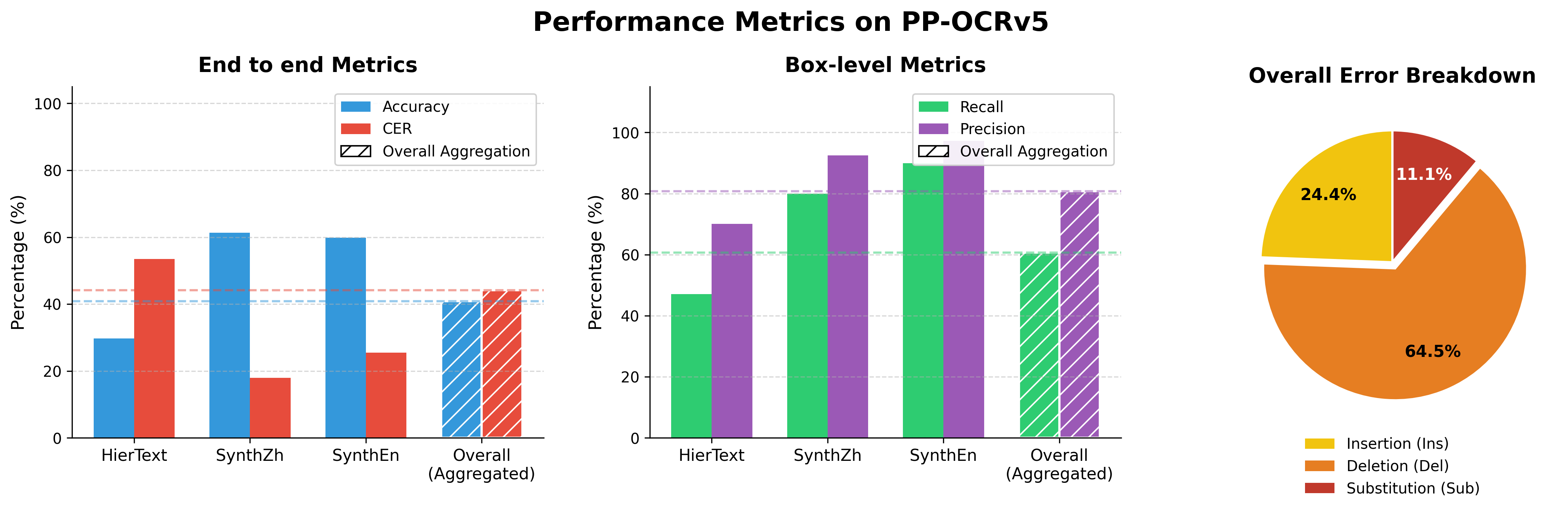}
    \caption{Empirical error profile of PPOCR-V5 on scene text. We decouple failures into \textbf{box-level} localization errors and \textbf{character-level} recognition errors. The character error rate is further decomposed into normalized proportions of insertion (\textbf{Ins}), deletion (\textbf{Del}), and substitution (\textbf{Sub}).}
    \label{fig:placeholder}
\end{figure}

\paragraph{Stochastic Prior Injection (SPI).}
Recent studies~\cite{mohammadshirazi2024dlava,lu2024bounding} suggest that injecting OCR-derived spatial priors (e.g., boxes with text cues) can improve answer accuracy and localization, motivating the use of OCR priors as a lightweight training augmentation. In TextAnchor-26M, raw OCR outputs are available for the scene-text subset as a by-product of pseudo-label generation. For the synthetic subset, we avoid running an additional OCR pipeline and instead \emph{simulate} OCR priors with an empirical noise model calibrated using PPOCR-V5~\cite{cui2025paddleocr} on HierText~\cite{long2022towards} and our bilingual synthetic data (Fig.~\ref{fig:placeholder}). We do not inject OCR priors for document pages due to token-budget constraints.

Let $\mathcal{V}=\{(b_i,t_i)\}$ denote ground-truth text instances and $\mathcal{V}_{raw}$ denote raw OCR outputs when available. During training, we sample $\gamma\in\{1.0,0.5,0.0\}$ for both scene-text and synthetic subsets and construct the injected prior set as
\begin{equation}
\tilde{\mathcal{V}}_{\gamma} =
\begin{cases}
\mathcal{V}_{raw}\ \text{(scene)}\ \text{or}\ \mathcal{V}\ \text{(synthetic)}, & \gamma=1.0 \\
\mathcal{V}_{keep}\cup \mathcal{F}(\mathcal{V}_{noise}), & \gamma=0.5 \\
\emptyset, & \gamma=0.0,
\end{cases}
\end{equation}
where $\mathcal{F}$ stochastically perturbs boxes and transcripts to emulate OCR outputs. This schedule exposes the model to priors that are present, noisy, or absent while keeping preprocessing cost manageable. Details of $\mathcal{F}$ are provided in the supplementary material.

\subsection{Training strategy}
A key property of Q-Mask is that mask prediction is conditioned only on image and query tokens, rather than on answer-token representations. This design allows us to apply spatial supervision when available, while still performing standard instruction tuning on datasets without spatial annotations.

In Stage~1, we focus on learning fine-grained spatial priors using supervision from TextAnchor-26M, where each sample provides aligned queries or transcripts together with bounding boxes or masks. We optimize a joint objective that combines next-token prediction with mask supervision for Q-Mask on these spatially annotated samples.

In Stage~2, we perform instruction tuning with a heterogeneous mixture of (i) standard VQA data without mask annotations, (ii) replayed samples from Stage~1, and (iii) high-quality VQA-with-causal-mask pairs. For samples without spatial annotations, training reduces to the next-token prediction loss. For samples with masks, we additionally supervise Q-Mask with the mask loss. Since Q-Mask does not access answer-token representations by construction, this mixed supervision enables the model to retain query-conditioned spatial cues while improving downstream reasoning.

%% file: main/sections/TextAnchorBench.tex
\section{TABench}
\label{sec:benchmark_construction}

We introduce \textbf{TABench}, a benchmark for evaluating whether a vision-language model can (i) accurately \textbf{read} the text within a specified region (\textbf{Region-to-Text}, R2T) and (ii) \textbf{localize} the region(s) corresponding to a given text query (\textbf{Text-to-Region}, T2R). TABench contains \textbf{5,450} queries in total, with exact balance between the R2T and T2R tasks, defined over the same set of \textbf{$970+$} core images. The benchmark is curated by sampling from four public datasets (HierText~\cite{long2023icdar}, SVRD~\cite{yu2023icdar}, CDLA~\cite{CDLA}, and ICDAR2015~\cite{karatzas2015icdar}) and covers \textbf{12} representative scenarios spanning both scene text and document-centric settings. Detailed statistics (source composition, language distribution, and granularity) are provided in the supplementary material.

\subsection{Evaluation metrics}
We report R2T accuracy and T2R performance. Specifically, we use the following metrics:
\begin{itemize}
    \item \textbf{R2T Accuracy ($Acc_{R2T}$).} We compute exact-match accuracy between the normalized prediction $\hat{y}_i$ and the ground-truth string $y_i$:
    \begin{equation}
        Acc_{R2T} = \frac{1}{N}\sum_{i=1}^{N}\mathbb{I}(\hat{y}_i = y_i).
    \end{equation}

    \item \textbf{T2R F1-score ($F1_{T2R}$).}
    Following the standard protocol in scene text detection and spotting benchmarks such as ICDAR2015~\cite{karatzas2015icdar}, we compute detection-style $F1$ using greedy bipartite matching under an IoU threshold of $0.5$.

    \item \textbf{Overall.} We summarize bidirectional capability by averaging $Acc_{R2T}$ and $F1_{T2R}$. If a model does not support one task direction or cannot produce a valid output for that direction, we assign a score of 0 for that metric when computing the overall score.
\end{itemize}

\subsection{Current MLLM performance on TABench}
\input{main/tables/sota_fineground}
Table~\ref{tab:sota_fineground} shows that even strong models exhibit a substantial imbalance between the two directions of text anchoring. For example, Gemini 3.0 Pro achieves moderate T2R localization ($F1_{T2R}=62.58\%$) but much lower R2T accuracy ($Acc_{R2T}=25.85\%$), suggesting that robust bidirectional text anchoring remains non-trivial.

%% file: main/tables/sota_fineground.tex
\begin{table}[htbp]
\centering
\caption{Evaluation of representative models on \textbf{TABench}. We report Region-to-Text exact-match accuracy ($Acc_{R2T}$) and Text-to-Region localization $F1$ ($F1_{T2R}$). \textbf{Overall} is defined as the arithmetic mean of $Acc_{R2T}$ and $F1_{T2R}$. ``--'' indicates the metric is not applicable because the model does not support the corresponding output format. For models that do not support one task direction, the missing metric is counted as 0 when computing Overall.}
\label{tab:sota_fineground}
\resizebox{\linewidth}{!}{
\begin{tabular}{l c c c c}
\toprule
\textbf{Model} & \textbf{Size} & $Acc_{R2T}$ (\%) $\uparrow$ & $F1_{T2R}$ (\%) $\uparrow$ & \textbf{Overall} (\%) $\uparrow$ \\
\midrule
Gemini 3.0 Pro~\cite{gemini3pro2025}        & Closed  & 25.85 & 62.58 & 44.22 \\
GPT 5.2~\cite{openai2025gpt5.2}               & Closed  & 10.64 &  0.64 &  5.64 \\
Kimi K2.5~\cite{team2026kimi}             & 1T      & 49.54 & 57.73 & 53.64 \\
Qwen3.5~\cite{qwen3.5}               & 397B    & 61.10 & 72.80 & 66.95 \\
Qwen3-VL-Instruct~\cite{bai2025qwen3vl}    & 235B    & 60.90 & 60.40 & 60.65 \\
\midrule
DeepSeek OCR2~\cite{deepseekocr-2}         & 3B      &  --   & 11.66 &  5.83   \\

Qwen3-VL-Instruct~\cite{bai2025qwen3vl}  & 2B      & 38.35 & 37.19 & 37.77 \\
\midrule
\rowcolor{gray!10}
\textbf{Q-Mask(Ours)}         & \textbf{3B} & \textbf{50.64} & \textbf{40.36} & \textbf{45.50} \\
\bottomrule
\end{tabular}
}
\end{table}

%% file: main/sections/Experiments.tex
\section{Experiments}
\subsection{Experimental Setup}
We use AdamW with a cosine learning-rate schedule in all training stages. Across all experiments, CQMD has a fixed size of \textbf{153.76M} parameters. For runtime benchmarking, we fix the input resolution and output length across methods and report the average latency over 10,000 cases. Under these identical settings, Q-Mask increases end-to-end latency by approximately \textbf{2.71\%} relative to the corresponding backbone without Q-Mask.

The Q-Mask-2B and Q-Mask-3B models are initialized from Qwen3-VL-2B-Instruct and Qwen2.5-VL-3B-Instruct, respectively, while the parameters of CQMD are randomly initialized. We follow a two-stage training recipe. In Stage~1, we train all parameters for one epoch on TextAnchor-26M with a base learning rate of $3\times10^{-5}$. In Stage~2, we freeze the vision encoder (ViT) and train for one epoch on an open-source OCR corpus containing \textbf{22.3M} samples, using a base learning rate of $5\times10^{-6}$. Detailed information on the data distribution and sources of this 22.3M OCR corpus is provided in the supplementary material.

All experiments are conducted on 64 NVIDIA H800 GPUs. Unless otherwise specified, we follow the official evaluation protocols of each benchmark and use a unified prompting template across models. We do not use external OCR engines or test-time tool augmentation during evaluation, so the reported results reflect the model’s internal perception and text anchoring ability.

\subsection{Effectiveness of Q-Mask}
\input{main/tables/VQA.tex}
We compare Q-Mask with representative multimodal large language models on multiple VQA and document understanding benchmarks (Table~\ref{tab:main_results}). On TextVQA, Q-Mask-3B achieves 88.5, improving upon the Qwen2.5-VL-3B baseline by 9.2 points. We attribute the gain primarily to (i) TextAnchor-26M, which provides large-scale text-centric supervision, and (ii) Stage-1 spatial pre-training that encourages the model to establish query-conditioned spatial cues before answer generation. We observe consistent improvements when applying the same training recipe to different backbones (e.g., Qwen2.5-VL-3B and Qwen3-VL-2B), suggesting that the paradigm is not specific to a particular base model.

In terms of parameter efficiency, Q-Mask-3B is competitive with larger open-source generalist models (e.g., InternVL3.5-8B~\cite{wang2025internvl3}) on text-heavy and visually grounded evaluations. We also compare against OCR-oriented models that incorporate spatial supervision (e.g., Martern~\cite{Martern} and the TokenFD series~\cite{TokenFD}). While these approaches employ masks during training, they are typically not designed to ensure that the learned spatial cues remain available as a query-conditioned prior throughout subsequent training and downstream decoding. Our results indicate that enforcing query-only conditioning in Q-Mask provides a practical way to preserve and leverage spatial cues for reasoning.

\subsection{Ablation Studies}

\subsubsection{w/o CQMD}
Table~\ref{tab:qmask_data_ablation} isolates the effect of CQMD under a fixed Stage-1 data budget. With the same 10M Stage-1 samples, enabling CQMD yields consistent gains on VQA-style benchmarks (notably TextVQA(+7.33) and InfoVQA(+3.18)), suggesting that query-conditioned spatial cues learned in Stage-1 transfer to downstream reasoning tasks.

\subsubsection{w/o SPI}
Table~\ref{tab:spi_ablation} evaluates SPI under a zero-shot protocol. SPI increases $Acc_{R2T}$ from 46.53 to 49.90 (+3.37) and T2R recall from 39.46 to 44.56 (+5.10), while $F1_{T2R}$ decreases from 40.09 to 35.00 due to more false positives under bipartite matching. Under our overall score definition, SPI yields a net improvement from 42.02 to 43.15 (+1.13), indicating more balanced bidirectional performance.

\begin{table*}[t]
\centering
\begin{minipage}{0.49\textwidth}
\centering
\caption{Ablation of CQMD under a fixed Stage-1 data budget (10M samples). Both settings share the same backbone and Stage-2 instruction tuning; the only difference is whether CQMD is enabled during Stage-1.}
\label{tab:qmask_data_ablation}
\input{main/tables/ablation}
\end{minipage}
\hfill
\begin{minipage}{0.49\textwidth}
\centering
\caption{Ablation of Stochastic Prior Injection (SPI) at the 5M checkpoint on TABench. Overall is $(Acc_{R2T}+F1_{T2R})/2$.}
\label{tab:spi_ablation}
\input{main/tables/SPI.tex}
\end{minipage}
\end{table*}

\begin{figure}[tbp]
    \centering
    \includegraphics[width=0.98\linewidth]{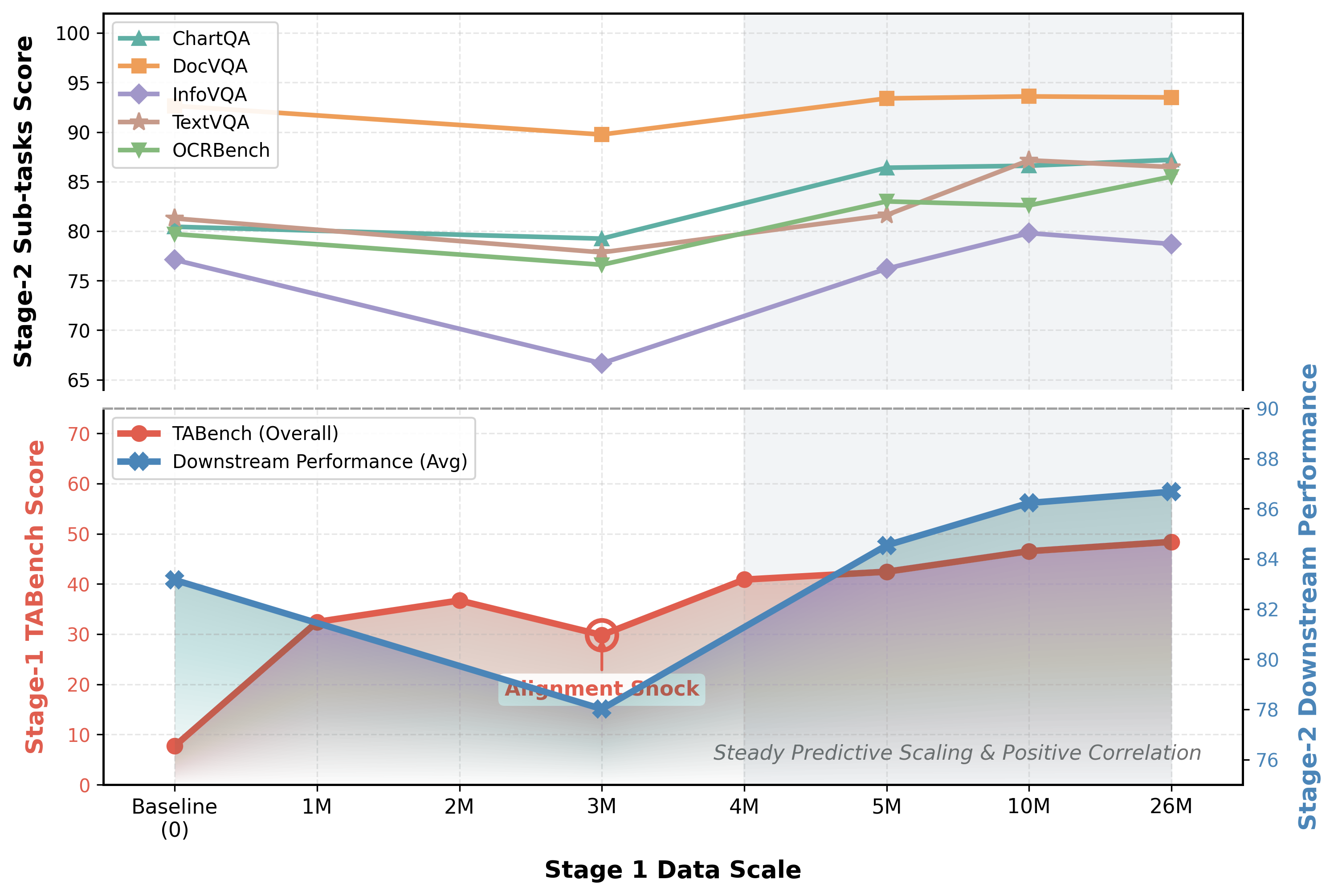}
    \caption{Stage-1 spatial pre-training signals versus Stage-2 downstream performance under a fixed Stage-2 recipe.}
    \label{fig:correlation_trajectory}
\end{figure}

\subsubsection{Scaling and Correlation of Stage-1 Spatial Supervision}
To understand how Stage-1 spatial supervision translates to downstream tasks, we vary the amount of Stage-1 data while keeping Stage-2 training identical. Figure~\ref{fig:correlation_trajectory} shows that Stage-1 TABench metrics correlate with the final Stage-2 performance, suggesting that TABench can serve as a lightweight diagnostic signal for predicting downstream gains. We further observe a non-monotonic regime around the 3M-sample scale, where $F1_{T2R}$ drops despite relatively stable $Acc_{R2T}$ and recall, indicating more false positives under bipartite matching. When the Stage-1 scale exceeds 4M, $F1_{T2R}$ and downstream performance recover and continue improving, suggesting that sufficient Stage-1 scale is important for stabilizing query-conditioned localization.



\subsection{Qualitative Analysis}
\label{sec:qualitative}

\begin{figure}[t]
    \centering
    \includegraphics[width=\linewidth]{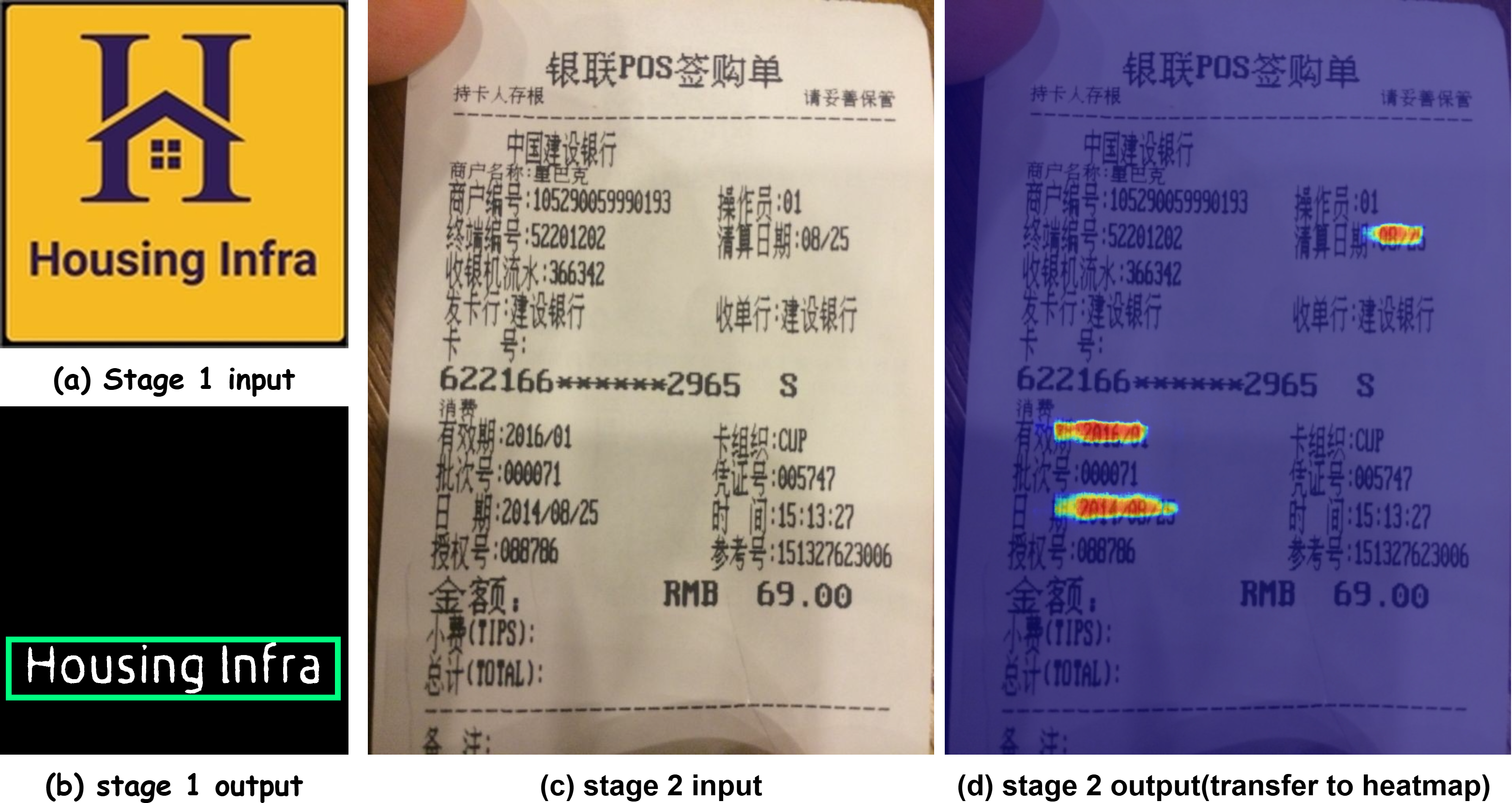}
    \caption{\textbf{Qualitative visualization of Q-Mask.} (a) Stage~1 input. (b) Stage~1 output, where the green box indicates the predicted text region. (c) Stage~2 input with the query \emph{``What is the transaction date?''}. (d) Stage~2 output visualized as a query-conditioned heatmap, highlighting candidate regions before the model generates the answer.}
    \label{fig:qualitative_vis}
\end{figure}

As shown in Fig.~\ref{fig:qualitative_vis}, we provide qualitative examples to illustrate how Q-Mask operates across the two-stage training pipeline. Figures~\ref{fig:qualitative_vis}(a--b) visualize the Stage~1 model on a text grounding example. The predicted bounding box (green) tightly encloses the target text, indicating that Stage~1 spatial supervision training effectively equips the model with strong localization ability. Figures~\ref{fig:qualitative_vis}(c--d) show a Stage~2 example on receipt VQA with the query \emph{``What is the transaction date?''}. Before generating the final answer, Q-Mask produces a query-conditioned heatmap that highlights multiple candidate regions relevant to the query. The model then correctly reads and outputs \emph{``2014/08/25''}. This example suggests that query-driven spatial cues learned in Stage~1 can be retained and leveraged during Stage~2 reasoning, providing grounded evidence prior to autoregressive decoding.

%% file: main/tables/VQA.tex
\begin{table*}[htbp]
\centering
\caption{Comprehensive evaluation on downstream multimodal reasoning and fine-grained document understanding benchmarks. By establishing an explicit "Look before Read" spatial prior, our 3B model significantly elevates the baseline performance. Strikingly, our 3B model consistently outperforms heavily engineered OCR-specialized models (e.g., TokenVL-8B, DocOwl) and even rivals or surpasses open-source generalist MLLMs that are more than twice its size (e.g., 7B--8B). \textbf{Bold} indicates the best results among models under 10B parameters.}
\label{tab:main_results}
\resizebox{0.95\textwidth}{!}{
\begin{tabular}{l c c c c c c }
\toprule
\textbf{Model} & \textbf{Size} & \textbf{ChartQA} & \textbf{DocVQA} & \textbf{InfoVQA} & \textbf{TextVQA} & \textbf{OCRBench} \\
\midrule
\multicolumn{7}{c}{\textit{Closed-source Generalist Models}} \\
\midrule
Gemini 3.0 Pro~\cite{gemini3pro2025}       & Closed & - & - & 57.2 & - & 940 \\
GPT 5.2~\cite{openai2025gpt5.2}            & Closed & 57.0 & 91.7 & 84.0 & 72.8 & 807 \\
Claude-3.5-Sonnet~\cite{anthropic2024claude35}  & Closed & 90.8 & 95.2 & 74.3 & 74.1 & 788 \\
GPT-4o~\cite{achiam2023gpt}             & Closed & 85.7 & 92.8 & 66.4 & 70.5 & 736 \\
Claude-3-Opus~\cite{anthropic2024claude3}      & Closed & 80.8 & 89.3 & 55.6 & 67.5 & 694 \\
\midrule
\multicolumn{7}{c}{\textit{Open-source Generalist MLLMs}} \\
\midrule
Qwen3-VL~\cite{bai2025qwen3vl}            & 235B(A22B) & 90.3 & 97.1 & 89.2 & - &920 \\
GLM-4.5V~\cite{vteam2026glm45vglm41vthinkingversatilemultimodal}            & 106B(A12B) & 86.6 & 94.5 & 84.1 & 72.0& 872 \\
InternVL3.5~\cite{wang2025internvl3}        & 78B & 89.7 & 95.4 & 86.5 & 84.3 & 906 \\
Qwen2.5-VL~\cite{bai2025qwen25vltechnicalreport}         & 72B & 89.5 & 96.4 & 87.3 & 83.5 & 885 \\
Molmo~\cite{deitke2025molmo}              & 72B & 87.3 & 93.5 & 81.9 & 83.1 & -   \\
\midrule
InternVL3.5~\cite{wang2025internvl3}        & 38B & 88.8 & 94.0 & 83.8 & 82.7 & 870 \\
Gemma3~\cite{gemma3_2025}             & 27B & 78.0 & 86.6 & 65.1 & 65.1 & 717 \\
InternVL3.5~\cite{wang2025internvl3}        & 20B & 86.6 & 92.9 & 78.1 & 78.5 & 870 \\
\midrule
Pixtral~\cite{mistral2024pixtral}            & 12B & 71.8 & 87.7 & 49.5 & 76.1 & -   \\
LLaMA3.2~\cite{dubey2024llama}           & 11B & 23.8 & 82.7 & 36.6 & 54.3 & -   \\
GLM-4.1V~\cite{vteam2026glm45vglm41vthinkingversatilemultimodal}           & 9B  & 70.0 & 93.3 & 80.3 & 79.6 & 823 \\
InternVL3.5~\cite{wang2025internvl3}        & 8B  & 86.7 & 92.3 & 79.1 & 78.2 & 840 \\
Qwen3-VL~\cite{bai2025qwen3vl}          & 8B  & 89.6 & 96.1 & 83.1 & 82.1 & 896 \\
Qwen2.5-VL~\cite{bai2025qwen25vltechnicalreport}         & 7B  & 87.3 & 95.7 & 82.6 & 84.9 & 864 \\
Qwen3-VL~\cite{bai2025qwen3vl}           & 4B  & 84.6 & 95.3 & 80.3 & 80.6 & 881 \\
InternVL3.5~\cite{wang2025internvl3}        & 4B  & 86.0 & 92.4 & 78.0 & 77.9 & 822 \\
\midrule
\multicolumn{7}{c}{\textit{Open-source OCR \& Document Specialist Models}} \\
\midrule
Monkey~\cite{Monkey}             & 10B & 65.1 & 66.5 & 36.1 & 67.6 & -   \\
TokenVL-8B~\cite{TokenFD}         & 8B  & 86.5 & 93.8 & 75.3 & 79.3 & 860 \\
Martern~\cite{Martern}            & 8B  & 81.7 & 92.0 & 75.2 & 74.4 & 820 \\
DocOwl-1.5-Chat~\cite{mplug-docowl1.5}    & 8B  & 70.2 & 82.2 & 50.7 & 68.6 & 599  \\
TextMonkey~\cite{liu2024textmonkey}         & 8B  & 66.9 & 73.0 & 28.6 & 65.6 & 561   \\
TextHawk2~\cite{yu2024texthawk2}          & 7B  & 81.4 & 89.6 & 67.8 & 75.1 & -   \\
DocKylin~\cite{zhang2024dockylin}           & 7B  & 66.8 & 77.3 & 46.6 & -    & -   \\
UReader~\cite{UReader}            & 7B  & 59.3 & 65.4 & 42.2 & 57.6 & -   \\
TokenVL-2B~\cite{TokenFD}         & 2B  & 81.1 & 89.9 & 61.0 & 76.4 & 821   \\
Mini-Monkey~\cite{MiNi-Monkey}        & 2B  & 76.5 & 87.4 & 60.1 & 75.7 & -   \\
HunyuanOCR~\cite{hunyuanocr}         & 0.9B& 78.5 & 86.8 & 61.6 & 71.1 & \textbf{860} \\
\midrule
Qwen2.5-VL~\cite{bai2025qwen25vltechnicalreport} & 3B  & 84.0 & 93.9 & 77.1 & 79.3 & 797 \\
\textbf{Q-Mask(Ours)} & 3B & \textbf{87.2} & 93.5 & \textbf{78.7} & \textbf{88.5} & 855 \\
Qwen3-VL~\cite{bai2025qwen3vl}           & 2B & 79.1 & 93.3 & 72.4 & 79.3 & 858 \\
\textbf{Q-Mask(Ours)}  & 2B  & 81.7 & 93.3 & 74.4 & 82.6 & 833 \\
\bottomrule
\end{tabular}
}
\end{table*}

%% file: main/tables/ablation.tex
\resizebox{\linewidth}{!}{
\begin{tabular}{l | ccccc}
\toprule
\textbf{Stage-1 Setting (10M)} & \textbf{ChartQA} & \textbf{DocVQA} & \textbf{InfoVQA} & \textbf{TextVQA} & \textbf{OCRBench} \\
\midrule
w/o CQMD & 84.92 & 93.09 & 76.69 & 81.20 & \textbf{831} \\
\rowcolor{gray!10}
\textbf{w/ CQMD} & \textbf{86.56} & \textbf{93.60} & \textbf{79.87} & \textbf{88.53} & 826 \\
\bottomrule
\end{tabular}
}

%% file: main/tables/SPI.tex
\resizebox{\linewidth}{!}{
\begin{tabular}{l cccc}
\toprule
\multirow{2}{*}{\textbf{Training Setting (5M)}} & \multicolumn{1}{c}{\textbf{R2T}} & \multicolumn{2}{c}{\textbf{T2R}} & \multirow{2}{*}{\textbf{Overall} $\uparrow$} \\
\cmidrule(lr){2-2} \cmidrule(lr){3-4}
& $Acc_{R2T}$ $\uparrow$ & Recall $\uparrow$ & $F1_{T2R}$ $\uparrow$ & \\
\midrule
Standard Training (w/o SPI) & 46.53 & 39.46 & \textbf{40.09} & 42.02 \\
\midrule
\rowcolor{gray!10}
\textbf{w/ SPI (Ours)} & \textbf{49.90} & \textbf{44.56} & 35.00 & \textbf{43.15} \\
\bottomrule
\end{tabular}
}

%% file: main/sections/conclusion.tex
\section{Conclusion}
We propose \textbf{Q-Mask}, a query-driven mask generation mechanism that predicts spatial cues solely from image and query tokens, avoiding answer-conditioned dependencies that can conflict with autoregressive decoding. To evaluate text anchoring, we introduce \textbf{TABench}, which jointly measures region-to-text reading and text-to-region grounding under matched visual conditions. To support scalable training of spatial priors, we construct \textbf{TextAnchor-26M}, a large-scale dataset with aligned transcripts, bounding boxes, and masks spanning unconstrained scene text, academic documents, and synthetic text, together with de-stylized mask rendering to encourage modality-agnostic supervision. Motivated by recent findings that OCR-derived priors can benefit VQA, we further propose \textbf{SPI} as a data augmentation strategy that simulates realistic OCR prior variability based on empirical OCR error profiles. Extensive experiments and ablations show that Q-Mask, TextAnchor-26M scale, and SPI contribute to consistent gains on VQA and document understanding benchmarks. We hope that the query-conditioned spatial cues produced by Q-Mask can serve as a useful intermediate representation for future studies on grounded reasoning and interpretability in vision-language models. Future work includes extending text anchoring to broader scripts and domains, and developing more token-efficient spatial priors for long documents.

%% file: main/sections/license.tex
\section{License and compliance note}
Academic document pages used in TextAnchor-26M are collected from publicly accessible arXiv papers for non-commercial academic research only. 

%% file: Supplymentary_Material/sections/details_of_TABench.tex
\section{TABench: Detailed Construction and Evaluation Protocol}
\label{sec:tabench_supp}

This section provides implementation-level details of \textbf{TABench} omitted from the main paper, including data sources, annotation workflow, scenario taxonomy, query construction, sampling strategy, and evaluation-time prompt / interface adaptation.

\subsection{Construction Pipeline}

TABench is constructed from four public datasets with text--region annotations: \textbf{HierText}~\cite{long2023icdar}, \textbf{SVRD}~\cite{yu2023icdar}, \textbf{ICDAR2015}~\cite{karatzas2015icdar}, and \textbf{CDLA}~\cite{CDLA}. For HierText, SVRD, and ICDAR2015, we directly use the original annotations. We include CDLA solely to improve Chinese coverage, which remains limited in existing open-source grounding benchmarks. Since the original CDLA annotations are layout-oriented rather than text-region grounding annotations, we first parse the selected pages with MinerU~\cite{wang2024mineruopensourcesolutionprecise} to obtain line-level text regions and then manually review and correct the resulting annotations where necessary. From CDLA, we retain only digitally native \emph{text-only} Chinese pages and exclude charts, tables, formulas, and embedded figures, because these elements often admit multiple valid textual serializations and may lead to ambiguous grounding targets.

The final benchmark contains 5,450 queries in total, with a strict 1:1 balance between the two task directions: 2,725 region-to-text (R2T) queries and 2,725 text-to-region (T2R) queries. These queries are defined over the same pool of more than 970 images.

\begin{figure}[t]
    \centering
    \includegraphics[width=\linewidth]{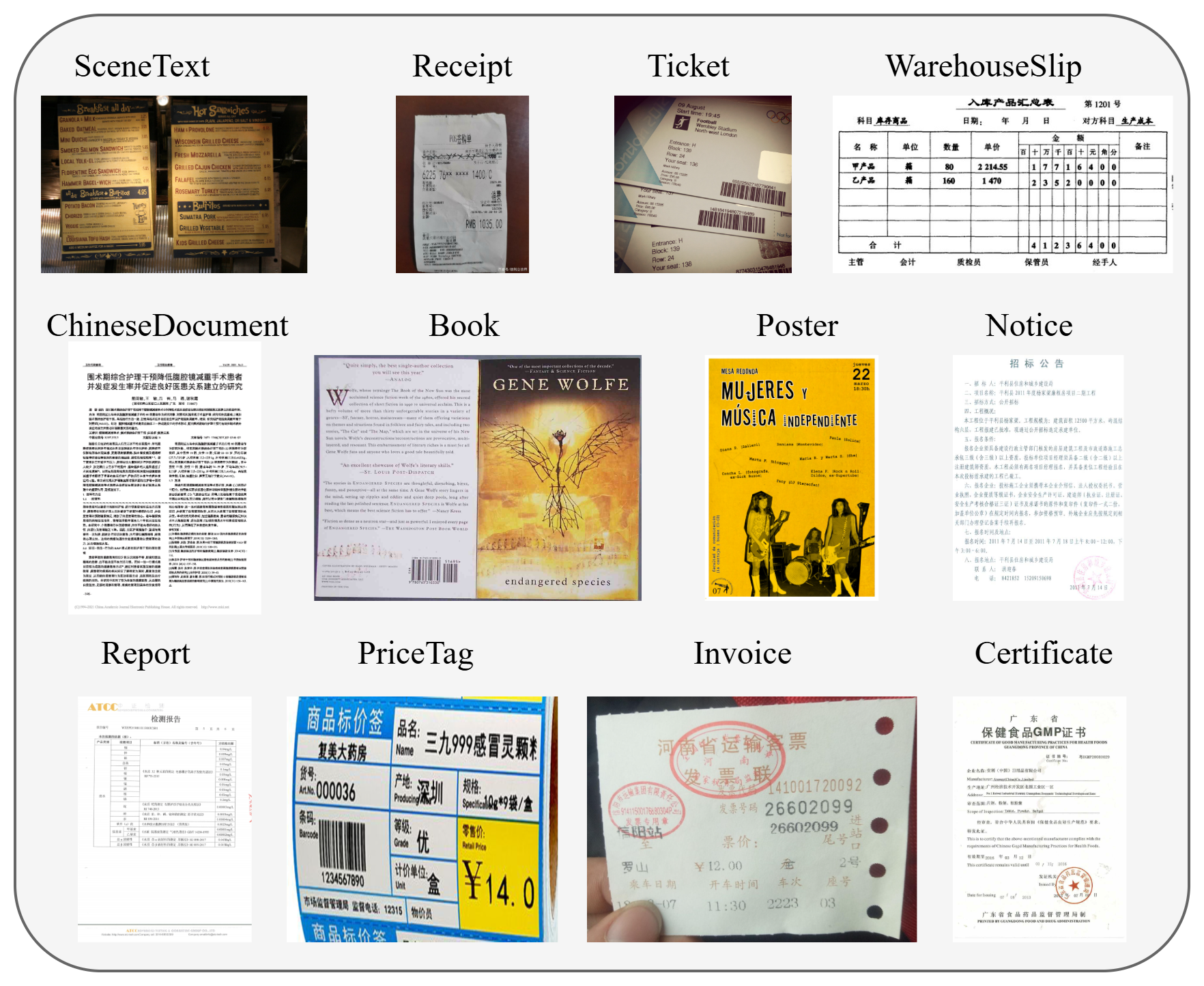}
    \caption{Representative examples of the 12 categories covered by TABench.}
    \label{fig:tabench_overview_supp}
\end{figure}

Each image is assigned to one of 12 categories: \texttt{SceneText}, \texttt{Receipt}, \texttt{Ticket}, \texttt{WarehouseSlip}, \texttt{Report}, \texttt{ChineseDocument}, \texttt{Book}, \texttt{Poster}, \texttt{Notice}, \texttt{PriceTag}, \texttt{Invoice}, and \texttt{Certificate}. Initial category labels are generated by Qwen3-VL-32B~\cite{bai2025qwen3vl} and then finalized through human verification. Representative examples of the 12 categories are shown in Figure~\ref{fig:tabench_overview_supp}.

Table~\ref{tab:tabench_stats_pair} summarizes the category-wise and global statistics of TABench. The benchmark covers both scene-text and document-centric settings, with English, Chinese, and mixed-language samples.

\begin{table*}[t]
\centering
\caption{Statistics of TABench. Left: category-wise counts with strict 1:1 parity between R2T and T2R. Right: global source and language distributions.}
\label{tab:tabench_stats_pair}
\begin{minipage}[t]{0.6\linewidth}
\centering
\resizebox{\linewidth}{!}{
\begin{tabular}{l r r l l}
\toprule
\textbf{Category} & \textbf{Samples} & \textbf{Ratio} & \textbf{R2T/T2R} & \textbf{Source} \\
\midrule
SceneText & 2{,}760 & 50.6\% & 1{,}380/1{,}380 & HierText 2{,}712; ICDAR2015 48 \\
Receipt & 560 & 10.3\% & 280/280 & SVRD 560 \\
Ticket & 460 & 8.4\% & 230/230 & SVRD 371; HierText 89 \\
WarehouseSlip & 390 & 7.2\% & 195/195 & SVRD 390 \\
Report & 280 & 5.1\% & 140/140 & SVRD 139; HierText 141 \\
ChineseDocument & 270 & 5.0\% & 135/135 & CDLA 270 \\
Book & 210 & 3.9\% & 105/105 & HierText 210 \\
Poster & 180 & 3.3\% & 90/90 & HierText 180 \\
Notice & 120 & 2.2\% & 60/60 & SVRD 112; HierText 8 \\
PriceTag & 80 & 1.5\% & 40/40 & SVRD 77; HierText 3 \\
Invoice & 80 & 1.5\% & 40/40 & SVRD 80 \\
Certificate & 60 & 1.1\% & 30/30 & SVRD 55; HierText 5 \\
\bottomrule
\end{tabular}}
\end{minipage}\hfill
\begin{minipage}[t]{0.3\linewidth}
\centering
\resizebox{\linewidth}{!}{
\begin{tabular}{@{}ll r@{}}
\toprule
\multicolumn{2}{@{}l}{\textbf{Distribution}} & \textbf{Count} \\
\midrule
\multirow{4}{*}{\textbf{Source Dataset}}
& HierText & 3{,}348 \\
& SVRD & 1{,}784 \\
& CDLA & 270 \\
& ICDAR2015 & 48 \\
\midrule
\multirow{2}{*}{\textbf{Lang. (R2T)}}
& English (EN) & 2{,}095 \\
& ZH / Mixed & 630 \\
\cmidrule(l){2-3}
\multirow{2}{*}{\textbf{Lang. (T2R)}}
& English (EN) & 2{,}017 \\
& ZH / Mixed & 708 \\
\bottomrule
\end{tabular}}
\end{minipage}
\end{table*}

All queries are generated deterministically from the underlying region annotations. For \textbf{R2T}, given an annotated bounding box $b=[x_{min},y_{min},x_{max},y_{max}]$, we use the query
\texttt{``What is the text at location $[x_{min},y_{min},x_{max},y_{max}]$?''}
and take the transcript of that region as the target.

\begin{figure}[t]
    \centering
    \includegraphics[width=\linewidth]{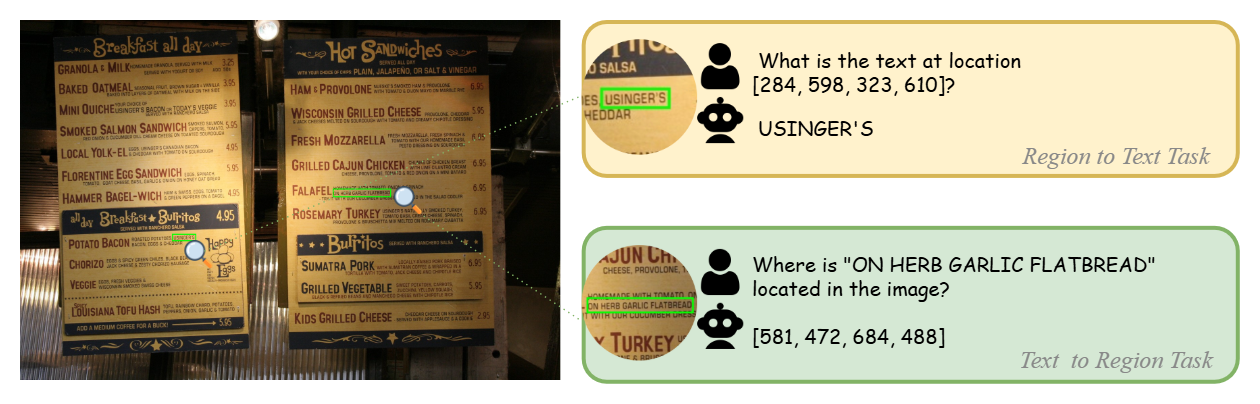}
    \caption{Illustration of the two tasks in TABench using representative single-instance examples. Region-to-Text (R2T) asks the model to read the text inside a specified region, while Text-to-Region (T2R) asks the model to localize the region corresponding to a queried text string.}
    \label{fig:tabench_task_supp}
\end{figure}

For \textbf{T2R}, given a text string $t$, we use the query
\texttt{``Where is `\{Text\}' located in the image?''}
and require the model to return bounding boxes in a standardized JSON format. Representative examples of the two task directions are shown in Figure~\ref{fig:tabench_task_supp}.

A main source of ambiguity in T2R is that the same text string may appear multiple times in one image. To avoid arbitrarily selecting a single target instance in such cases, we canonicalize text strings within each image by removing whitespace, stripping both English and CJK punctuation, and applying Unicode normalization (NFKC). All instances sharing the same canonicalized string are merged into one query, whose ground truth is defined as the set of all corresponding boxes in the image. During evaluation, predicted boxes are matched to ground-truth boxes using greedy bipartite matching under an IoU threshold of 0.5. Matched pairs are counted as true positives, while unmatched predicted boxes and unmatched ground-truth boxes are counted as false positives and false negatives, respectively. The final T2R score is computed by aggregating true positives, false positives, and false negatives over all T2R queries in the benchmark, and then reporting the resulting dataset-level F1 score.

After collecting candidate samples from the four source datasets and applying the post-processing rules above, we construct the final benchmark by random sampling under category-wise quotas, while enforcing equal numbers of R2T and T2R queries within each category. For each category, we first determine the target quota for each task direction and then sample from the corresponding candidate pool using a fixed random seed of 42. To keep the released benchmark compact and reproducible, the number of queries for each task direction within each category is rounded down to the nearest multiple of 5. This rounding affects only a small number of samples and does not materially change the overall distribution of the candidate pool.

\subsection{Evaluation Prompt and Interface Adaptation}

All applicable models are evaluated on the same TABench samples. We apply only interface-compatible adaptation required by each model's native interface, including prompt syntax, coordinate convention, and output parsing, without any label-driven prompt tuning or benchmark-specific prompt search. This design helps reduce confounding factors related to instruction-following and formatting compliance when comparing text anchoring ability across models.

For most evaluated MLLMs, including the Qwen series~\cite{bai2025qwen3vl, bai2025qwen25vltechnicalreport} and GLM-4.6V~\cite{vteam2025glm45vglm41vthinkingversatilemultimodal}, both input and output follow the standard box convention $(x_{\min},\allowbreak y_{\min},\allowbreak x_{\max},\allowbreak y_{\max})$. Different versions may expect either absolute pixel coordinates or relative coordinates; in such cases, we only convert coordinates to match the native interface while keeping the task definition unchanged. For models with native structured grounding outputs, such as the Qwen series, format-related parsing ambiguity is minimal in practice, since they directly support JSON-style fields such as \texttt{bbox\_2d} and \texttt{label}.

A few models require additional interface-specific handling. Gemini 3.0 Pro~\cite{gemini3pro2025} uses the coordinate order $(y_{\min}, x_{\min}, y_{\max}, x_{\max})$, so we reorder coordinates in the prompt and adjust the parser accordingly before converting predictions back to the canonical format used by our evaluation script. Kimi K2.5~\cite{team2026kimi} only supports normalized coordinates in $[0,1]$, so we convert both benchmark boxes and parsed outputs to the native normalized $[0,1]$ coordinate format. DeepSeek OCR2~\cite{wei2026deepseek} uses a dedicated grounding-style prompt rather than our generic natural-language template. For example, the T2R query

\texttt{Where is "WWW.NATIONALENQUIRER.COM" located in the image?}

is rewritten as

\texttt{Locate <|ref|>WWW.NATIONALENQUIRER.COM<|/ref|> in the image.}

For R2T, we extract the textual content when possible and otherwise fall back to the raw decoded string before normalization. We then normalize both predictions and ground-truth transcripts by applying Unicode normalization (NFKC), normalizing a small set of CJK punctuation variants, removing zero-width characters, and collapsing consecutive whitespace, so that the reported reading accuracy is less affected by superficial formatting differences.

For T2R, outputs that remain unparsable after the above adaptation---such as invalid JSON, missing boxes, or malformed numeric fields---are treated as empty predictions. Predicted boxes are then matched against the ground-truth box set using the IoU-based matching protocol defined in the main paper, with a threshold of 0.5.

Table~\ref{tab:prompt_adaptation} summarizes the interface differences and the corresponding evaluation adaptations.

\begin{table}[t]
\centering
\caption{Model-specific prompt and interface adaptations used in TABench evaluation. We only adapt prompt syntax, coordinate convention, and output parsing to match each model's native interface.}
\label{tab:prompt_adaptation}
\resizebox{\linewidth}{!}{
\begin{tabular}{l l l l}
\toprule
\textbf{Model family} & \textbf{Supported task(s)} & \textbf{Native coordinate/interface} & \textbf{Adaptation in TABench} \\
\midrule
Qwen-series & R2T + T2R & $(x_{\min},y_{\min},x_{\max},y_{\max})$; absolute or relative varies by version & coordinate conversion only \\
GLM-4.6V & R2T + T2R & $(x_{\min},y_{\min},x_{\max},y_{\max})$; relative coordinates in $[0,1000]$ & use native relative-coordinate format \\
Gemini 3.0 Pro & R2T + T2R & $(y_{\min},x_{\min},y_{\max},x_{\max})$ & reorder prompt/parser to native convention \\
Kimi K2.5 & R2T + T2R & normalized coordinates in $[0,1]$ & convert GT/predictions to normalized $[0,1]$ format \\
DeepSeek OCR2 & T2R & grounding prompt with \texttt{<|ref|>} tags & rewrite natural-language query to native grounding prompt \\
\bottomrule
\end{tabular}}
\end{table}

%% file: Supplymentary_Material/sections/details_of_TA-26M.tex
\section{TextAnchor-26M Construction}
\label{sec:data_engine}

\begin{figure}[t]
\centering
\includegraphics[width=\linewidth]{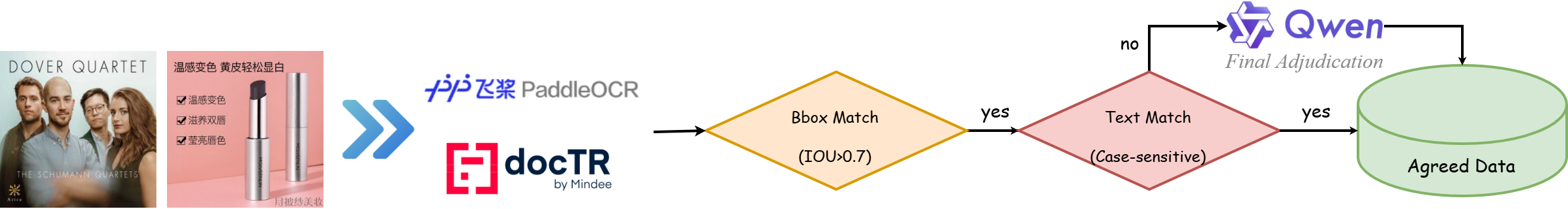}
\caption{
A simplified schematic of the pseudo-label generation pipeline for the scene-text subset of TextAnchor-26M.
Raw image--text pairs are first processed by two OCR engines (PPOCR-V5 and docTR).
Bounding boxes are retained only when the two predictions form a mutually best match with IoU $\ge 0.7$.
For retained candidates, transcripts are further verified by exact text matching.
Samples passing both checks are accepted into the agreed subset, while a subset of remaining transcript disagreements is sent to Qwen2.5-VL-72B for final adjudication.
}
\label{fig:scene_pipeline}
\end{figure}

TextAnchor-26M is constructed to provide large-scale spatial supervision for Q-Mask training while remaining aligned with the TABench evaluation protocol. 
The training corpus comprises four complementary components and contains approximately \textbf{26.7M} instances with bounding boxes and corresponding masks.
\paragraph{1. Unconstrained scene text.}
We mine raw image--text pairs from large-scale multimodal corpora, including Wukong~\cite{gu2022wukong} and TextDiffuser-MARIO-10M~\cite{chen2024textdiffuser, chen2023textdiffuser}. To generate spatial pseudo-labels, we apply a two-engine consensus pipeline based on PPOCR-V5~\cite{cui2025paddleocr} and docTR~\cite{doctr2021}. Each engine first produces candidate text boxes and transcripts. A candidate pair is retained only when the two boxes form a mutually best match with IoU $\ge 0.7$. For retained pairs, we compare the transcripts after whitespace normalization without case conversion. Exact matches are directly accepted, yielding approximately 19.02M agreed line-level box--text pairs. For a subset of the remaining disagreements, we use Qwen2.5-VL-72B as an auxiliary adjudicator to recover an additional 1.65M valid pairs. These numbers correspond to the intermediate pseudo-labeling yield; after deduplication, quality filtering, and corpus balancing, we retain 14.4M scene-text instances in the final training corpus. A schematic overview of this pipeline is shown in Fig.~\ref{fig:scene_pipeline}.

\paragraph{2. Academic documents and complex layouts.}
To improve coverage on dense document pages, we collect academic pages from arXiv and process them with MinerU~\cite{wang2024mineruopensourcesolutionprecise}. For regions with low parsing confidence (confidence $<0.7$), we additionally query PaddleOCR-VL~\cite{cui2025paddleocr} as an auxiliary validator to determine whether the corresponding page remains recoverable. This step is used only during data filtering: it helps retain pages whose low-confidence MinerU regions are still structurally consistent and discard pages with unresolved parsing failures. The final training annotations are kept in a unified MinerU-style line-level format, and no PaddleOCR-VL annotations are directly mixed into the final training annotations.

\paragraph{3. Typography-rich synthetic text.}
To increase font and style diversity, we synthesize text images with SynthDog~\cite{Kim2021SEMANTIC_Donut_Document_Understanding} using a large private corpus and a diverse collection of freely usable Chinese and English fonts. This subset covers both Chinese and English content and provides controlled typography-rich supervision that complements real-world scene text and academic documents.

\paragraph{4. VQA with causal mask.}
We further construct a small but high-quality VQA with causal mask subset from the training split of SVRD. Given an image together with text regions and their coordinates, we prompt a large VLM to generate 3--6 question--answer pairs. This subset contains approximately 10K samples. Unlike the other three sources, it is used \emph{only} in stage-2 training, where it provides semantic question-answering signals with explicit answer localization and supports reasoning over spatially grounded evidence. It is not used in first-stage training.

Across all sources, supervision is converted into a unified instance format containing the image, transcript or question--answer target, bounding box, binary mask, and source tag. The stage-1 training uses only the first three sources, while the VQA with causal mask subset is reserved for the second stage.

\begin{table}[t]
\centering
\caption{Composition of the TextAnchor-26M training data. The first three components are used in first-stage training, while the VQA with causal mask subset is used only in second-stage training.}
\label{tab:stage1_data}
\begin{tabular}{l r r l}
\toprule
\textbf{Component} & \textbf{\#Instances} & \textbf{Ratio} & \textbf{Usage} \\
\midrule
Scene text & 14.4M & 54.0\% & Stage-1 \\
Synthetic text & 10.8M & 40.4\% & Stage-1 \\
Academic document pages & 1.53M & 5.7\% & Stage-1 \\
VQA with causal mask & 10K & --- & Stage-2 only \\
\midrule
\textbf{Total} & \textbf{26.7M} & \textbf{100\%} & \\
\bottomrule
\end{tabular}
\end{table}

\begin{figure}[tbp]
    \centering
    \includegraphics[width=1.0\linewidth]{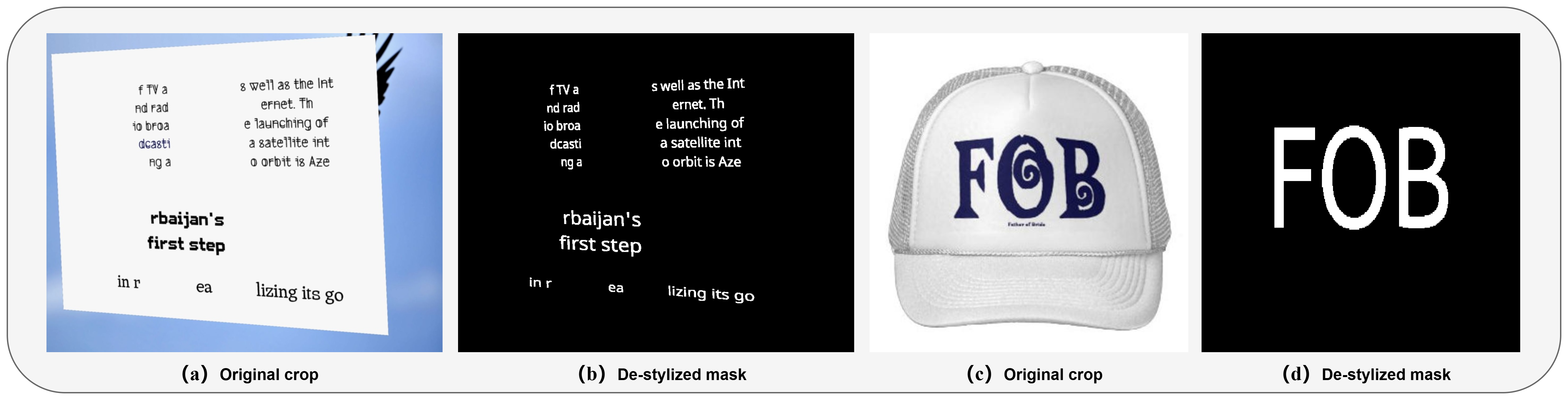}
    \caption{
    Examples of de-stylized mask rendering. For each pair, we show the original cropped text region and the corresponding rendered mask constructed only from the transcript and bounding box. The resulting mask preserves coarse spatial layout while removing source-specific style, color, texture, and background appearance.
}
    \label{fig:destylized_mask}
\end{figure}

\paragraph{De-stylized mask rendering.}
The de-stylized mask is constructed from the transcript and its corresponding bounding box. For Latin text, we use \textbf{Ubuntu Regular}; for Chinese text, we use \textbf{Noto Sans SC Regular}. Given a target box and transcript, we first compute the maximum font size that allows the rendered text to fit within the box. The text is then rendered on a blank canvas, cropped to its tight foreground region, and resized to match the spatial extent of the original annotation box. As illustrated in Fig.~\ref{fig:destylized_mask}, this procedure preserves approximate spatial alignment to the annotated text region while removing source-specific font style, texture, color, and other low-level appearance cues.

\subsection{SPI Details}

SPI is applied to the scene-text and synthetic subsets during training. For the scene-text subset, raw OCR outputs are directly available as a by-product of pseudo-label generation. For the synthetic subset, instead of running an additional OCR pipeline, we simulate OCR priors using an empirical noise model calibrated from the PPOCR-V5 error profile reported in the main paper. We do not inject OCR priors for document pages due to token-budget constraints.

Following the empirical breakdown reported in the main paper, we decompose OCR failures into three types: missing text instances, box localization errors, and intra-line transcript corruption. Let $R$, $P$, and $\mathrm{CER}$ denote the box recall, box precision, and character error rate of PPOCR-V5, respectively. Let $\hat{E}_{\mathrm{del}}$ and $\hat{E}_{\mathrm{ins}}$ denote the normalized deletion and insertion proportions among character-level OCR errors. We define three unnormalized weights:
\begin{equation}
\begin{aligned}
w_{\mathrm{del}} &= 1 - R \approx 0.3931, \\
w_{\mathrm{jit}} &= 1 - P \approx 0.1928, \\
w_{\mathrm{txt}} &= R \cdot \mathrm{CER} \cdot (\hat{E}_{\mathrm{del}} + \hat{E}_{\mathrm{ins}}) \approx 0.2381.
\end{aligned}
\end{equation}
After normalization, these yield the categorical sampling probabilities
\begin{equation}
\begin{aligned}
p_{\mathrm{del}} &\approx 0.477, \\
p_{\mathrm{jit}} &\approx 0.234, \\
p_{\mathrm{txt}} &\approx 0.289.
\end{aligned}
\end{equation}
These three modes correspond to line-level deletion, box jitter, and intra-line transcript corruption, respectively. We do not explicitly model substitution noise, since scene-text confusion statistics transfer poorly to the typography-rich synthetic split and substitution accounts for only a small fraction of character-level errors in our measurements.

The stochastic corruption function $\mathcal{F}$ applied to a text instance $(b_i,t_i)$ is defined as
\begin{equation}
\mathcal{F}(b_i, t_i) =
\begin{cases}
\emptyset, & p_{\mathrm{del}} \approx 0.477,\\
(\mathrm{Jitter}(b_i), t_i), & p_{\mathrm{jit}} \approx 0.234,\\
(b_i, \mathrm{Perturb}(t_i)), & p_{\mathrm{txt}} \approx 0.289.
\end{cases}
\end{equation}

Here, \texttt{Jitter} denotes coordinate perturbation of the bounding box. In our implementation, each box boundary is perturbed independently with magnitude proportional to the box size, using a jitter ratio uniformly sampled from $[0.12, 0.17]$, which yields an average IoU of approximately $0.76$ between the original and perturbed boxes. \texttt{Perturb} denotes character-level transcript corruption implemented as a stochastic combination of deletion and insertion. Concretely, for a transcript of length $n$, we first sample a total corruption ratio uniformly from $[0.2, 0.6]$, convert it into an error budget proportional to $n$, and then split this budget into deletion and insertion operations using fixed shares of $0.7256$ and $0.2744$, respectively.

The main paper defines a three-state prior schedule with $\gamma \in \{1.0, 0.5, 0.0\}$, corresponding to present, noisy, and absent priors. In practice, these three states are materialized offline during data construction and then mixed during training, rather than being sampled on the fly inside the training loop. Let $\mathcal{V}=\{(b_i,t_i)\}$ denote the ground-truth text instances of an image, and let $\mathcal{V}_{raw}$ denote raw OCR outputs when available. The injected prior set is defined as
\begin{equation}
\tilde{\mathcal{V}}_{\gamma} =
\begin{cases}
\mathcal{V}_{raw}\ \text{(scene)}\ \text{or}\ \mathcal{V}\ \text{(synthetic)}, & \gamma = 1.0, \\
\mathcal{V}_{keep}\cup \mathcal{F}(\mathcal{V}_{noise}), & \gamma = 0.5, \\
\emptyset, & \gamma = 0.0.
\end{cases}
\end{equation}
When $\gamma=0.5$, we split the available prior instances into an unchanged subset $\mathcal{V}_{keep}$ and a corrupted subset $\mathcal{V}_{noise}$, and apply $\mathcal{F}$ independently to each instance in $\mathcal{V}_{noise}$. For the synthetic subset, this corruption is applied to pseudo OCR priors constructed from the ground-truth instances. For the scene-text subset, corruption is applied to a selected subset of OCR-derived priors aligned to the ground truth, while unmatched raw OCR items are preserved.

%% file: Supplymentary_Material/sections/details_of_training.tex
\section{Training Data Details}
\input{Supplymentary_Material/tables/stage2}
\paragraph{Stage-1 training details.}
Stage~1 is trained on the proposed \textbf{TextAnchor-26M} dataset, which contains approximately 26.7M training instances in total. The corpus is composed of three major parts: scene text, synthetic text, and academic document pages. As summarized in Table~\ref{tab:stage1_data}, the dataset is dominated by scene-text and synthetic-text supervision, while academic document pages provide an additional source of dense-layout text. Stage~1 uses only these three sources.

\paragraph{Stage-2 training details.}
Across all sources, supervision is converted into a unified instance format containing the image, transcript or question--answer target, bounding box, binary mask, and source tag. Stage~2 is trained on a 23.3M corpus, which includes 22.3M large-scale open-source OCR-related data, a 1.0M resampled subset drawn from the first three TextAnchor-26M sources, a 10K VQA with causal mask subset. The resampled subset preserves grounding-oriented supervision from Stage~1, while the VQA with causal mask subset introduces semantic question-answering signals with explicit answer localization, complementing the more basic reading and grounding supervision. As summarized in Table~\ref{tab:stage2_ocr_data}, the OCR portion covers text recognition, charts/tables, documents, formula recognition, and general OCR-VQA tasks, providing substantially broader open-source coverage than Stage~1.

%% file: Supplymentary_Material/tables/stage2.tex
\begin{table*}[!t]
\centering
\small
\caption{Composition of the Stage-2 training corpus.}
\label{tab:stage2_ocr_data}
\begin{tabular}{|p{0.20\textwidth}|p{0.15\textwidth}|p{0.59\textwidth}|}
\hline
\textbf{Category} & \textbf{Number} & \textbf{Datasets} \\
\hline
Text Recognition &
3.2M &
CASIA-HWDB2\cite{liu2013online}, K12\_Printing\cite{wiedmann2025finevisionopendataneed}, ORAND-CAR-2014\cite{ORAND-CAR-2014}, rendered\_text\cite{rendered_text}, IAM\cite{marti2002iam}, IIIT5K\cite{IIIT5K}, Imgur5K\cite{Imgur5K}, VisualMRC\cite{VisualMRC2021}, SROIE\cite{huang2019icdar2019sroie}, multiple public CAPTCHA datasets, COCO-Text\cite{COCO-Text} \\
\hline
Charts / tables &
3.9M &
ChartQA\cite{masry2022chartqa}, Chart2Text\cite{chat2text}, CoSyn-400K\cite{Cosyn-400K}, HiTab\cite{cheng2022hitab}, InfographicVQA\cite{mathew2022infographicvqa}, LRV-Chart, SciTSR\cite{SciTSR}, SimChart9K\cite{SimChart9K}, TabMWP\cite{TabMWP}, VisText\cite{VisText}, UniChart\cite{UniChart}, SynthChartNet\cite{SynthFormulaNet1, SynthChartNet2}, ArxivQA\cite{ArxivQA}, DVQA\cite{DVQA}, FigureQA\cite{FigureQA}, FinQA\cite{FinQA}, FUNSD\cite{FUNSD}, MMC\cite{MMC}, PlotQA\cite{PlotQA}, RoBUT\cite{RoBUT}, TAT-QA\cite{zhu2021tat}, VQAonBD\cite{VQAonBD1,ST-VQA,VQAonBD3,VQAonBD4}, Block-Diagram-Datasets\cite{Block-Diagram-Datasets} \\
\hline
Documents &
5.7M &
DocVQA\cite{DocVQA}, DocReason25K\cite{DocReason25KandDocStruct4M}, Sujet\cite{Sujet}, PDF-VQA\cite{PDF-VQA}, TAT-DQA\cite{TAT-DQA}, DocDownstream\cite{DocDownstream}, DocStruct4M\cite{DocReason25KandDocStruct4M}, Docmatix\cite{mplug-docowl1.5}, UReader\cite{UReader} \\
\hline
Formula recognition &
2.4M &
UniMER-1M\cite{UniMER-1M}, SynthFormulaNet\cite{SynthFormulaNet1, SynthFormulaNet2}, HME100K\cite{HME100K}, latex\_handwritten\cite{wiedmann2025finevisionopendataneed}, LatexFormulas\cite{LatexFormulas}, Chrome\_Writing\cite{rendered_text} \\
\hline
General OCR VQA &
7.1M &
EST-VQA\cite{EST-VQA}, MTVQA\cite{MTVQA}, ST-VQA\cite{ST-VQA}, TextVQA\_train\cite{TextVQA}, WebSRC\cite{WebSRC}, LSVT\cite{LSVT}, ReCTS\cite{ReCTS}, RCTW\cite{rctw}, ArT\cite{ArT}, COCO-Text\cite{COCO-Text}, EATEN\cite{EATEN}, invoices\_receipts\cite{in-voices_receipts} \\
\hline
\hline
Stage-1 resampled subset &
1.0M &
A resampled subset drawn from the scene-text, synthetic-text, and academic-document components of TextAnchor-26M \\
\hline
VQA-with-causal-mask &
10K &
A high-quality VQA subset with explicit answer localization constructed from SVRD \\
\hline
\end{tabular}
\end{table*}

%% file: Supplymentary_Material/sections/Q-Mask_Visualization.tex
\section{Q-Mask Visualization}
We present qualitative visualizations to illustrate the effectiveness of Q-Mask across multiple vision-language tasks. Figure~\ref{fig:vis_t1_examples} shows Region-to-Text (R2T) results, where Q-Mask produces more accurate text recognition within specified regions compared to GPT-5.2, Qwen3-VL-235B-A22B-Instruct, and Qwen2.5-VL-3B(Baseline). 

\begin{figure*}[htbp]
\centering
\small
\setlength{\tabcolsep}{6pt}

\begin{tabular}{ccc}

\begin{minipage}[t]{0.30\linewidth}
\centering
\textbf{(a)} \\[2pt]
\includegraphics[width=\linewidth]{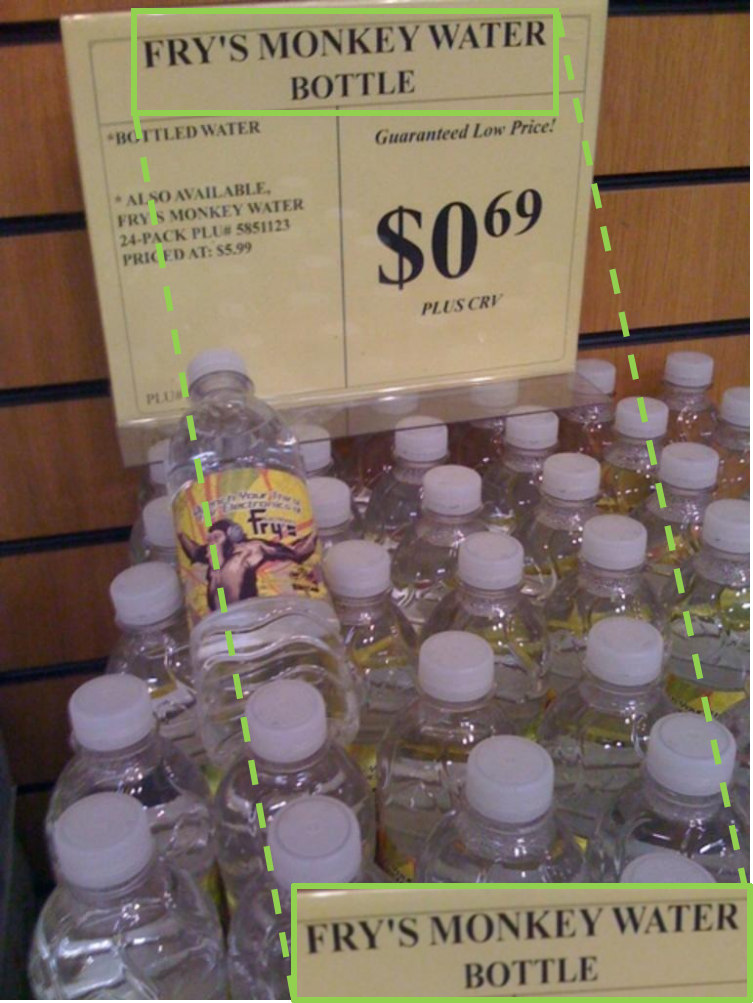}

\vspace{0.4em}

\raggedright \scriptsize
\textbf{Query:} What is .69? \\
\textcolor{gray}{GPT-5.2:} price \\
\textcolor{gray}{Qwen3-VL:} price \\
\textcolor{gray}{Baseline:} Price \\
\textcolor{blue}{\textbf{Ours:} Fry's monkey water bottle}

\end{minipage}

&
\begin{minipage}[t]{0.30\linewidth}
\centering
\textbf{(b)} \\[2pt]
\includegraphics[width=\linewidth]{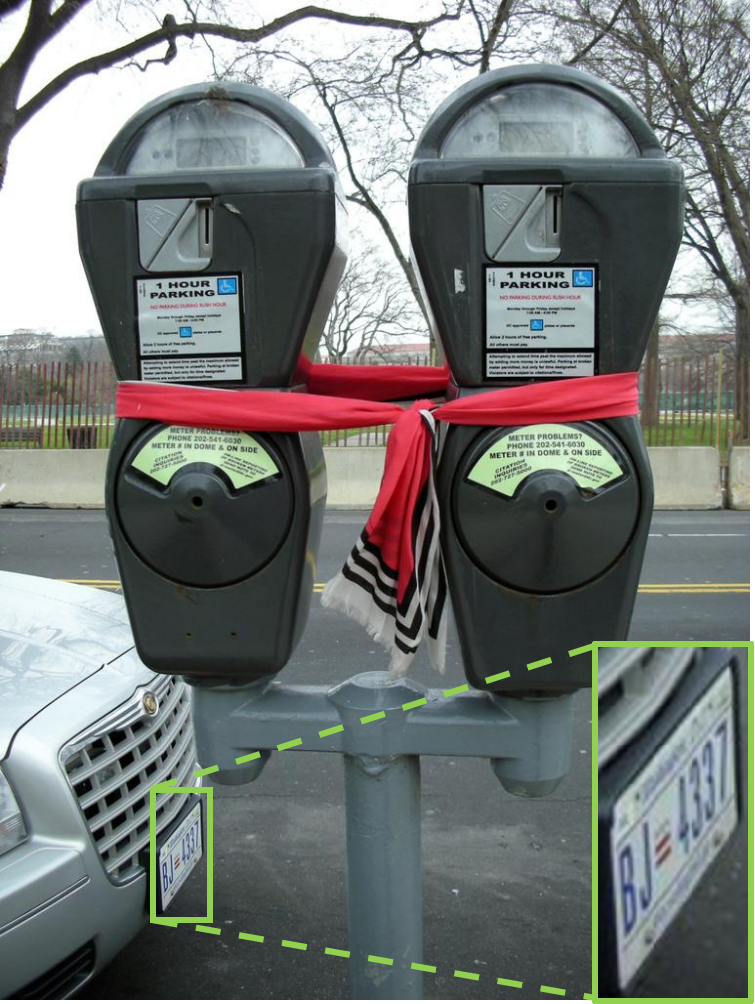}

\vspace{0.4em}

\raggedright \scriptsize
\textbf{Query:} Alphabet in car plate? \\
\textcolor{gray}{GPT-5.2:} B \\
\textcolor{gray}{Qwen3-VL:} B \\
\textcolor{gray}{Baseline:} B \\
\textcolor{blue}{\textbf{Ours:} Bj}

\end{minipage}

&
\begin{minipage}[t]{0.30\linewidth}
\centering
\textbf{(c)} \\[2pt]
\includegraphics[width=\linewidth]{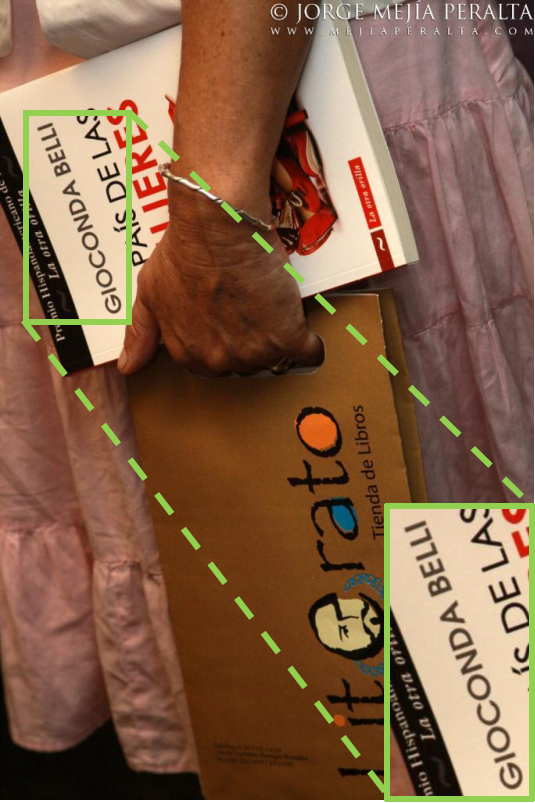}

\vspace{0.4em}

\raggedright \scriptsize
\textbf{Query:} Who is the author of the book? \\
\textcolor{gray}{GPT-5.2:} Gioconda Belli \\
\textcolor{gray}{Qwen3-VL:} Giocomo Bell... \\
\textcolor{gray}{Baseline:} Giacomo Giolitti \\
\textcolor{blue}{\textbf{Ours:} Gioconda Belli}

\end{minipage}

\\[1em]

\begin{minipage}[t]{0.30\linewidth}
\centering
\textbf{(d)} \\[2pt]
\includegraphics[width=\linewidth]{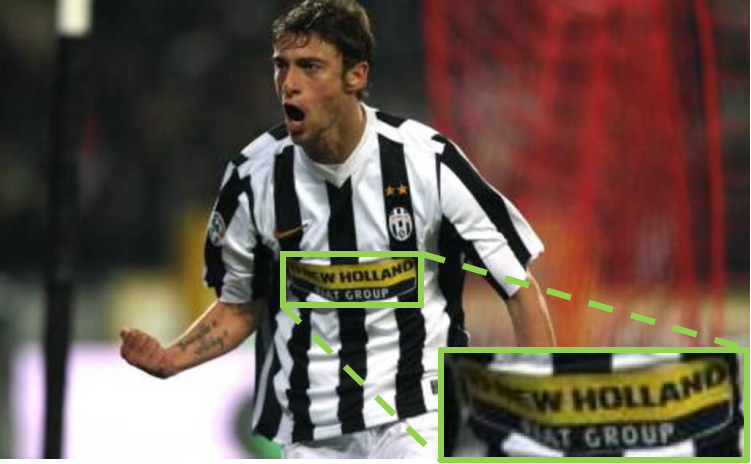}

\vspace{0.4em}

\raggedright \scriptsize
\textbf{Query:} What country does he play for? \\
\textcolor{gray}{GPT-5.2:} Italy \\
\textcolor{gray}{Qwen3-VL:} Italy \\
\textcolor{gray}{Baseline:} Italy \\
\textcolor{blue}{\textbf{Ours:} Holland}

\end{minipage}

&
\begin{minipage}[t]{0.30\linewidth}
\centering
\textbf{(e)} \\[2pt]
\includegraphics[width=\linewidth]{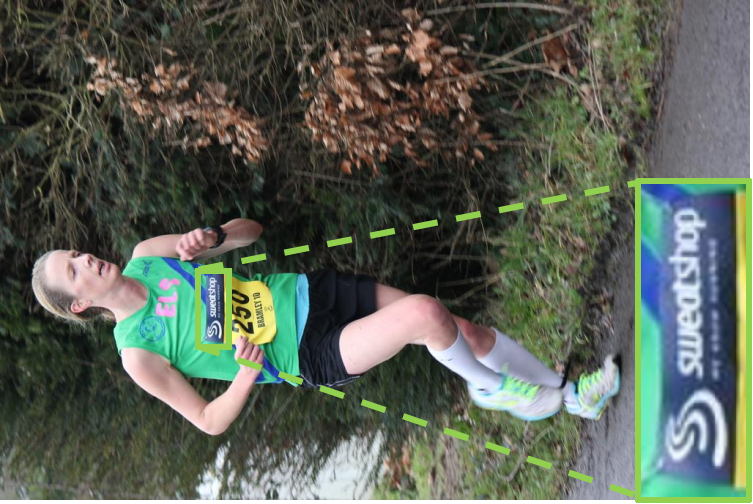}

\vspace{0.4em}

\raggedright \scriptsize
\textbf{Query:} Word on blue background? \\
\textcolor{gray}{GPT-5.2:} Sudtipk \\
\textcolor{gray}{Qwen3-VL:} superdrug \\
\textcolor{gray}{Baseline:} EL5 \\
\textcolor{blue}{\textbf{Ours:} Sweatshop}

\end{minipage}

&
\begin{minipage}[t]{0.30\linewidth}
\centering
\textbf{(f)} \\[2pt]
\includegraphics[width=\linewidth]{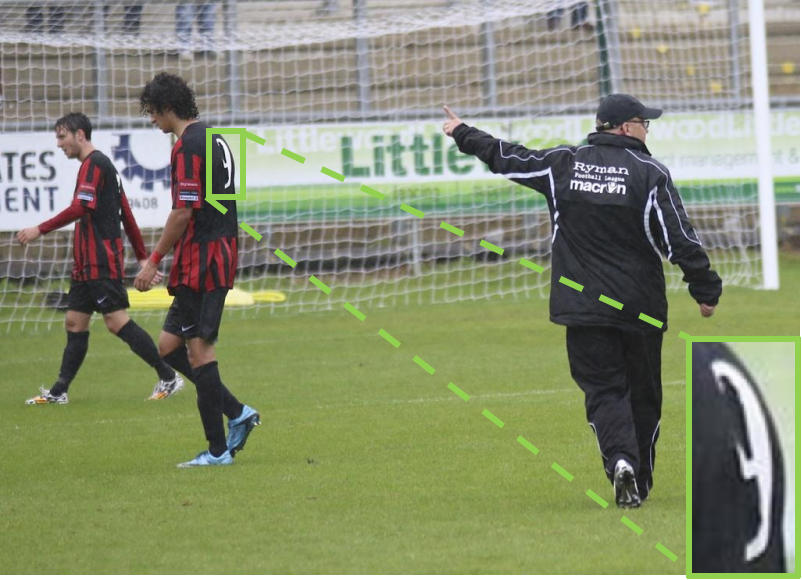}

\vspace{0.4em}

\raggedright \scriptsize
\textbf{Query:} Number of player on right? \\
\textcolor{gray}{GPT-5.2:} 3 \\
\textcolor{gray}{Qwen3-VL:} 9 \\
\textcolor{gray}{Baseline:} 3 \\
\textcolor{blue}{\textbf{Ours:} 9}

\end{minipage}

\end{tabular}

\caption{
\textbf{Qualitative comparison on diverse VQA tasks.}
We evaluate Q-Mask-3B(Ours) against GPT-5.2, Qwen3-VL-235B-A22B-Instruct(Qwen3-VL), and Qwen2.5-VL-3B(baseline).
Labels (a)--(f) correspond to different test cases.
}
\label{fig:main_comparison}

\end{figure*}

\begin{figure*}[htbp]
\centering
\small
\setlength{\tabcolsep}{3pt}

\begin{tabular}{c c c}

\textbf{(a) Input image} & \textbf{(b) Heatmap with Q-Mask} & \textbf{(c) Label Mask} \\

\includegraphics[width=0.29\linewidth]{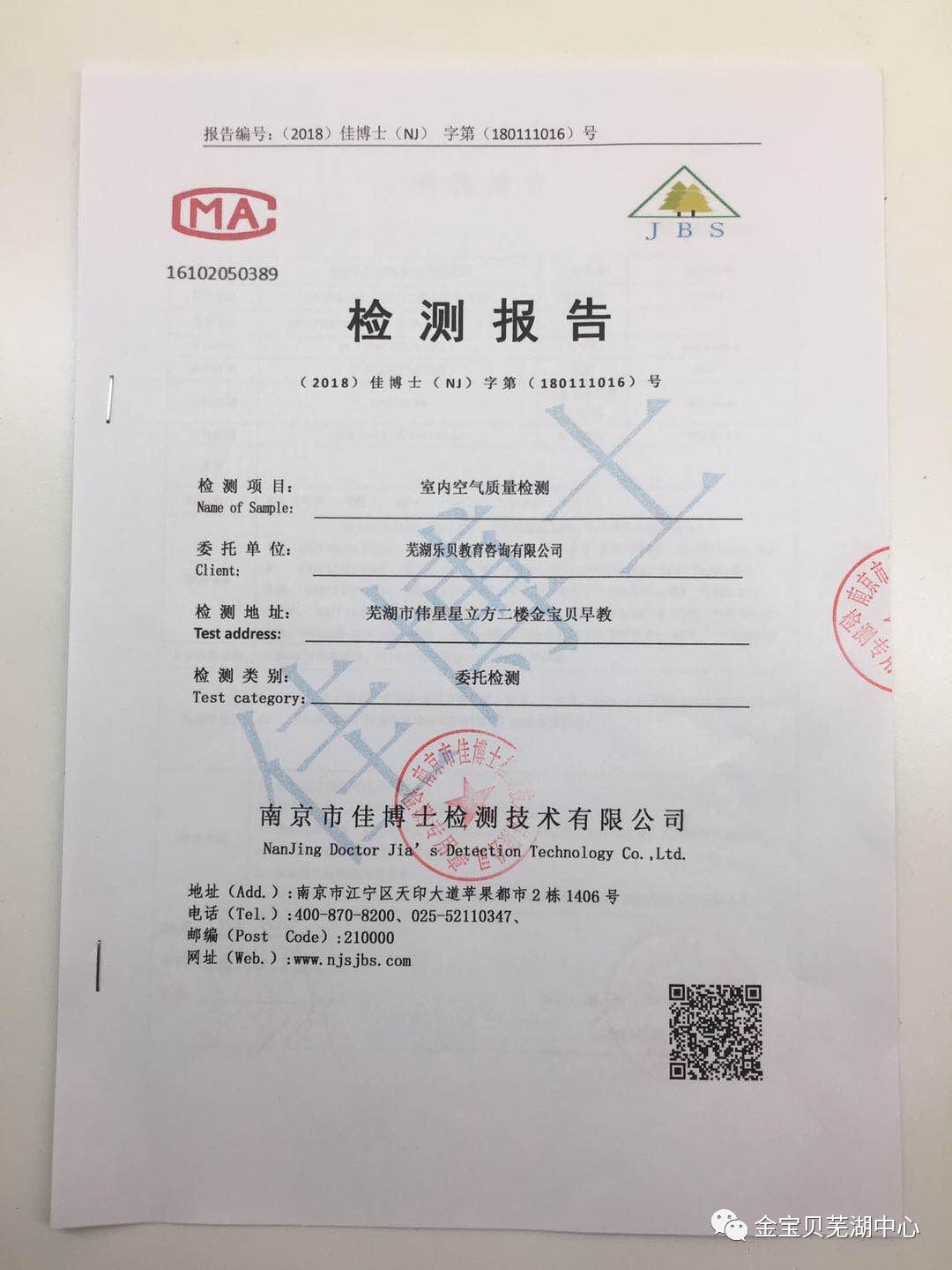} &
\includegraphics[width=0.29\linewidth]{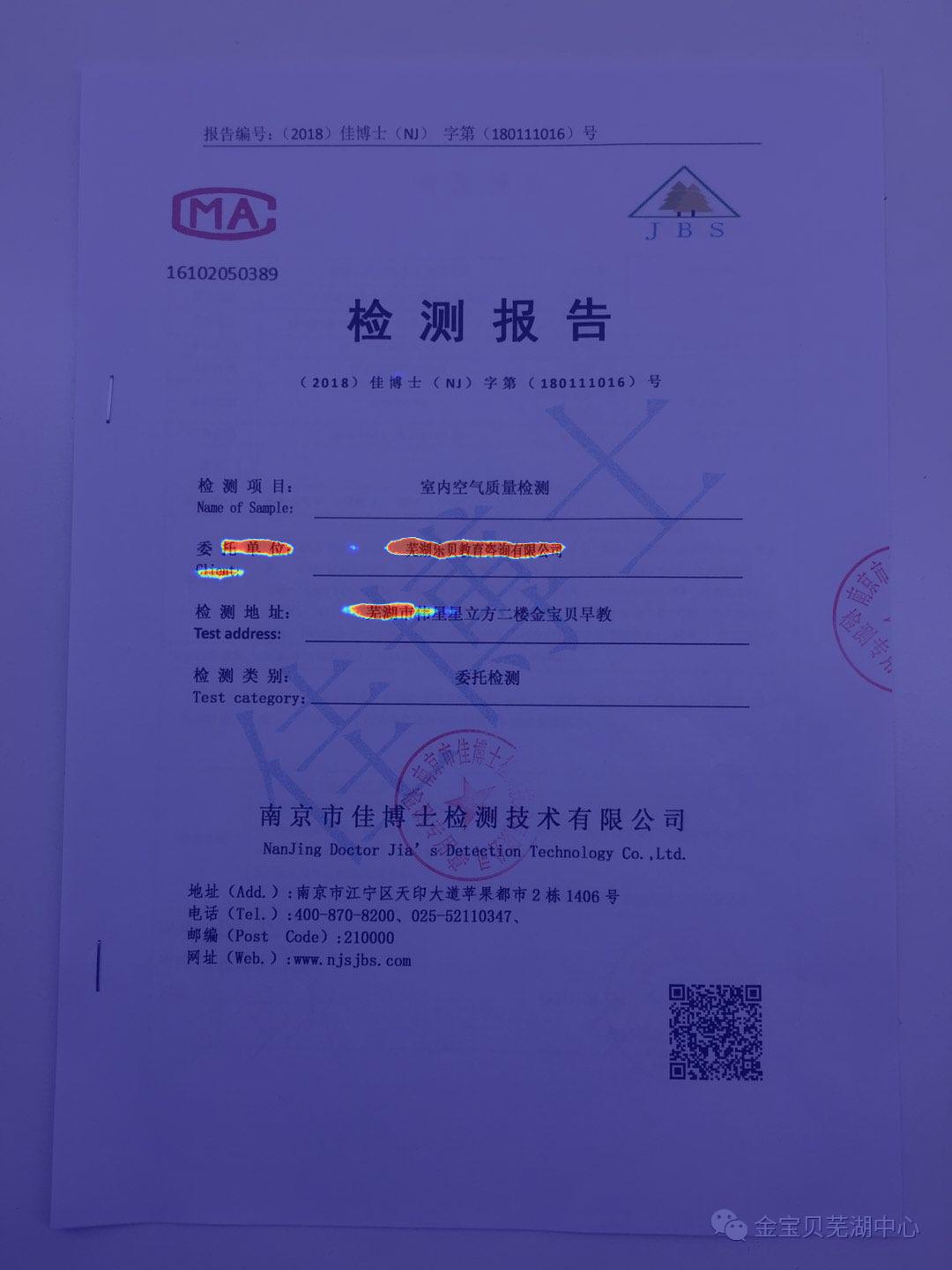} &
\includegraphics[width=0.29\linewidth]{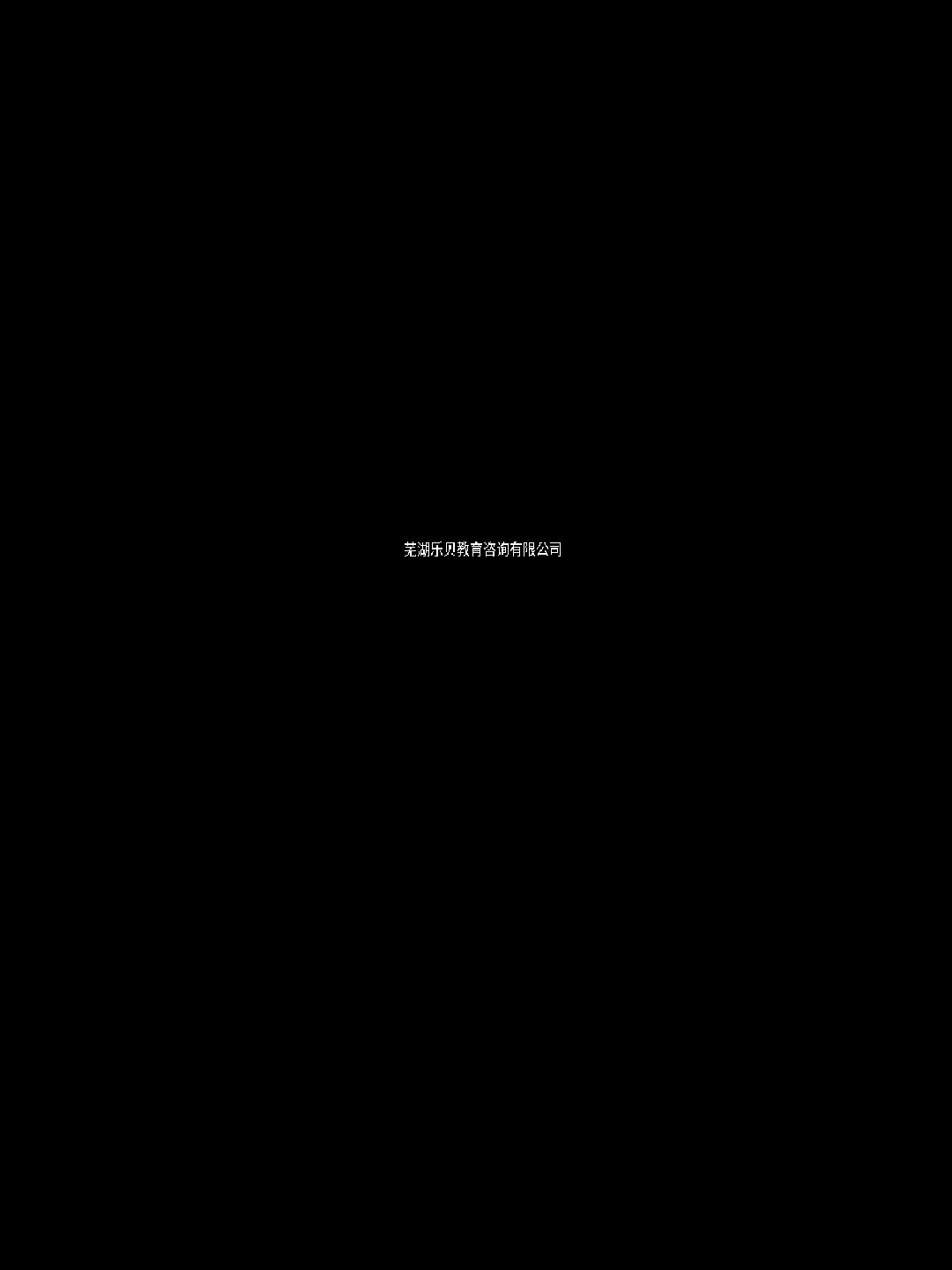} \\

\multicolumn{3}{l}{
\begin{minipage}{0.8\linewidth}
\footnotesize
\raggedright
\begin{CJK}{UTF8}{gbsn}
\textbf{Q:} Who is the client for this test? \textbf{A:} 芜湖乐贝教育咨询有限公司
\end{CJK}
\end{minipage}
} \\[2pt]

\includegraphics[width=0.29\linewidth]{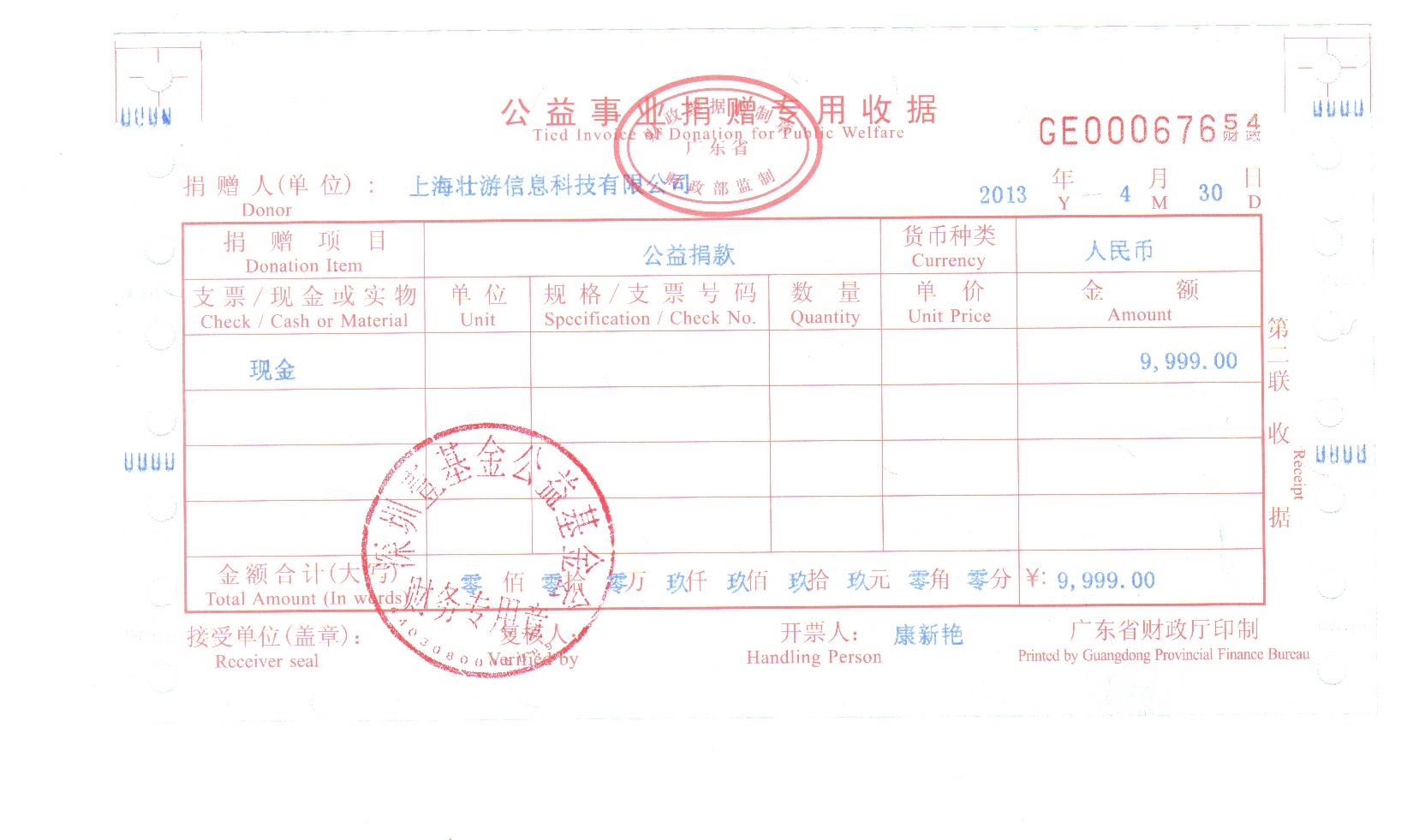} &
\includegraphics[width=0.29\linewidth]{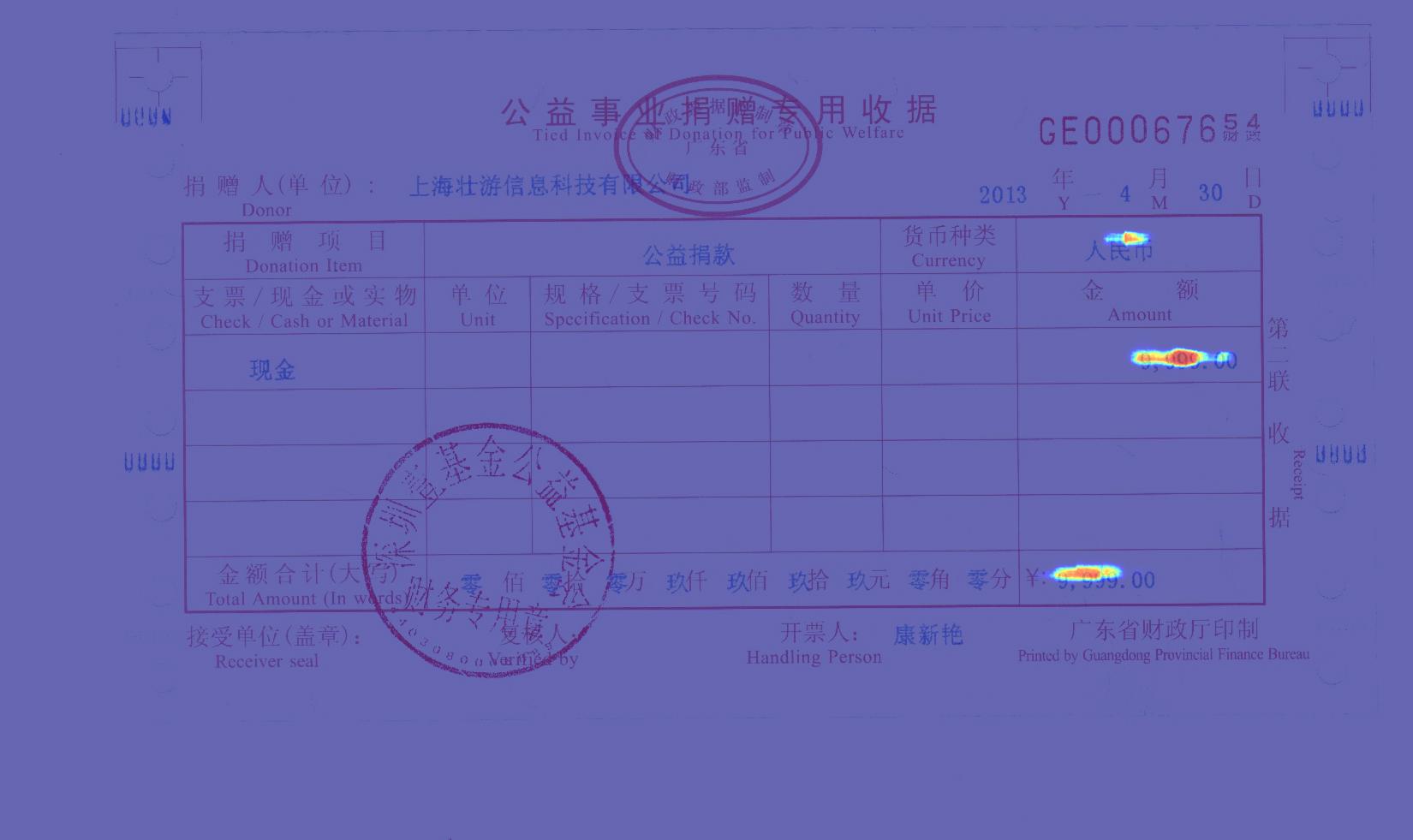} &
\includegraphics[width=0.29\linewidth]{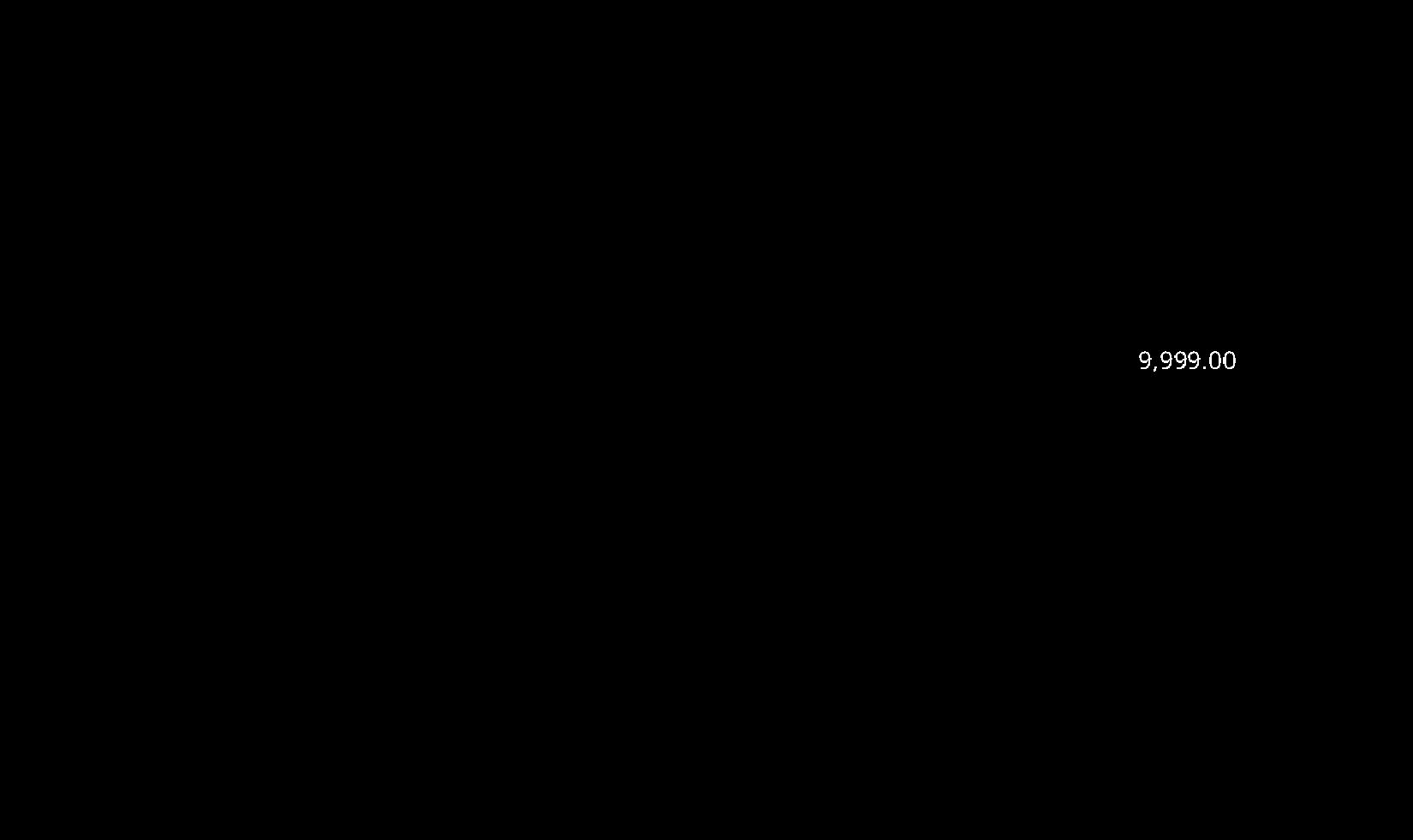} \\

\multicolumn{3}{l}{
\footnotesize
\raggedright
\textbf{Q:} What is the donation amount? \textbf{A:} 9,999.00
} \\[2pt]

\includegraphics[width=0.29\linewidth]{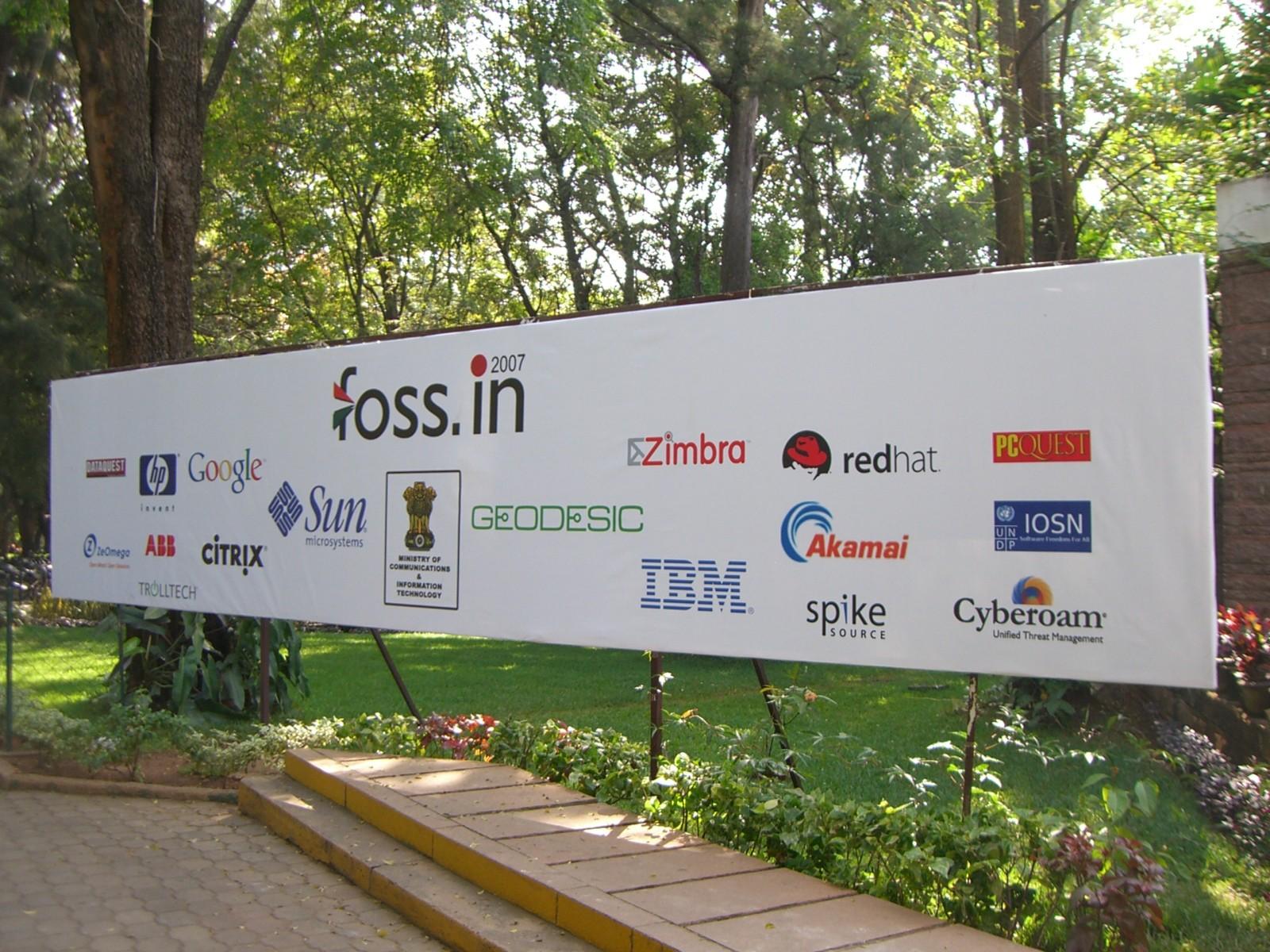} &
\includegraphics[width=0.29\linewidth]{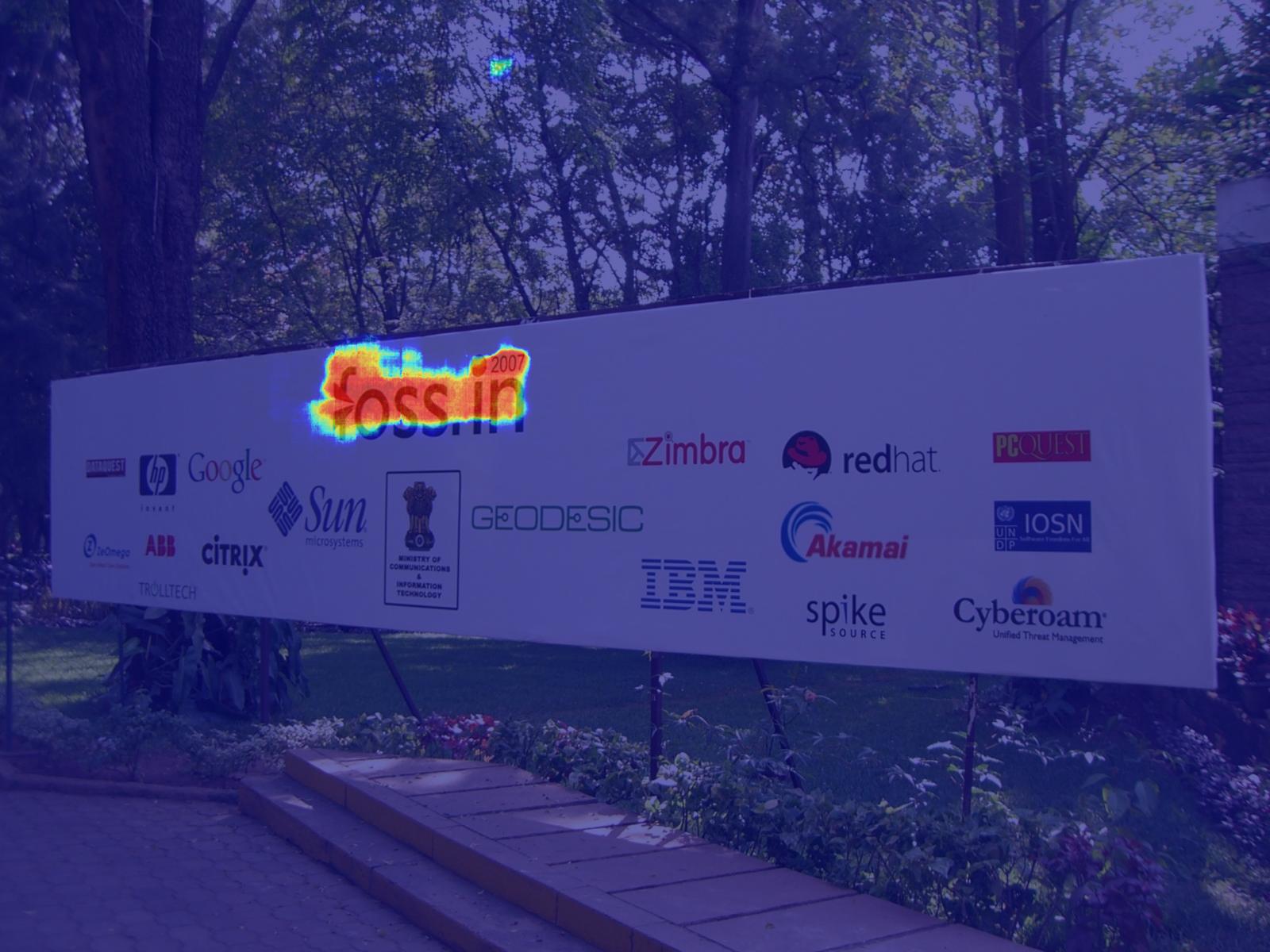} &
\includegraphics[width=0.29\linewidth]{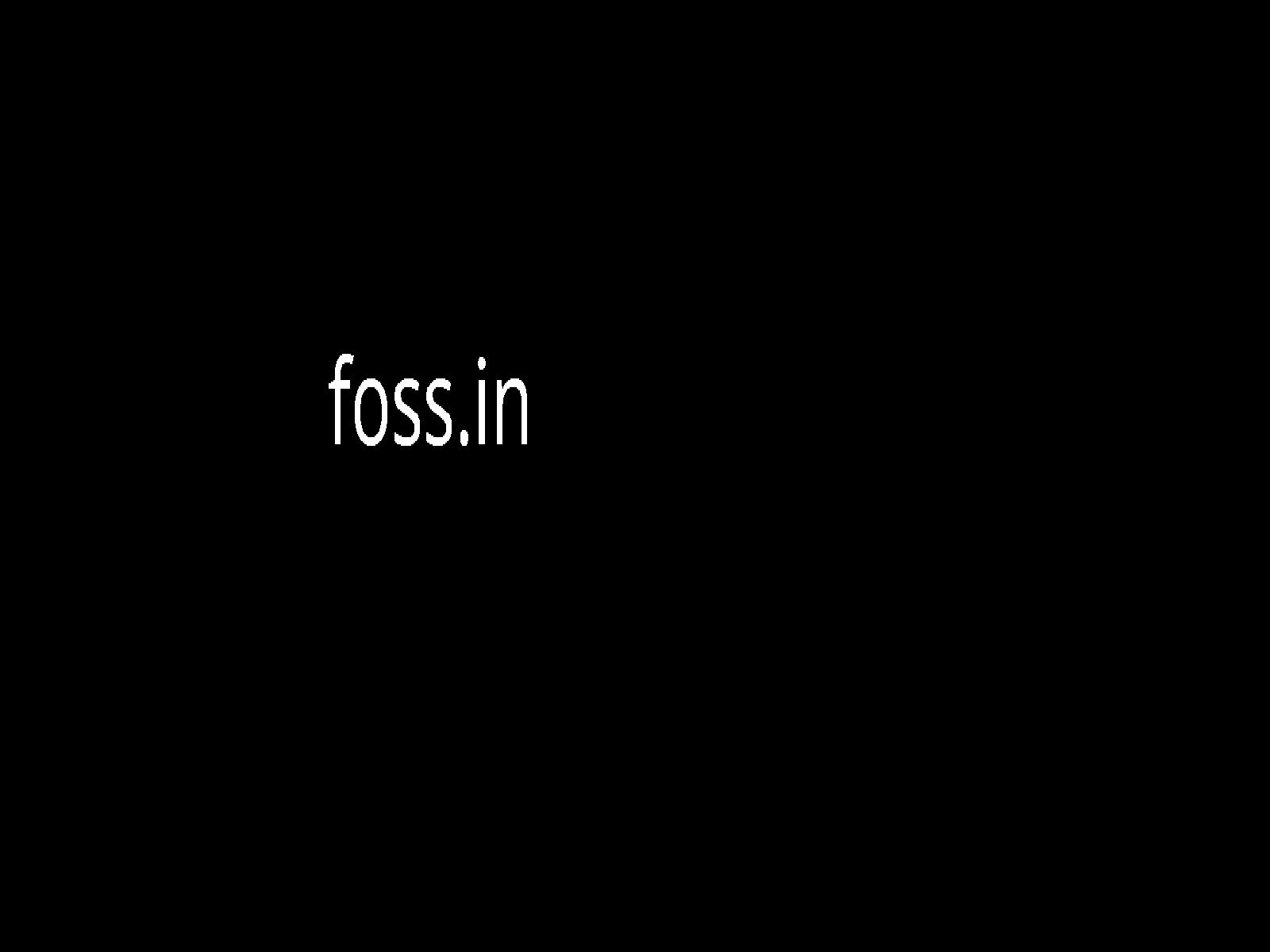} \\

\multicolumn{3}{l}{
\footnotesize
\raggedright
\textbf{Q:} What is the main event or organization name displayed on the banner? \textbf{A:} foss.in
} \\[2pt]

\includegraphics[width=0.29\linewidth]{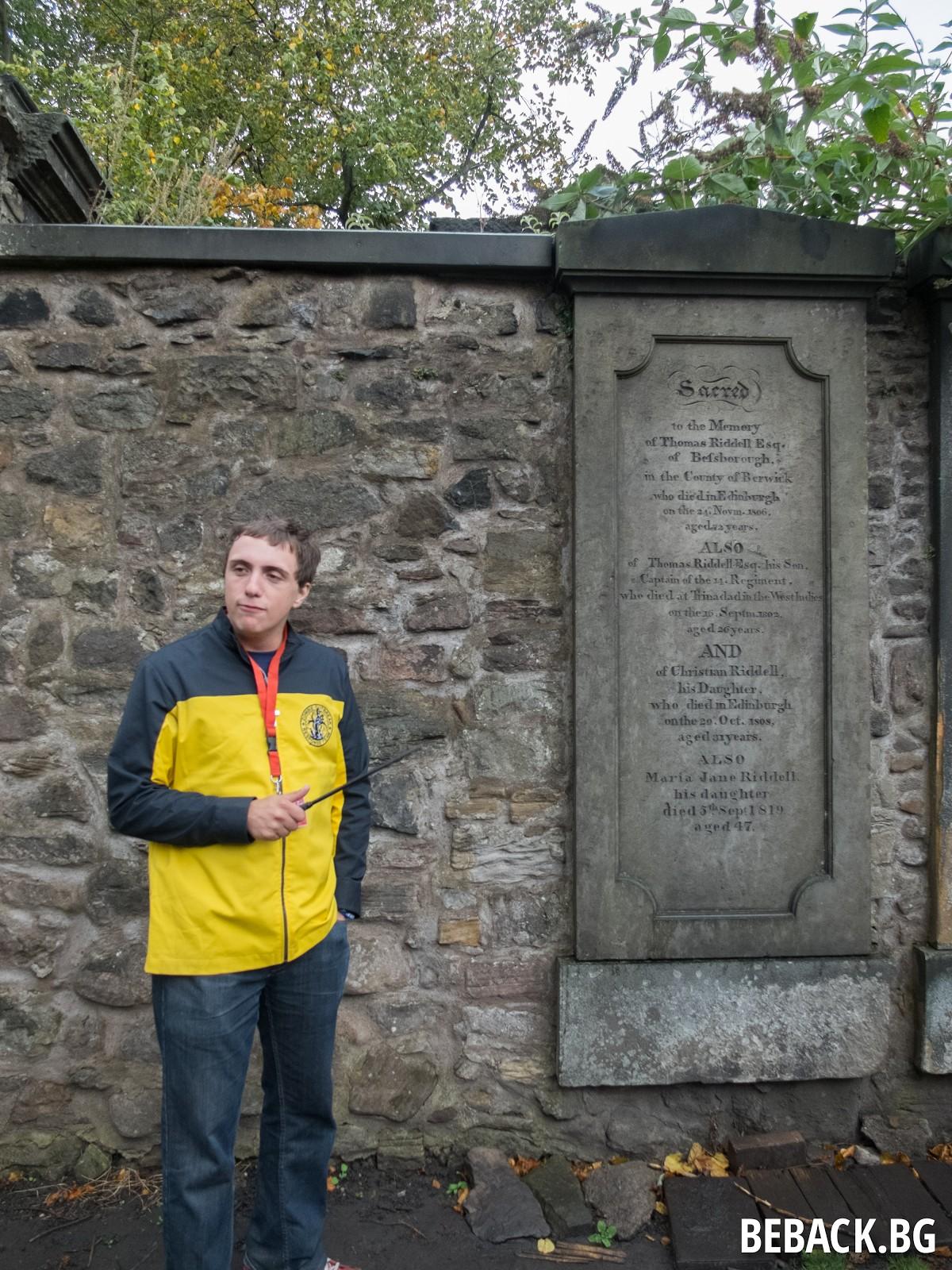} &
\includegraphics[width=0.29\linewidth]{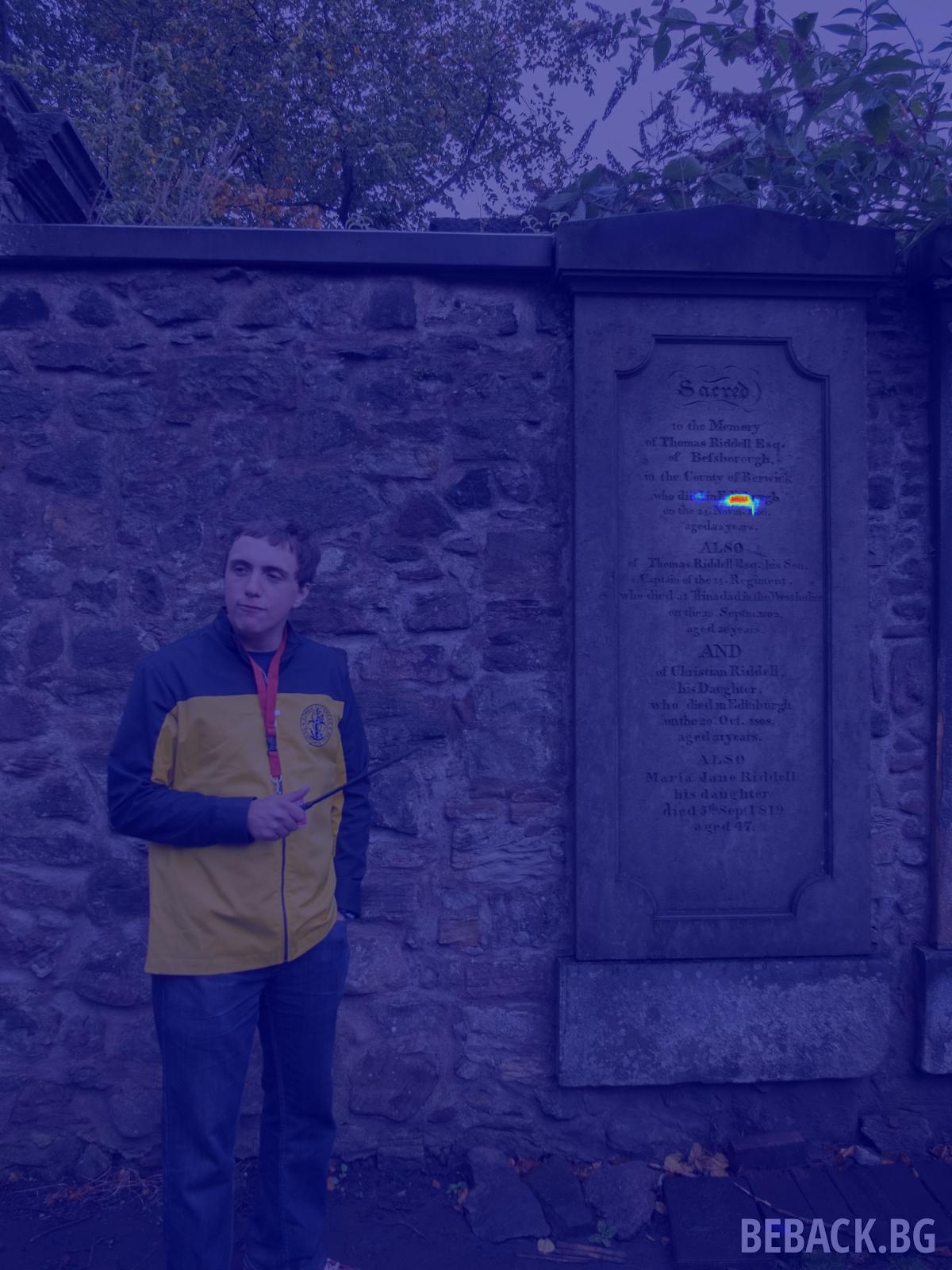} &
\includegraphics[width=0.29\linewidth]{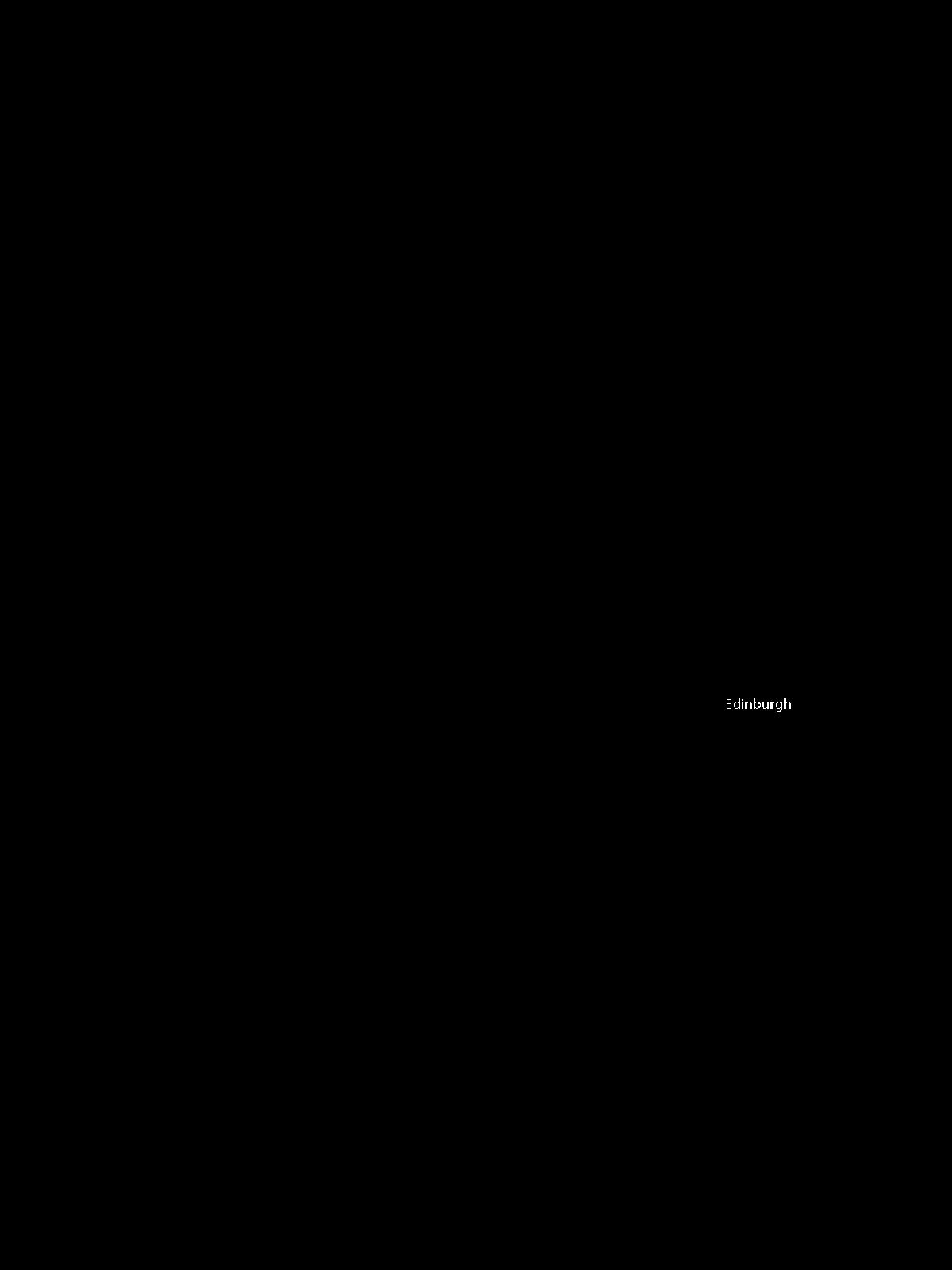} \\

\multicolumn{3}{l}{
\footnotesize
\raggedright
\textbf{Q:} Where did Thomas Riddell Esq. die? \textbf{A:} Edinburgh
} \\

\end{tabular}

\caption{
Visualization of Q-Mask-3B on representative VQA examples. For each example, we show (a) Input image, (b) Heatmap with Q-Mask-3B, (c) Label Mask.
}
\label{fig:vis_mask_examples}
\end{figure*}

Figure~\ref{fig:vis_t2_examples} presents Text-to-Region (T2R) grounding results, demonstrating more precise localization of query-relevant text. 
\input{Supplymentary_Material/tables/t2r}
\input{Supplymentary_Material/tables/r2t}
Figure~\ref{fig:main_comparison} further compares model outputs on diverse VQA tasks. Finally, Figure~\ref{fig:vis_mask_examples} visualizes Q-Mask heatmaps, showing that the model attends to query-relevant textual regions.

%% file: Supplymentary_Material/tables/t2r.tex
\begin{figure*}[htbp]
\centering
\small
\setlength{\tabcolsep}{2pt}
\renewcommand{\arraystretch}{0.95}

\begin{tabular}{cccc}

\textbf{(a) GPT-5.2} &
\textbf{(b) Qwen3-VL} &
\textbf{(c) Baseline} &
\textbf{(d) Ours} \\[1pt]

\includegraphics[width=0.235\linewidth]{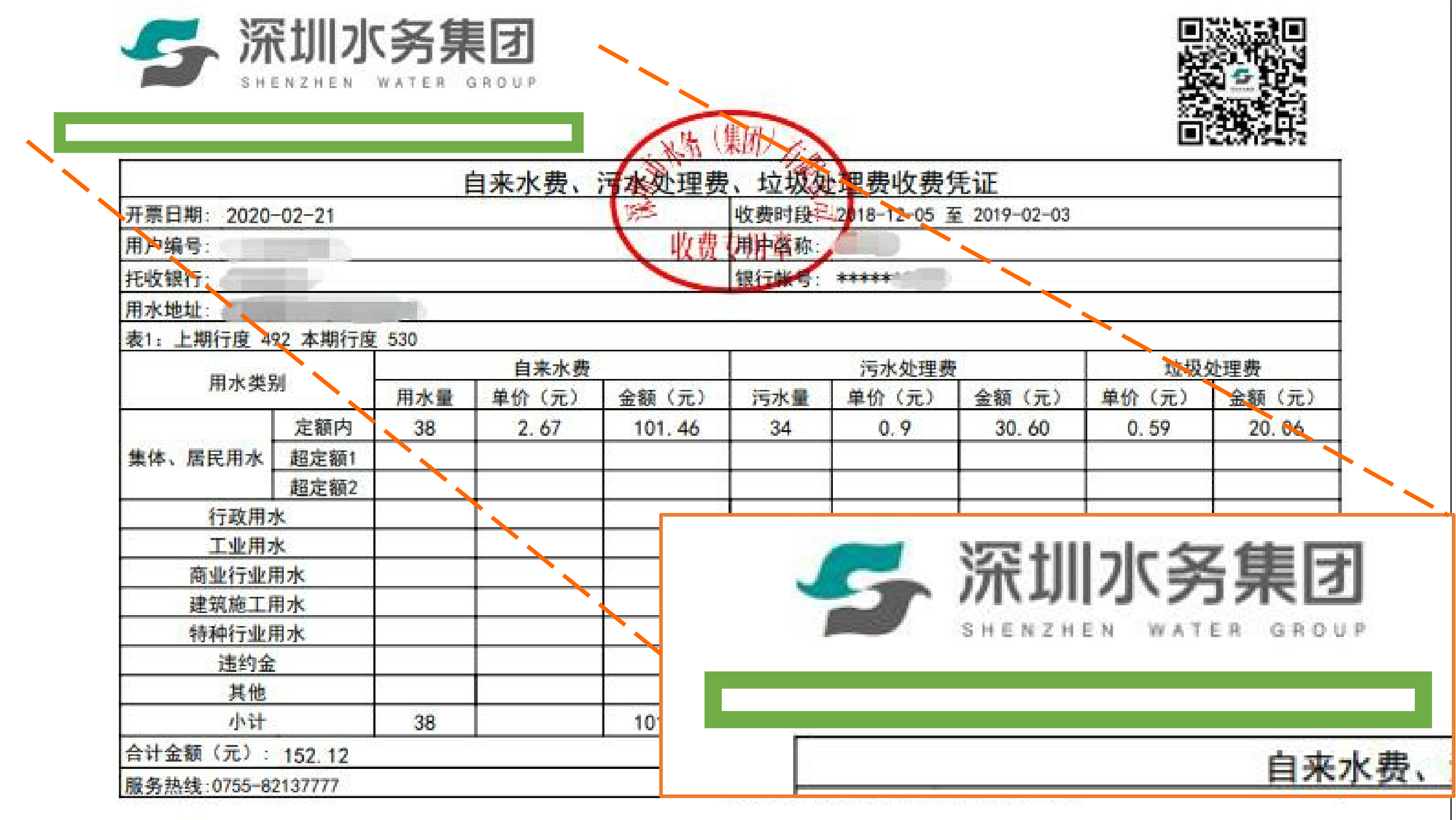} &
\includegraphics[width=0.235\linewidth]{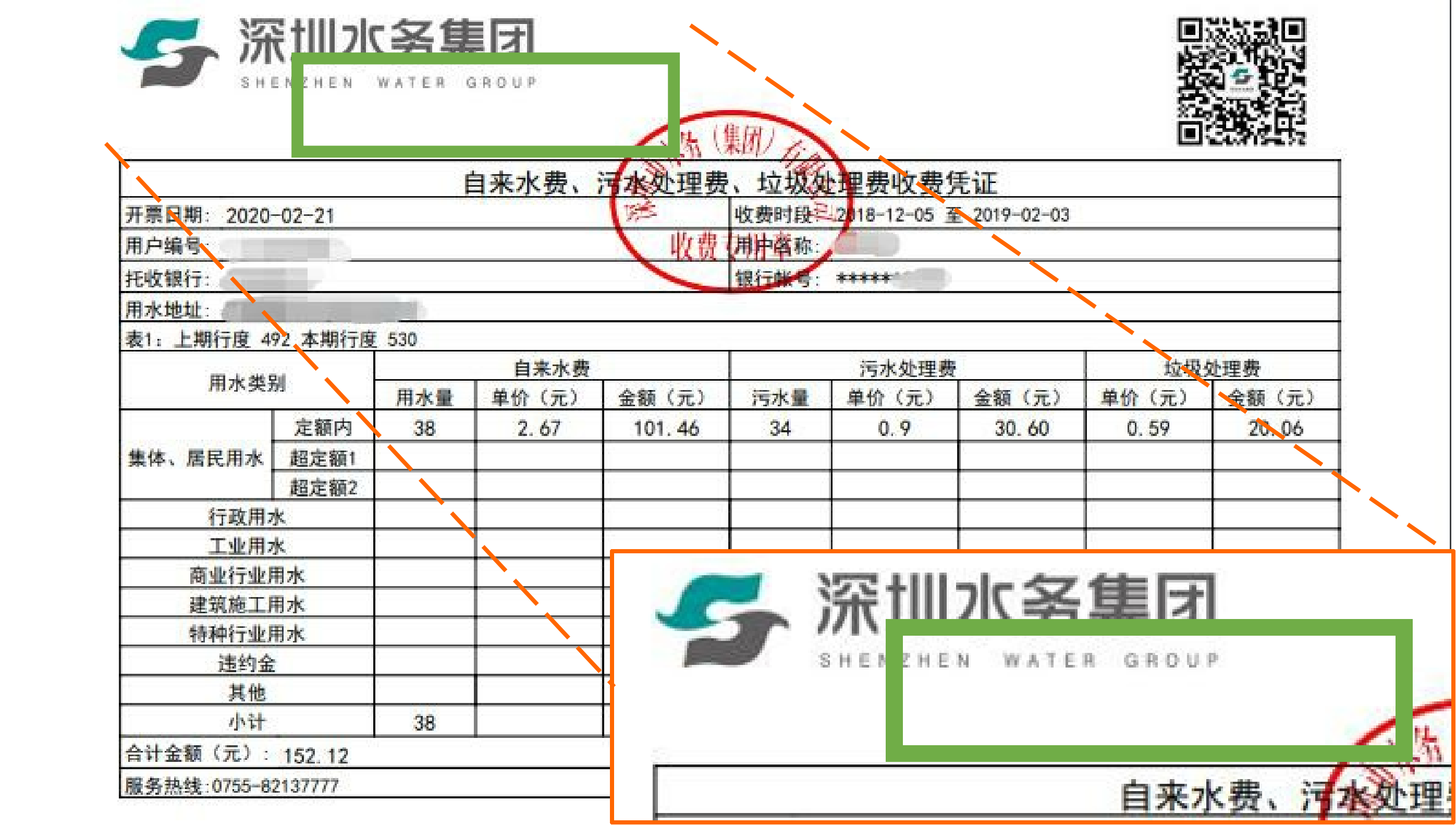} &
\includegraphics[width=0.235\linewidth]{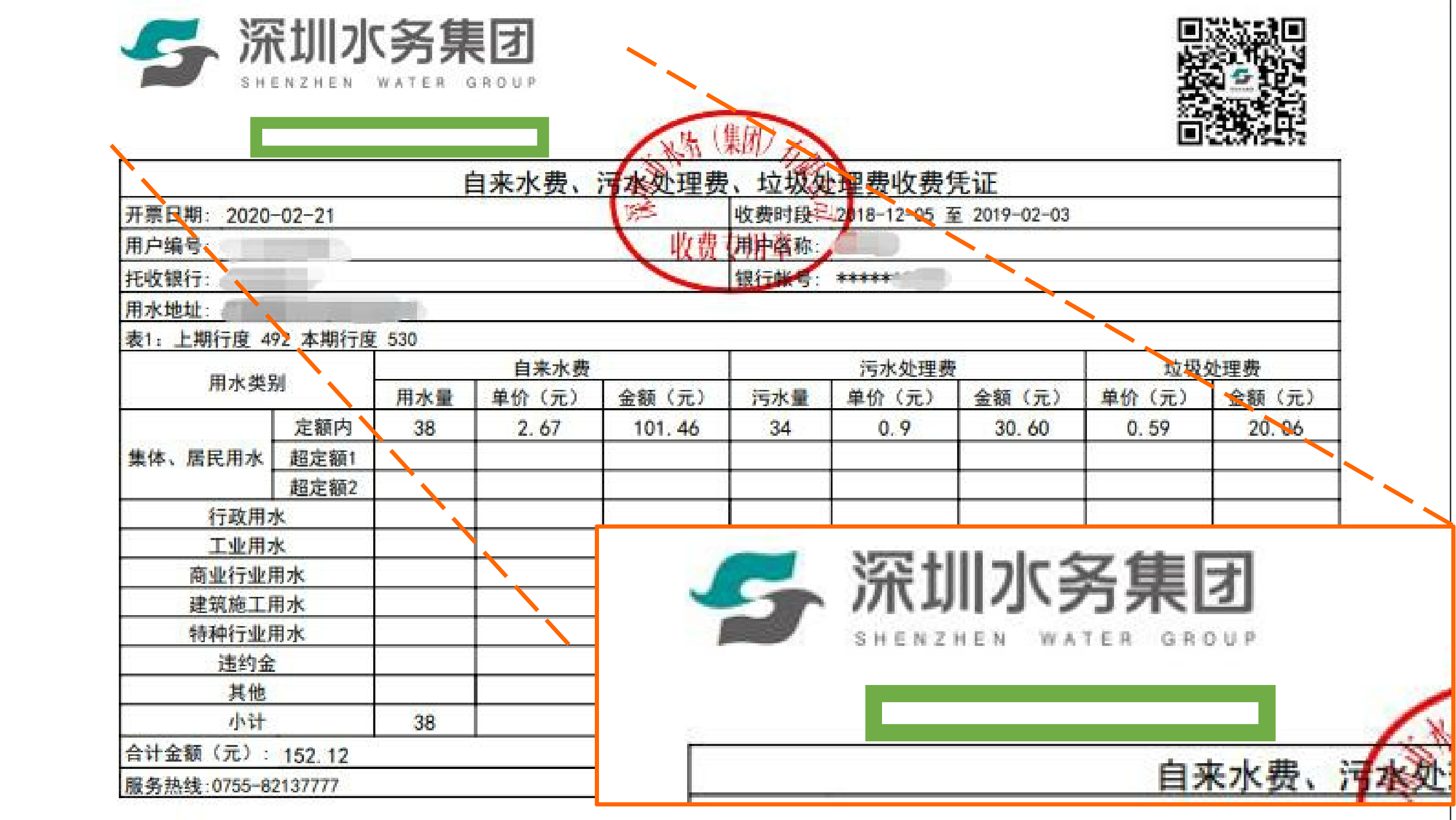} &
\includegraphics[width=0.235\linewidth]{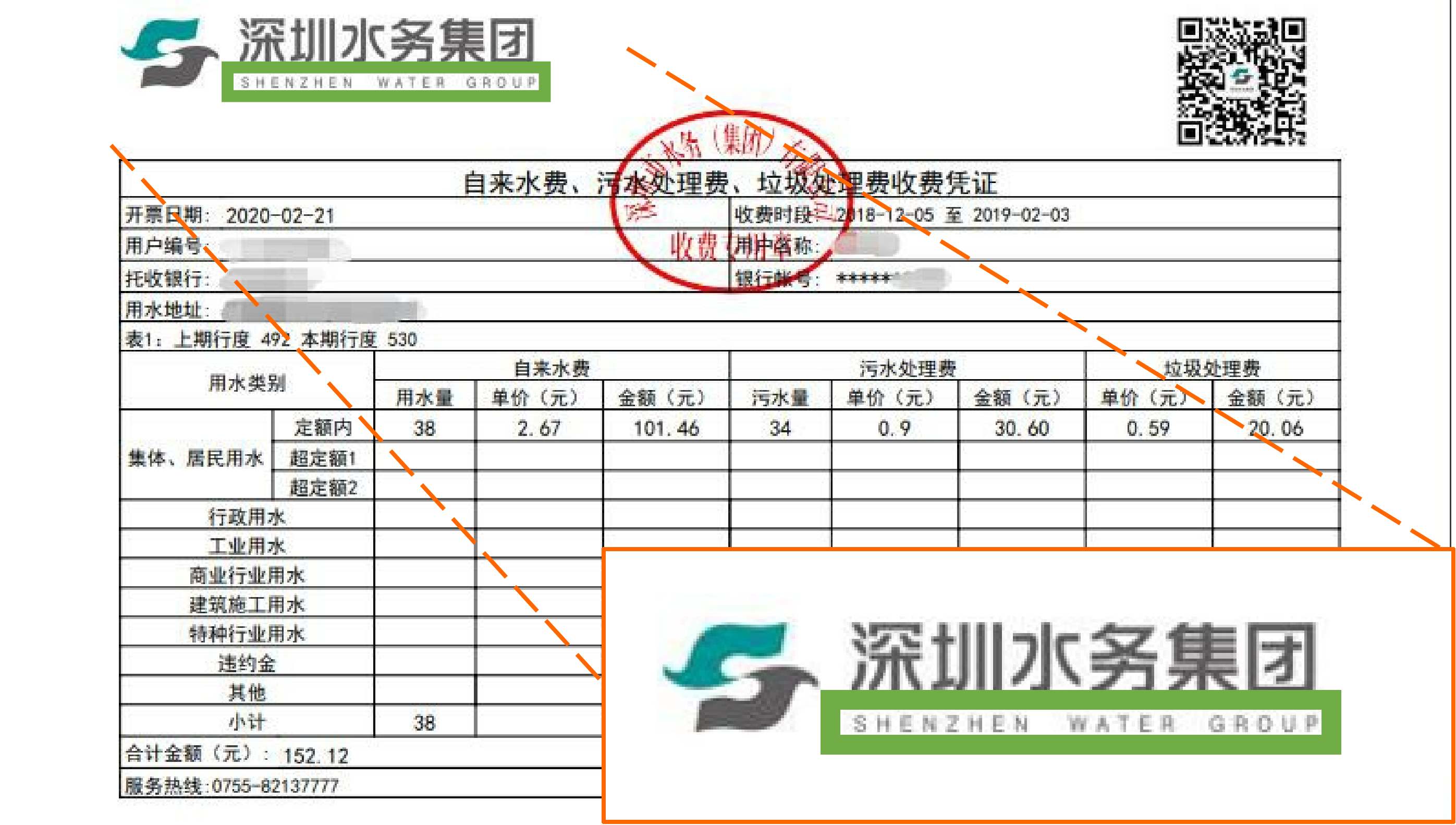} \\

\multicolumn{4}{l}{\scriptsize\textbf{Query:} SHENZHEN WATER GROUP} \\[0.35em]

\includegraphics[width=0.235\linewidth]{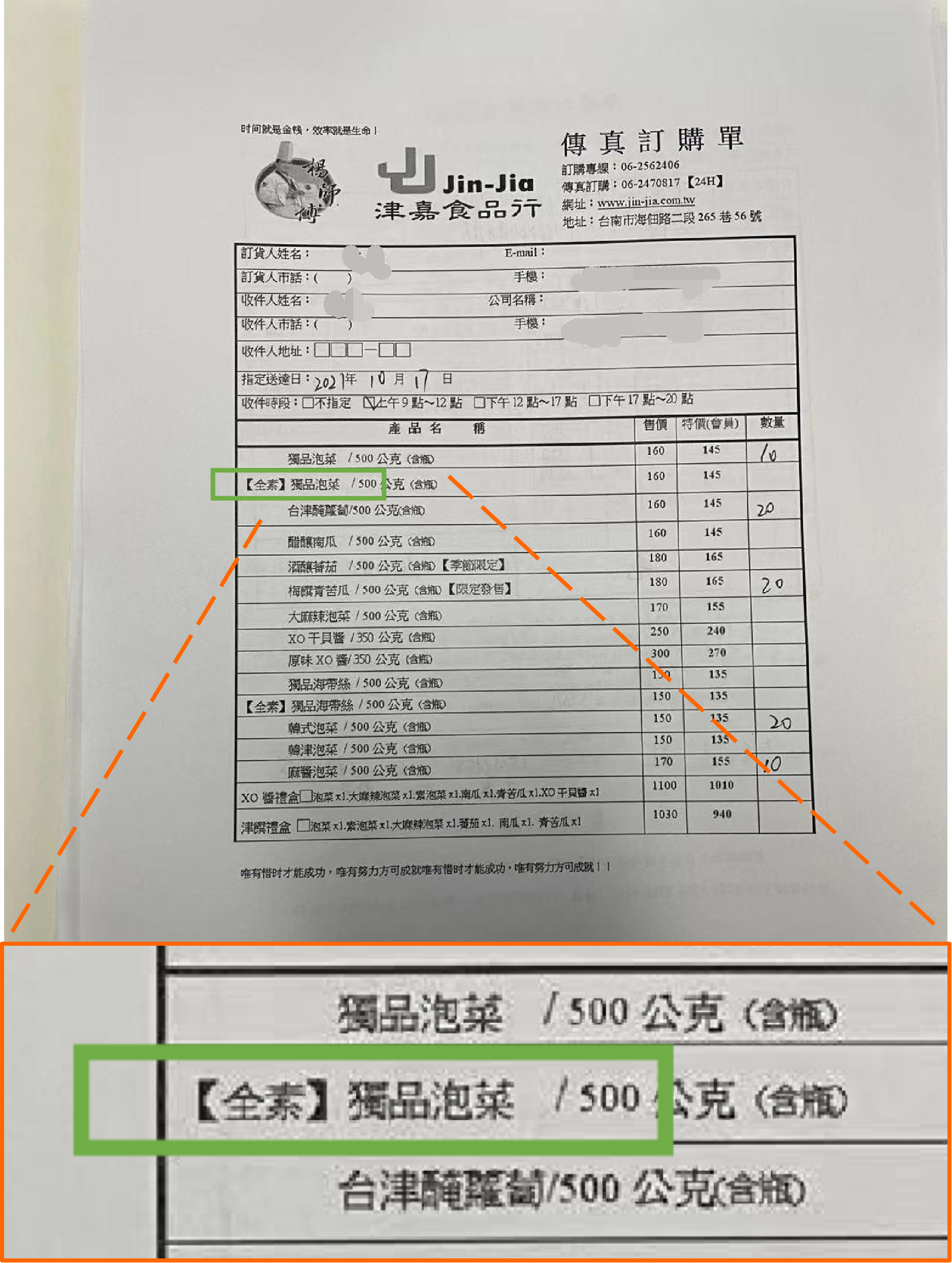} &
\includegraphics[width=0.235\linewidth]{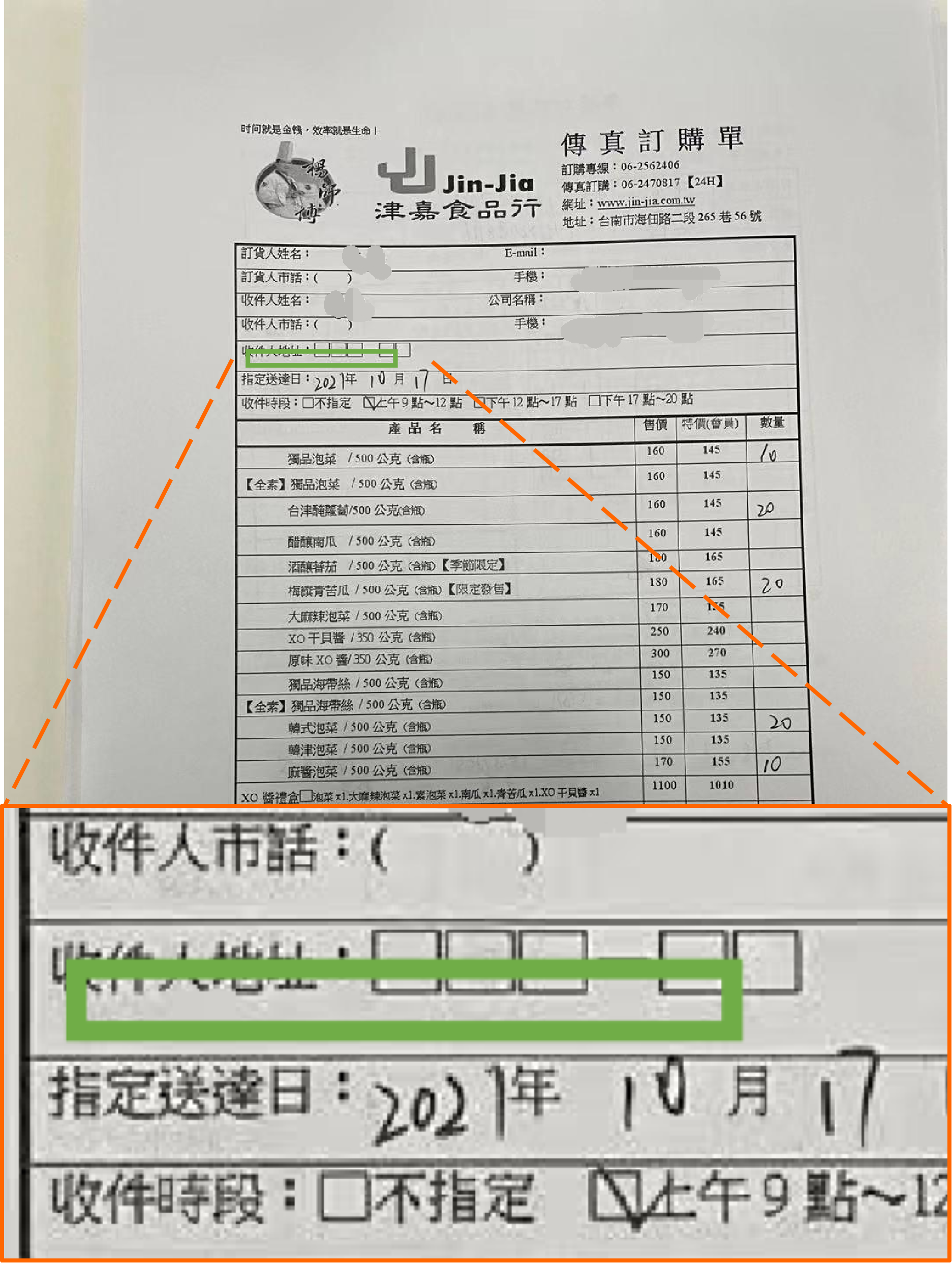} &
\includegraphics[width=0.235\linewidth]{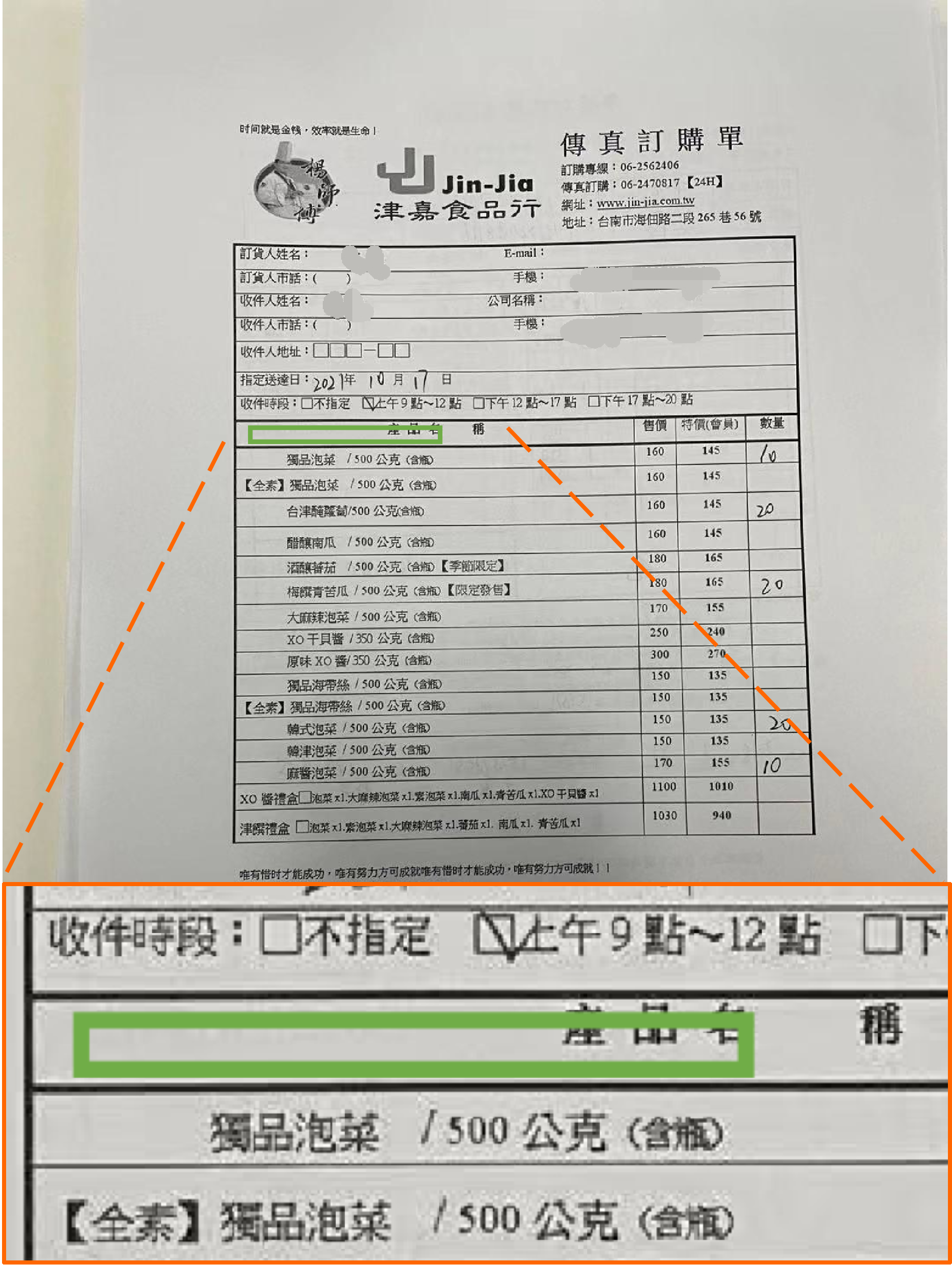} &
\includegraphics[width=0.235\linewidth]{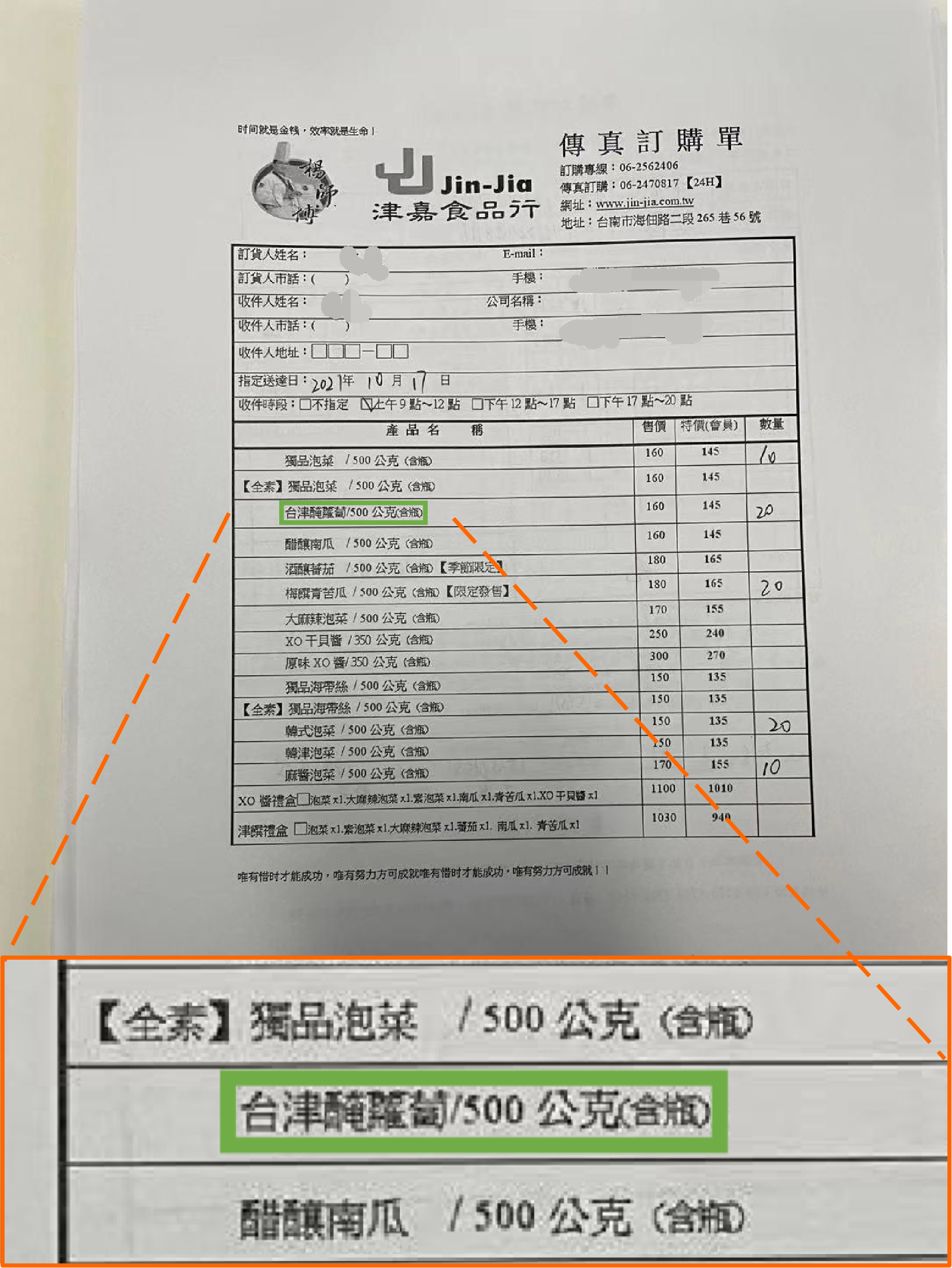} \\

\multicolumn{4}{l}{\scriptsize
\begin{CJK}{UTF8}{bsmi}
\textbf{Query:} 台津醃籮萄/500公克（含瓶）
\end{CJK}} \\[0.35em]

\includegraphics[width=0.235\linewidth]{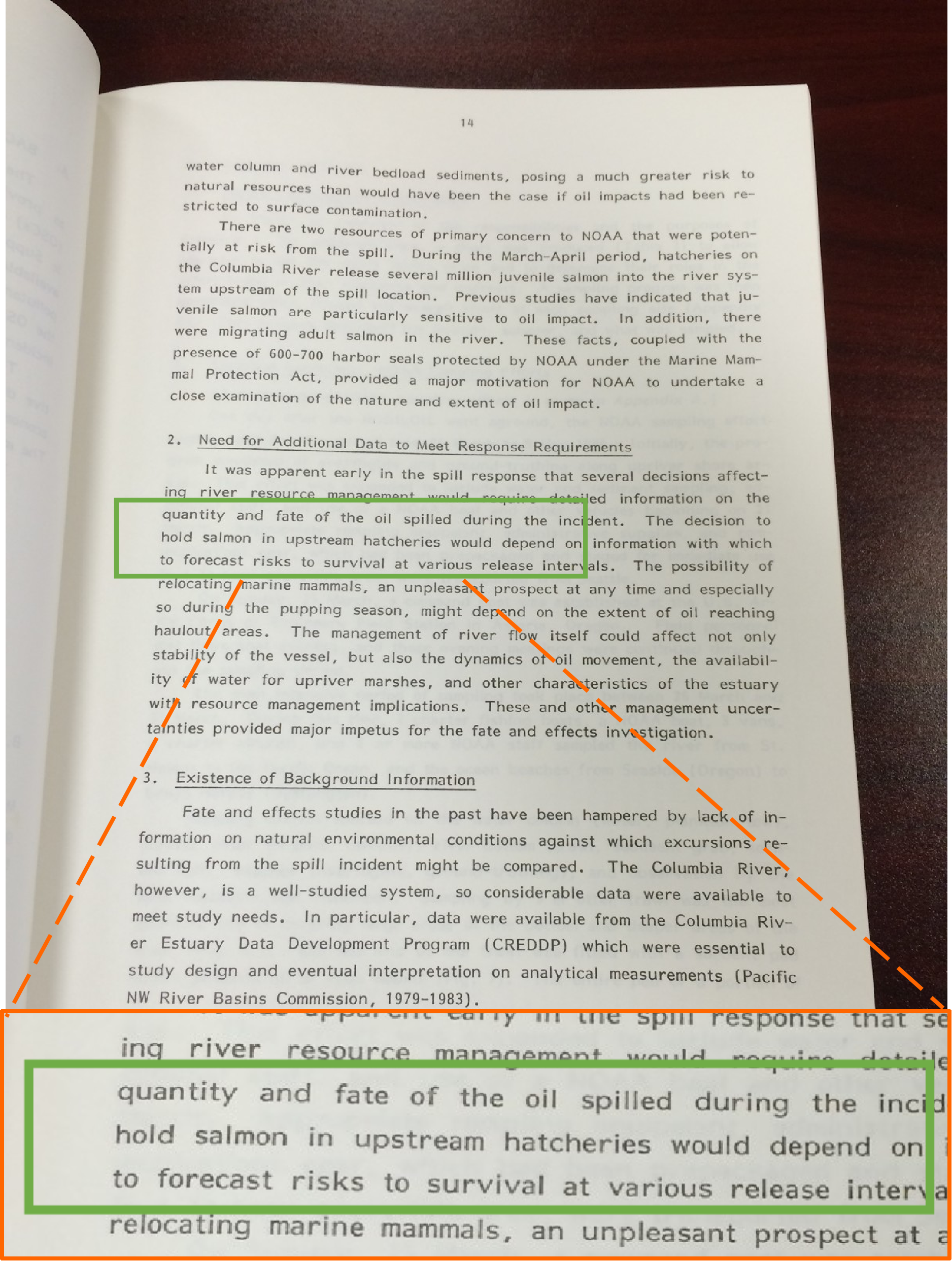} &
\includegraphics[width=0.235\linewidth]{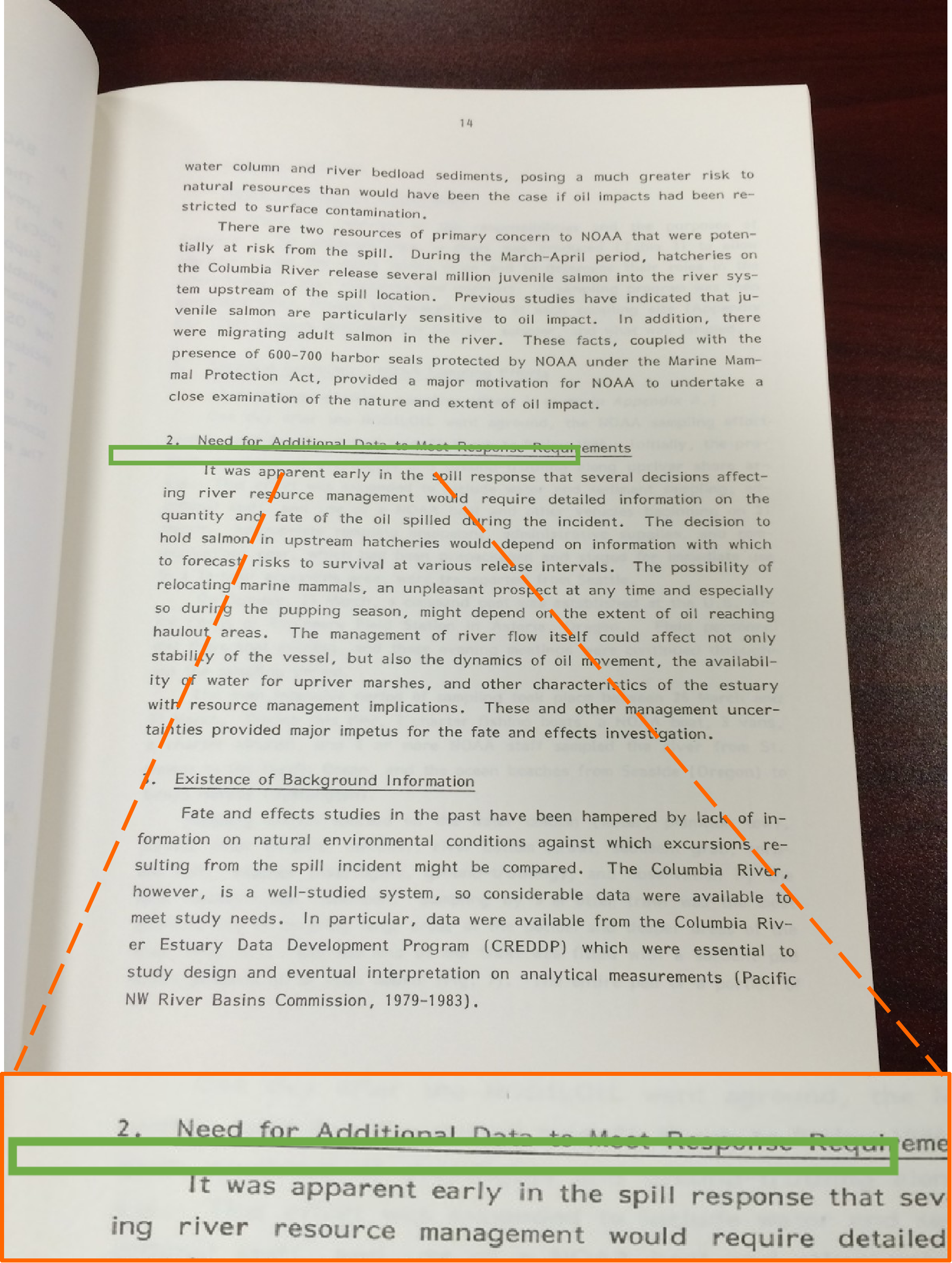} &
\includegraphics[width=0.235\linewidth]{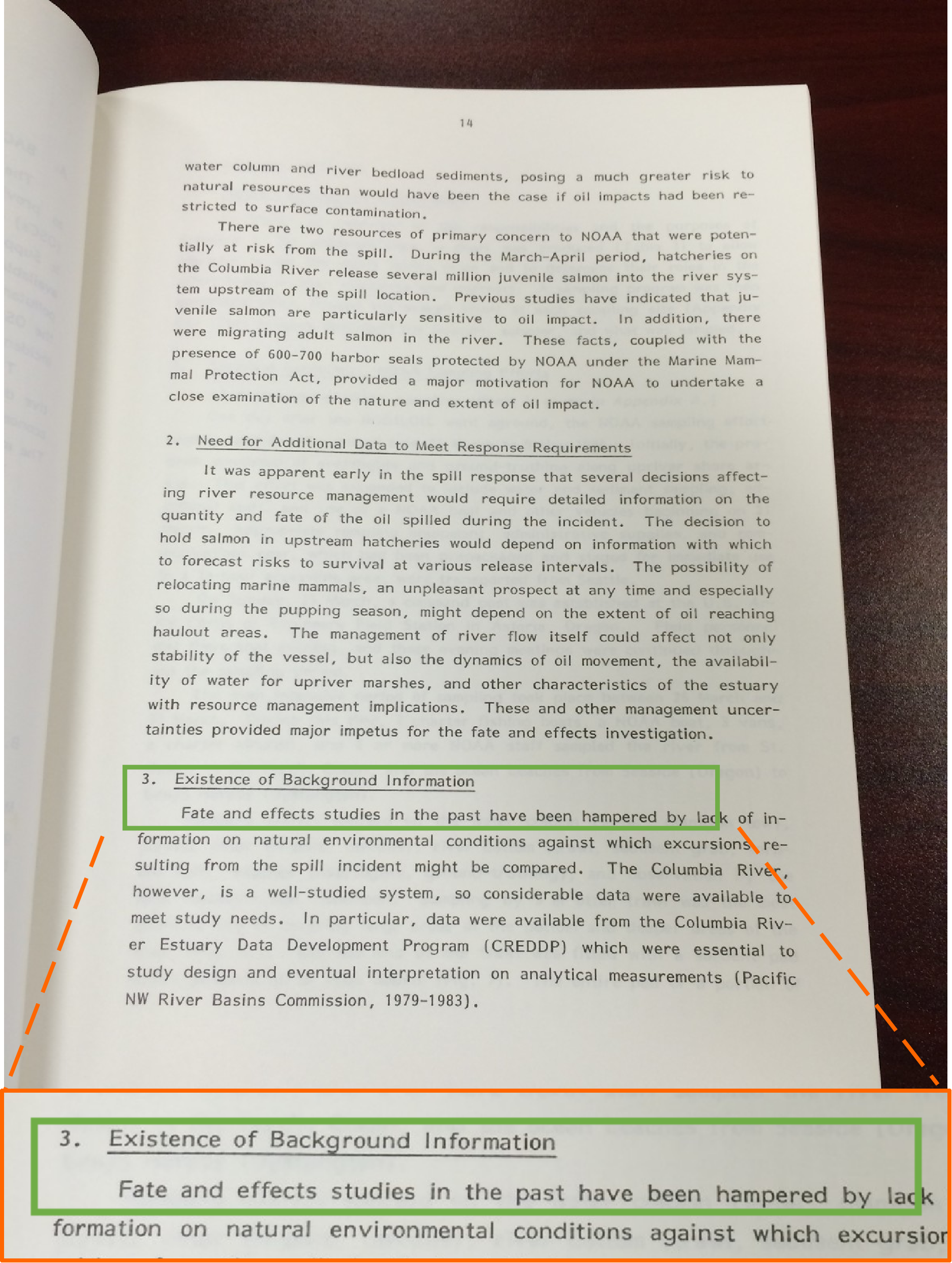} &
\includegraphics[width=0.235\linewidth]{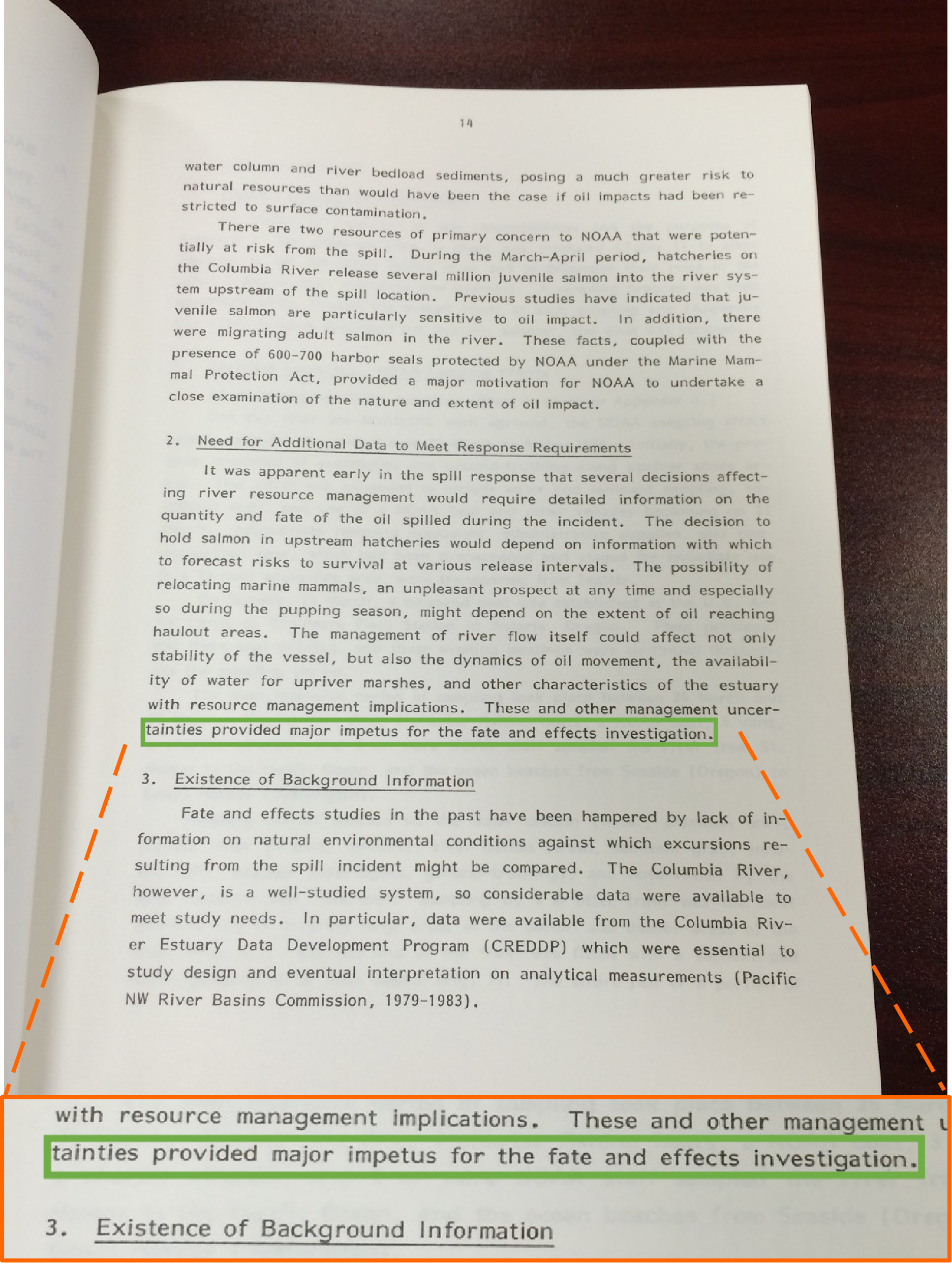} \\

\multicolumn{4}{l}{\scriptsize\textbf{Query:} tainties provided major impetus for the fate and effects investigation.} \\[0.35em]

\includegraphics[width=0.235\linewidth]{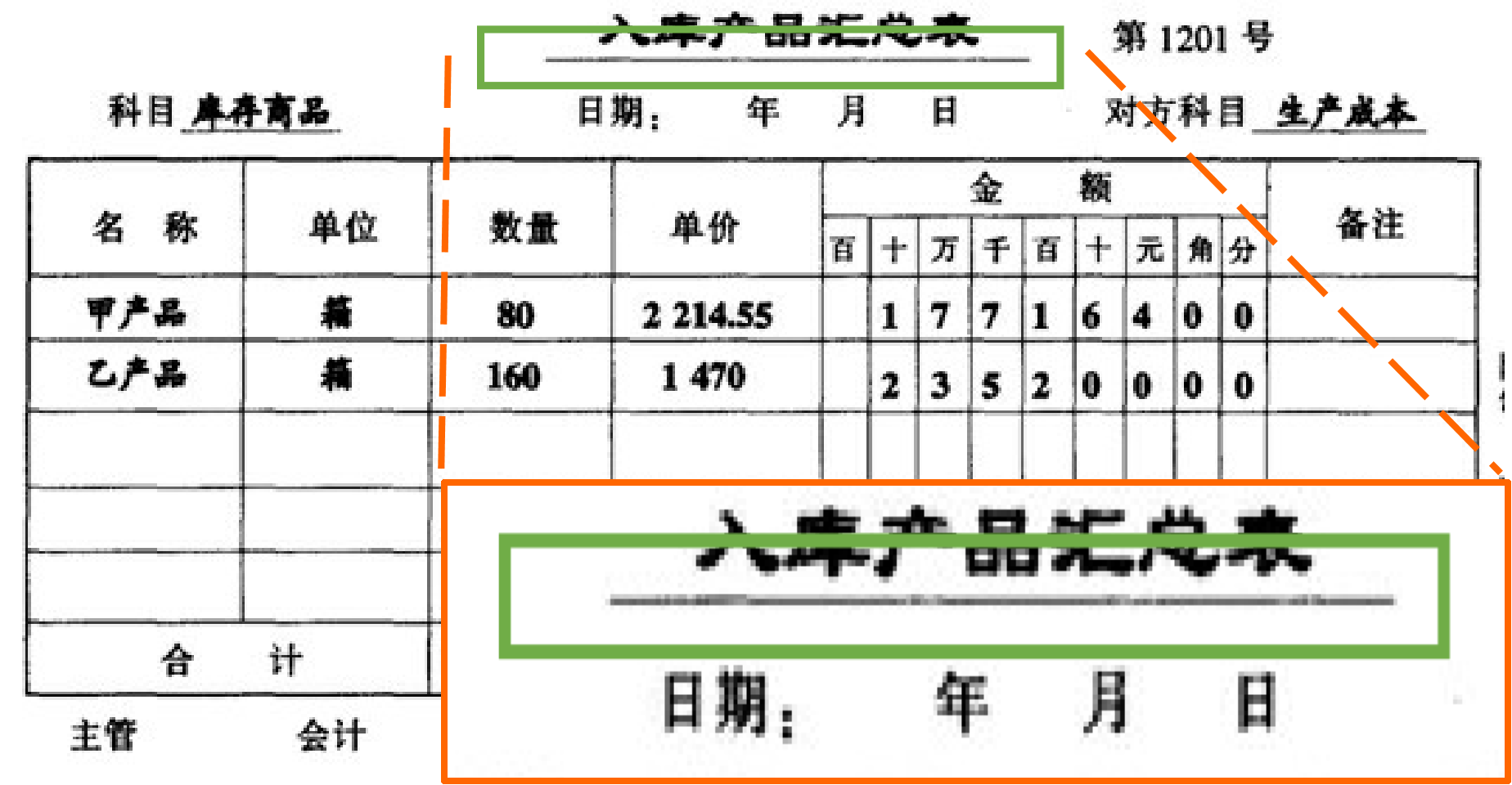} &
\includegraphics[width=0.235\linewidth]{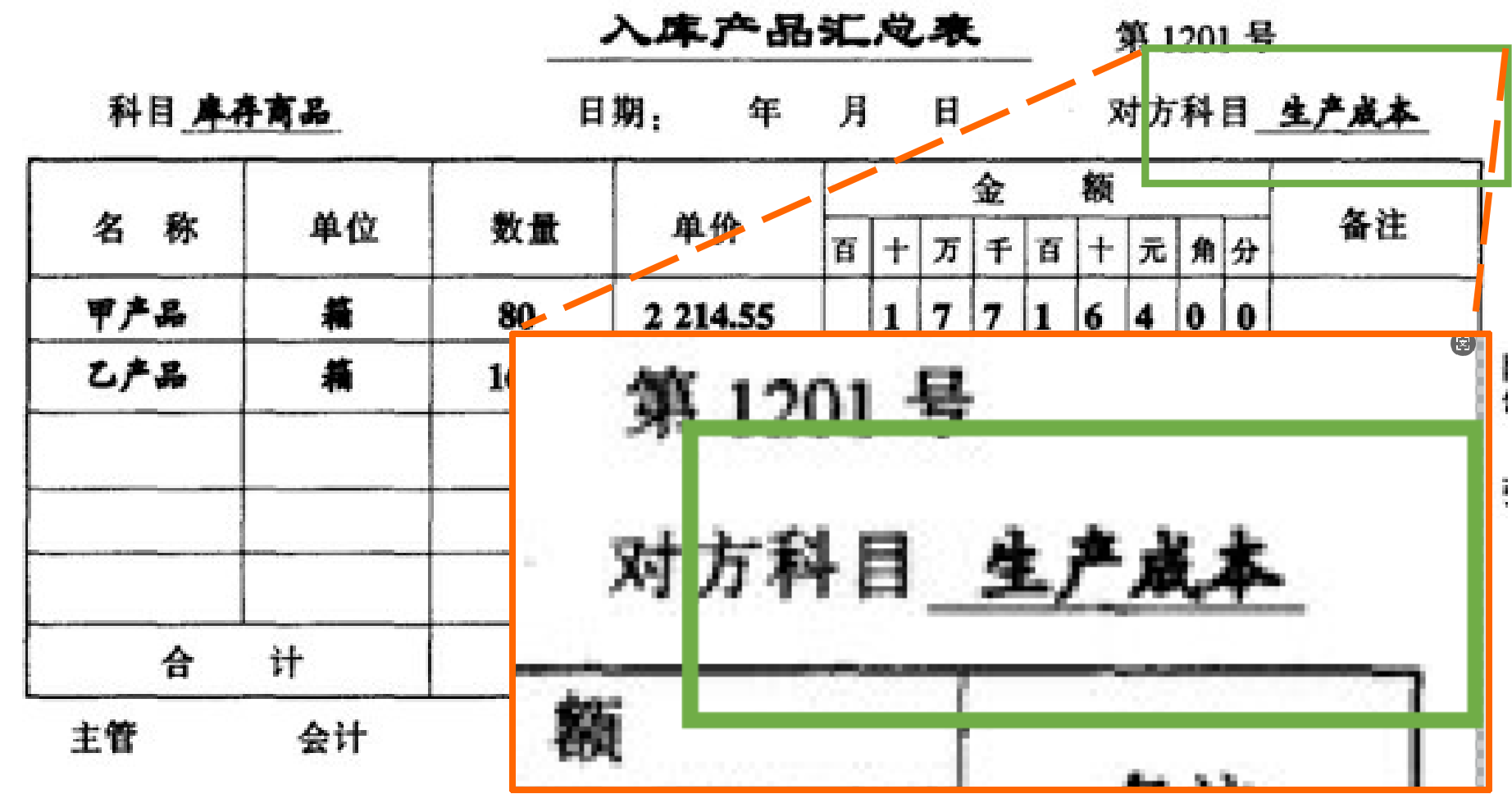} &
\includegraphics[width=0.235\linewidth]{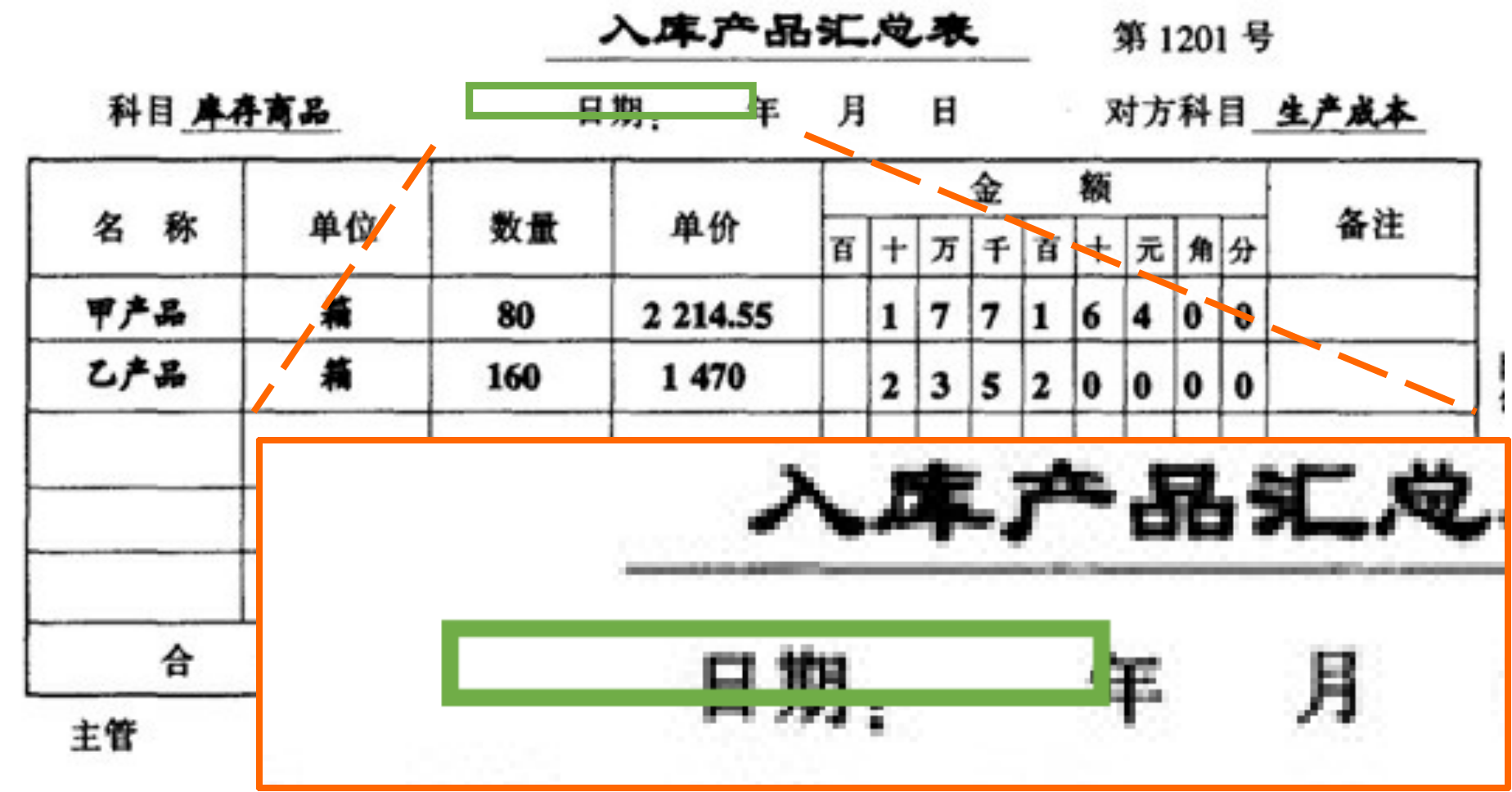} &
\includegraphics[width=0.235\linewidth]{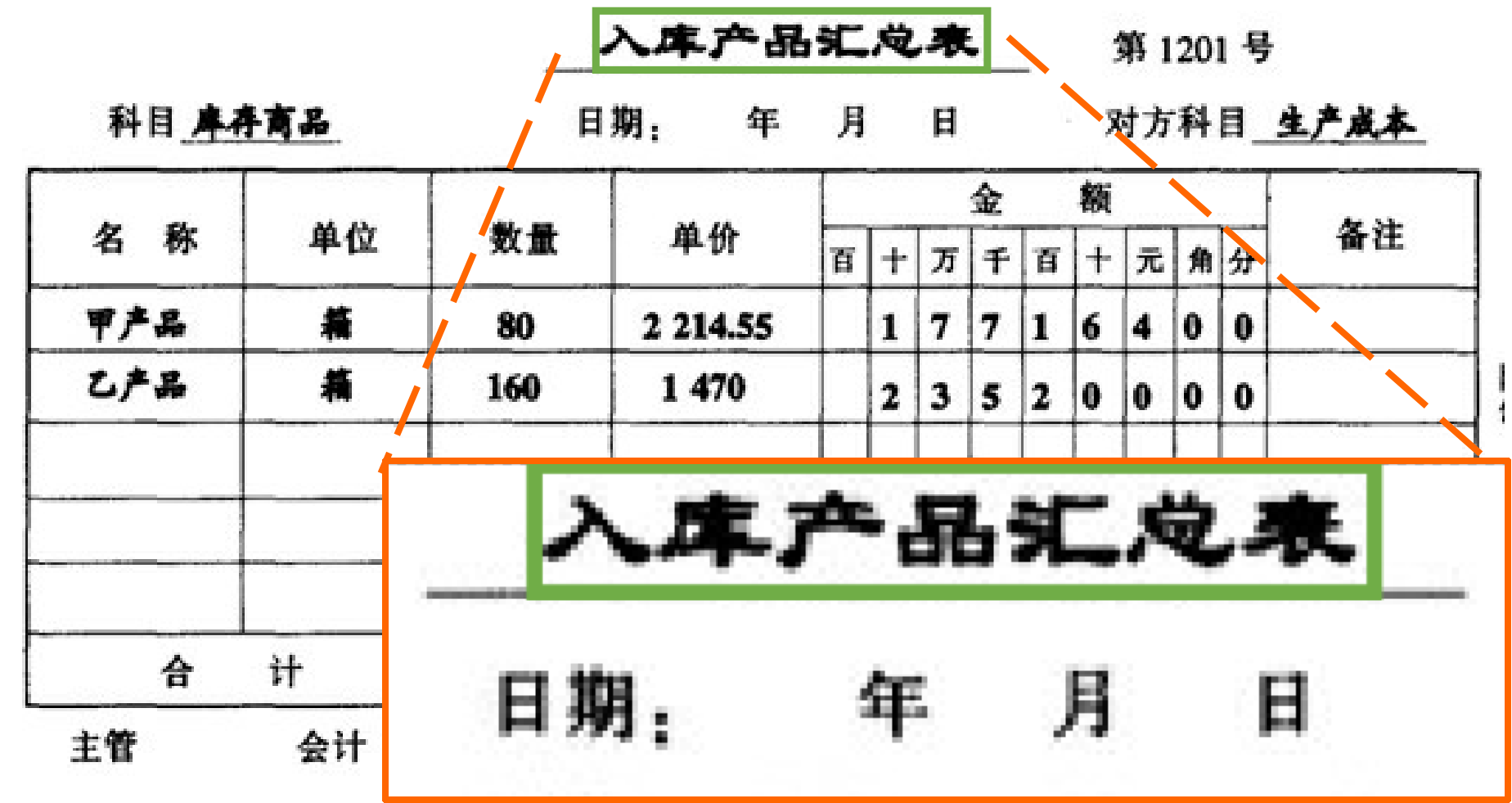} \\

\multicolumn{4}{l}{\scriptsize
\begin{CJK}{UTF8}{gbsn}
\textbf{Query:} 入库产品汇总表
\end{CJK}} \\

\end{tabular}

\vspace{0.3em}
\caption{
\textbf{Qualitative comparison of Text-to-Region (T2R) results.}
Columns correspond to GPT-5.2, Qwen3-VL-235B-A22B-Instruct (Qwen3-VL), Qwen2.5-VL-3B (Baseline), and Q-Mask-3B (Ours). Each row shows the predicted location for a given text query.
}
\label{fig:vis_t2_examples}
\end{figure*}

%% file: Supplymentary_Material/tables/r2t.tex
\begin{figure*}[htbp]
\centering
\small
\setlength{\tabcolsep}{6pt}

\begin{tabular}{cc}

\begin{minipage}[t]{0.36\linewidth}
\centering
\textbf{(a)} \\[2pt]
\includegraphics[width=\linewidth]{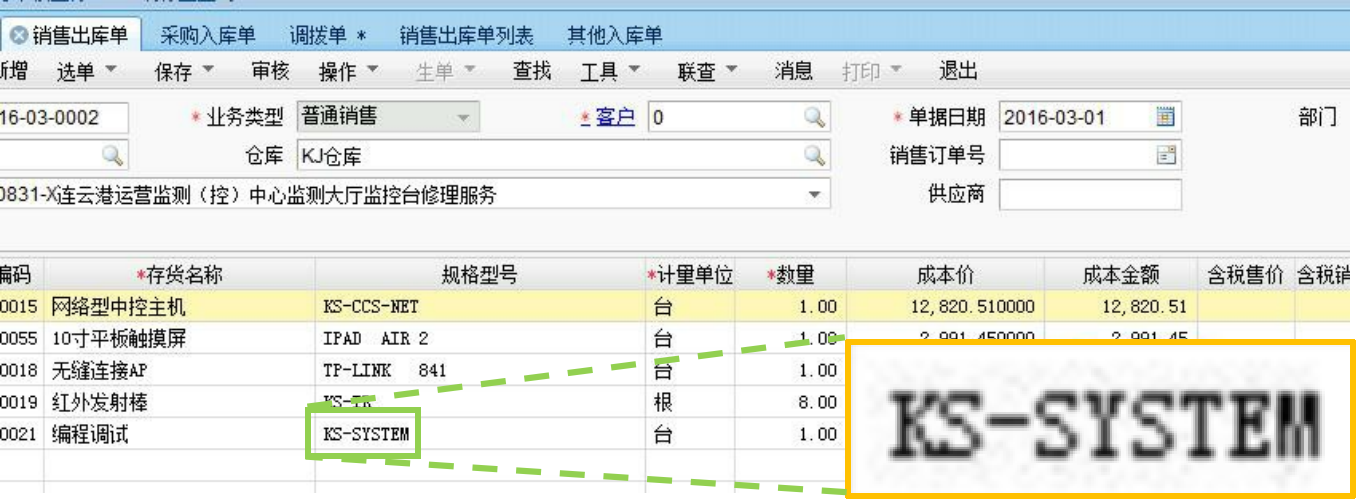}

\vspace{0.4em}

\raggedright \scriptsize
\textbf{Query:} What is the text at location [201.0, 267.0, 256.0, 276.0]? \\
\begin{CJK}{UTF8}{gbsn}
\textbf{GT:} KS-SYSTEM \\
\textcolor{gray}{GPT-5.2:} 0.00 \\
\textcolor{gray}{Qwen3-VL:} 编程调试 \\
\textcolor{gray}{Baseline:} KJ仓库 \\
\textcolor{blue}{\textbf{Ours:}} KS-SYSTEM
\end{CJK}
\end{minipage}

&
\begin{minipage}[t]{0.42\linewidth}
\centering
\textbf{(b)} \\[2pt]
\includegraphics[width=\linewidth]{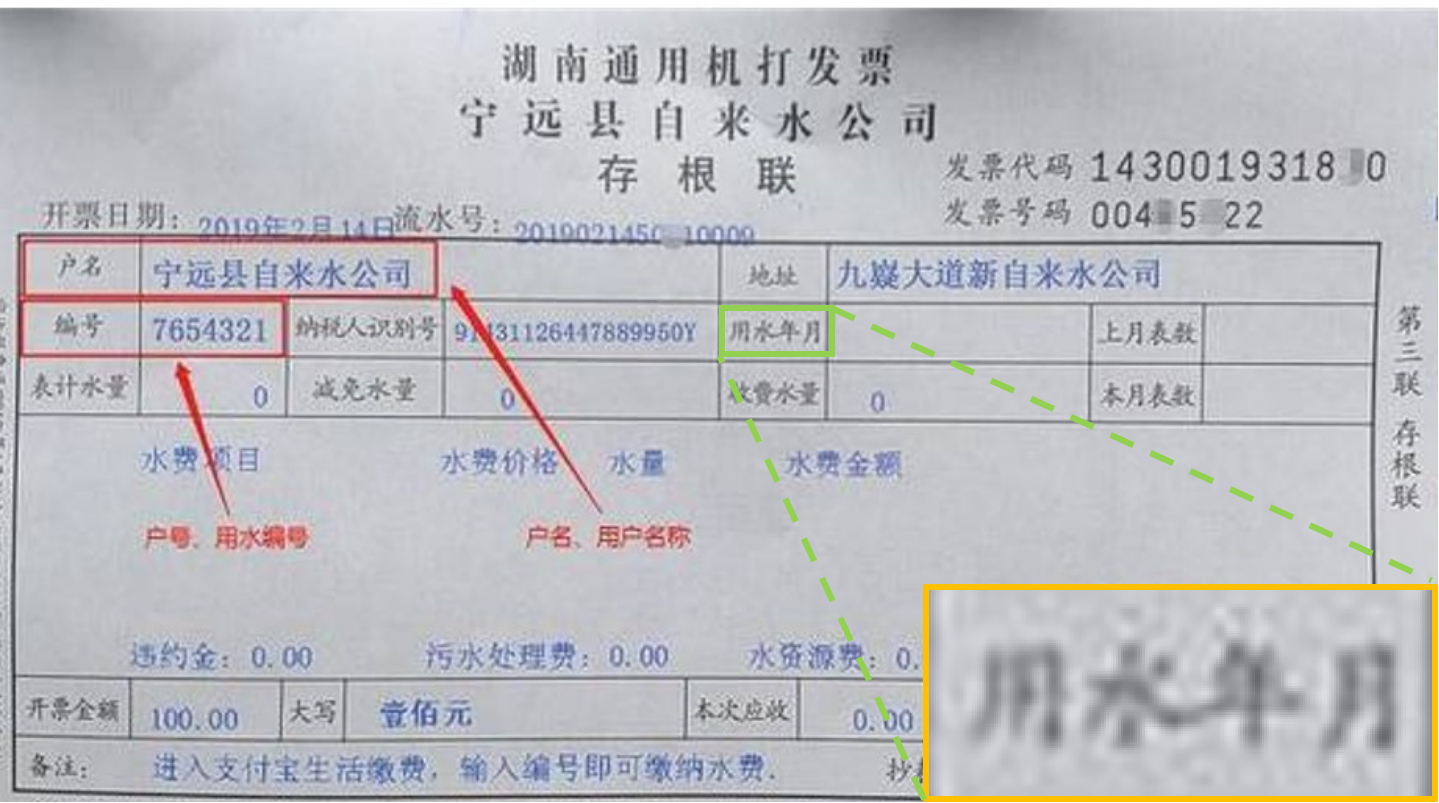}

\vspace{0.4em}

\raggedright \scriptsize
\textbf{Query:} What is the text at location [464.0, 201.0, 524.0, 220.0]? \\
\begin{CJK}{UTF8}{gbsn}
\textbf{GT:} 用水年月 \\
\textcolor{gray}{GPT-5.2:} 宁远县自来水公司 \\
\textcolor{gray}{Qwen3-VL:} 用水年月 \\
\textcolor{gray}{Baseline:} 九嶷大道新自来水公司 \\
\textcolor{blue}{\textbf{Ours:}} 用水年月
\end{CJK}
\end{minipage} \\[1em]

\begin{minipage}[t]{0.42\linewidth}
\centering
\textbf{(c)} \\[2pt]
\includegraphics[width=\linewidth]{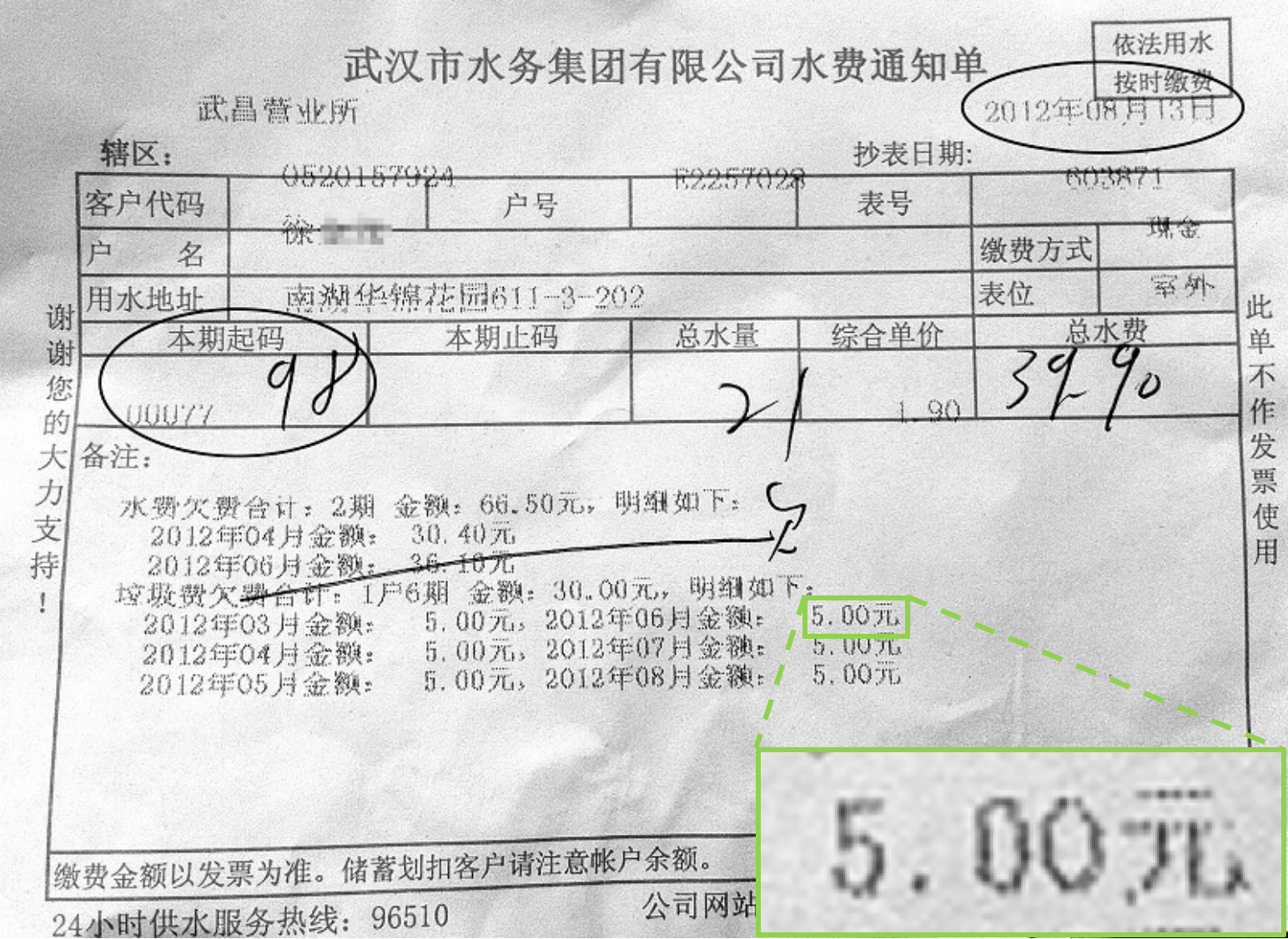}

\vspace{0.4em}

\raggedright \scriptsize
\textbf{Query:} What is the text at location [502.0, 373.0, 558.0, 389.0]? \\
\begin{CJK}{UTF8}{gbsn}
\textbf{GT:} 5.00元 \\
\textcolor{gray}{GPT-5.2:} 00 \\
\textcolor{gray}{Qwen3-VL:} 5.00元 \\
\textcolor{gray}{Baseline:} 3990 \\
\textcolor{blue}{\textbf{Ours:}} 5.00元
\end{CJK}
\end{minipage}

&
\begin{minipage}[t]{0.48\linewidth}
\centering
\textbf{(d)} \\[2pt]
\includegraphics[width=\linewidth]{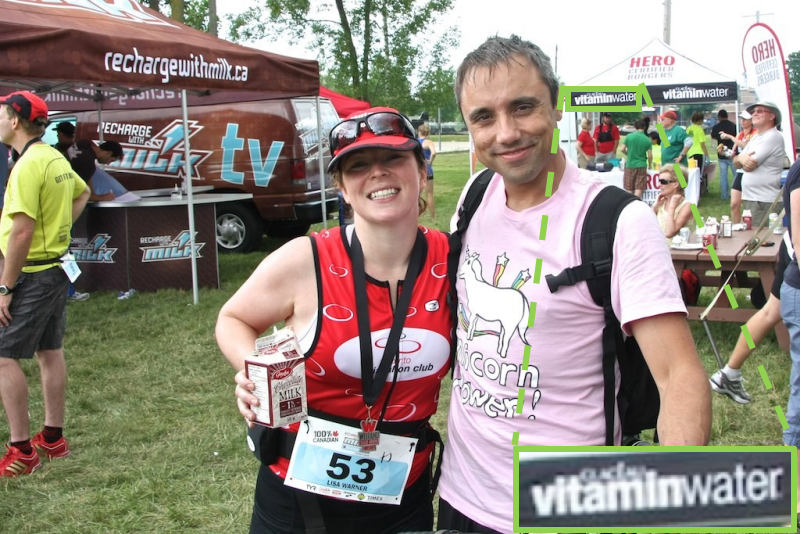}

\vspace{0.4em}

\raggedright \scriptsize
\textbf{Query:} What is the text at location [735.0, 117.0, 816.0, 133.0]? \\
\begin{CJK}{UTF8}{gbsn}
\textbf{GT:} vitaminwater \\
\end{CJK}
\textcolor{gray}{GPT-5.2:} Unicorn Power! \\
\textcolor{gray}{Qwen3-VL:} vitaminwater \\
\textcolor{gray}{Baseline:} HERO \\
\textcolor{blue}{\textbf{Ours:}} vitaminwater
\end{minipage} \\

\end{tabular}

\caption{
\textbf{Representative R2T cases.}
Q-Mask reads the target text more reliably in small, low-contrast, or densely surrounded regions. In all examples, the queried box is fixed, and the comparison highlights recognition robustness under identical spatial constraints.
}
\label{fig:vis_t1_examples}
\end{figure*}